\documentclass[]{openwam}
\setcitestyle{sort}

\newcommand{\finding}[2]{
    \begin{tcolorbox}[
        colback=blue!5,
        colframe=blue!70!black,
        arc=5pt,
        boxsep=5pt,
        left=2pt,
        right=2pt,
        top=2pt,
        bottom=2pt,
        boxrule=0.8pt,
        drop shadow=gray!50!white,
        enhanced jigsaw,
        before skip=12pt plus 2pt minus 2pt, 
        after skip=12pt plus 2pt minus 2pt,  
    ]
        \phantomsection\label{finding:#1}%
        \noindent\textbf{\textit{Finding #1:}} #2
    \end{tcolorbox}
}

\newcommand{\findingref}[2]{\hyperref[findingref:#1]{\textcolor{blue!50!black}{\textit{#2}}}}

\newcommand{\takeaway}[2]{
    \begin{tcolorbox}[
        colback=blue!5,
        colframe=blue!70!black,
        arc=5pt,
        boxsep=5pt,
        left=2pt,
        right=2pt,
        top=2pt,
        bottom=2pt,
        boxrule=0.8pt,
        drop shadow=gray!50!white,
        enhanced jigsaw,
        before skip=12pt plus 2pt minus 2pt, 
        after skip=12pt plus 2pt minus 2pt,  
    ]
        \phantomsection\label{takeaway:#1}%
        \noindent\textbf{\textit{Takeaway #1:}} #2
    \end{tcolorbox}
}

\newcommand{\question}[2]{
    \begin{tcolorbox}[
        enhanced,
        frame hidden,
        colback=blue!5,
        borderline west={2pt}{0pt}{blue!70!black},
        sharp corners,
        boxsep=0pt,
        left=8pt,
        right=4pt,
        top=8pt,
        bottom=8pt,
        before skip=12pt plus 2pt minus 2pt,
        after skip=8pt plus 2pt minus 2pt,
    ]
        \phantomsection\label{question:#1}%
        \noindent\textcolor{blue!70!black}{\textbf{\textit{Question #1:}}} #2
    \end{tcolorbox}
}

\newcommand{\suggref}[2]{\hyperref[suggestion:#1]{\textcolor{blue!50!black}{\textit{#2}}}}

\usepackage[dvipsnames]{xcolor}
\iftrue   
    \newcommand{\displaytodo}[1]{#1}
\else
    \newcommand{\displaytodo}[1]{}
\fi

\usepackage{makecell}
\usepackage{epsfig}
\usepackage{amsmath}
\usepackage{amssymb}
\usepackage{fontawesome5}
\usepackage{algorithmic}
\usepackage{colortbl} 
\usepackage{arydshln}
\usepackage{tikz}
\usetikzlibrary{tikzmark}
\usepackage{siunitx} 
\usepackage{mdframed}
\usepackage{tablefootnote} 
\usepackage{xspace}
\definecolor{tbdmagenta}{HTML}{FF1493}

\newcolumntype{g}{>{\columncolor{gray!30}}c}
\usepackage{wrapfig}
\usepackage{float}
\definecolor{citecolor}{HTML}{6B4AA0}

\definecolor{thupurple}{HTML}{57068C}
\definecolor{nusorange}{HTML}{EF7C00}
\definecolor{pkured}{HTML}{8C1515}
\definecolor{zjublue}{HTML}{003F88}
\definecolor{hkugreen}{HTML}{007B5E}
\definecolor{cuhkyellow}{HTML}{F65550}

\usepackage{fdsymbol}
\newcommand{\nusmark}{1}
\newcommand{\thumark}{2}
\newcommand{\pkumark}{3}
\newcommand{\hkumark}{4}
\newcommand{\zjumark}{5}
\newcommand{\cuhkmark}{6}
\newcommand{\sjtumark}{7}

\tcbuselibrary{skins}

\newenvironment{wamquote}{%
    \begin{list}{}{%
        \setlength{\leftmargin}{1.25em}%
        \setlength{\rightmargin}{1.25em}%
        \setlength{\topsep}{0.9em}%
        \setlength{\partopsep}{0pt}%
        \setlength{\parsep}{0pt}%
    }%
    \item\relax
    \fontfamily{ppl}\itshape\selectfont
}{%
    \end{list}
    \vspace{0.15em}
}

\newtcolorbox{introquestions}{
    enhanced,
    frame hidden,
    colback=blue!5,
    borderline west={2pt}{0pt}{blue!70!black},
    sharp corners,
    boxsep=0pt,
    left=8pt,
    right=8pt,
    top=6pt,
    bottom=4pt,
    before skip=7pt plus 2pt minus 1pt,
    after skip=10pt plus 2pt minus 2pt,
}

\newcommand{\introquestion}[2]{%
    \noindent\textcolor{blue!70!black}{\textbf{\textit{Question #1:}}} #2\par\vspace{3pt}%
}

\makeatletter
\DeclareRobustCommand{\openwamalpha}{%
    OpenWAM-\ensuremath{%
        \ifx\f@series\mddefault
            \alpha
        \else
            \boldsymbol{\alpha}%
        \fi
    }%
}
\makeatother
\definecolor{userbg}{RGB}{245, 245, 245}
\definecolor{userborder}{RGB}{210, 229, 255}
\definecolor{userfont}{RGB}{0, 0, 0}

\tcbset{
  enhanced,
  boxrule=1pt,
  colback=userbg,
  colframe=userborder,
  fonttitle=\bfseries,
  coltitle=userfont,
  boxsep=5pt,
  arc=5pt,
  outer arc=5pt,
  left=10pt,
  right=10pt,
  top=2pt,
  bottom=2pt,
  attach boxed title to top left={yshift=-0.25\baselineskip-1mm,xshift=2mm},
  boxed title style={
    colback=userborder,
    sharp corners,
    rounded corners=northeast,
    arc=3pt,
    outer arc=3pt,
    boxrule=0pt,
    boxsep=2pt,
  },
}
\usepackage{pgf}
\usepackage{adjustbox}
\usepackage[shortlabels,inline]{enumitem} 
\usepackage{array}
\definecolor{listcolor}{RGB}{50,120,230}

\definecolor{methodgreen}{RGB}{45, 106, 79} 

\usepackage{mathtools}
\usepackage{amsthm}

\titlespacing*{\paragraph} {0pt}{1.25ex plus 1ex minus .2ex}{0.75em}

\definecolor{DiTblue1}{HTML}{AEC6FE}
\definecolor{DiTblue2}{HTML}{9AB8FE}
\definecolor{DiTblue3}{HTML}{86AAFE}
\definecolor{DiTblue4}{HTML}{729CFE}
\definecolor{DiTblue5}{HTML}{5D8DFD}

\definecolor{DiTgreen1}{HTML}{B3CDA2}
\definecolor{DiTgreen2}{HTML}{A8C695}
\definecolor{DiTgreen3}{HTML}{9EBF88}
\definecolor{DiTgreen4}{HTML}{93B87A}
\definecolor{DiTgreen5}{HTML}{88B06D}

\definecolor{DiTyellow1}{HTML}{F9CB8A}
\definecolor{DiTyellow2}{HTML}{F8C277}
\definecolor{DiTyellow3}{HTML}{F7BA64}
\definecolor{DiTyellow4}{HTML}{F6B150}
\definecolor{DiTyellow5}{HTML}{F5A83D}

\definecolor{DiTred1}{HTML}{FBA09D}
\definecolor{DiTred2}{HTML}{FA8480}
\definecolor{DiTred3}{HTML}{F86762}
\definecolor{DiTred4}{HTML}{F65550}
\definecolor{DiTred5}{HTML}{F5433D}
\definecolor{EncoderSigLIP 2}{HTML}{729CFE}  
\definecolor{EncoderWebSSL}{HTML}{A8C695}   
\definecolor{EncoderDiNOv2}{HTML}{729CFE}   
\definecolor{EncoderSDVAE}{HTML}{F86762}    
\definecolor{EncoderFLUX}{HTML}{F65550}     
\definecolor{EncoderRawPixel}{HTML}{F7BA64} 

\definecolor{DataDrivenPurple}{HTML}{897bc1}
\definecolor{ExtendDiTPurple}{HTML}{D4B6FC}

\theoremstyle{plain}

\theoremstyle{definition}

\theoremstyle{remark}

\makeatletter
\newif\if@obnosep
\renewcommand\author[2][]{%
  \if@obnosep
    \begingroup
      \let\protect\@unexpandable@protect
      \xdef\authorlist{\expandafter{\authorlist}\protect\authorformat[#1]{#2}}%
    \endgroup
    \@obnosepfalse
  \else
    \addtolist[#1]{#2}{\authorlist}{\authorformat}{, }%
  \fi
}
\newcommand\authorbreak{%
  \begingroup
    \let\protect\@unexpandable@protect
    \xdef\authorlist{\expandafter{\authorlist},\protect\\}%
  \endgroup
  \@obnoseptrue
}
\newif\if@afnosep
\renewcommand\affiliation[2][]{%
  \if@afnosep
    \begingroup
      \let\protect\@unexpandable@protect
      \xdef\affiliationlist{\expandafter{\affiliationlist}\protect\affiliationformat[#1]{#2}}%
    \endgroup
    \@afnosepfalse
  \else
    \addtolist[#1]{#2}{\affiliationlist}{\affiliationformat}{, }%
  \fi
}
\newcommand\affiliationbreak{%
  \begingroup
    \let\protect\@unexpandable@protect
    \xdef\affiliationlist{\expandafter{\affiliationlist},\protect\\}%
  \endgroup
  \@afnoseptrue
}
\makeatother
\renewcommand\affiliationformat[2][]{{\small $^{#1}$#2}}

\title{OpenWAM: An Open, Modular Exploration Towards Systematic World--Action Model Pretraining}

\author[\nusmark, *, \ddagger]{Yuran~Wang}
\author[\thumark, *, \ddagger]{Siqiao~Huang}
\author[\pkumark, *]{Mingleyang~Li}
\author[\pkumark, *]{Chenhao~Zhang}
\author[\pkumark, *]{Jiaqi~Liang}
\author[\hkumark]{Weiyang~Jin}
\authorbreak
\author[\pkumark]{Yue~Chen}
\author[\zjumark]{Xuemin~Chi}
\author[\cuhkmark]{Donghao~Zhou}
\author[\pkumark]{Qize~Yu}
\author[\pkumark]{Yu-Kai~Wang}
\author[\pkumark]{Yuhan~Rui}
\author[\thumark]{Shenzhe~Yao}
\authorbreak
\author[\hkumark]{Zhen~Yuan}
\author[\pkumark]{Zhenhao~Shen}
\author[\pkumark]{Kefei~Zhu}
\author[\hkumark]{Zijie~Zhu}
\author[\sjtumark]{Ning~Gao}
\author[\pkumark]{Xiaowei~Chi}
\author[\thumark]{Guanqi~He}
\authorbreak
\author[\pkumark]{Shanghang~Zhang}
\author[\pkumark]{Hao~Dong}
\author[\nusmark, \dagger]{Lin~Shao}
\author[\thumark, \dagger]{Hang~Zhao}

\affiliation[\nusmark]{National University of Singapore}
\affiliation[\thumark]{Tsinghua University}
\affiliation[\pkumark]{Peking University}
\affiliation[\hkumark]{The University of Hong Kong}
\affiliationbreak
\affiliation[\zjumark]{Zhejiang University}
\affiliation[\cuhkmark]{The Chinese University of Hong Kong}
\affiliation[\sjtumark]{Shanghai Jiao Tong University}
\contribution[*]{Equal contribution}
\contribution[\ddagger]{Project lead}
\contribution[\dagger]{Equal advising}

\abstract{
World--Action Models inherit world knowledge from video-generative priors, and channel it into executable control signals through embodied experience. Existing systems, however, are monolithic: the generative backbone, visual representation, architecture, information flow, inference procedure, and training data are tightly coupled, obscuring which design choices matter and why. We introduce \texttt{OpenWAM}, an open research stack that turns world--action pretraining into a controlled experimental program. \texttt{OpenWAM-Infra} factorizes the WAM design space into composable modules with unified training, inference, deployment, and evaluation. On this substrate, \texttt{OpenWAM-Study} examines three questions through controlled experiments: what to inherit, how world and action learning interact, and how their synergy scales; and distills three principles: upstream knowledge transfers through a sufficiently capable generative backbone and a compact, information-rich latent space; world--action synergy requires dedicated action capacity, explicit world-to-action information flow, and synchronized joint denoising; and embodied pretraining principally improves out-of-domain generalization, with one-stage co-training over egocentric and robot data integrating world coverage and action grounding. Composing these principles, we build \texttt{\openwamalpha{}}, an open WAM pretrained on roughly 6{,}400 hours of egocentric human and robot data and evaluated across simulation and real-world benchmarks. Across the eight simulation benchmarks and the real-robot experiments, which together span embodiments from single-arm and bimanual manipulation to dexterous hands, \openwamalpha{} delivers consistently excellent performance, sustaining its top-tier standing from simulation to the physical world. We release the full stack, including infrastructure, evaluation protocols, pretrained models, and data recipes, to facilitate future research.
}

\project{\href{https://openwam-official.github.io/}{\sffamily \fontsize{8.8pt}{11pt}\selectfont \texttt{https://openwam-official.github.io/}}}
\code{\href{https://github.com/OpenWAM-Official/OpenWAM}{\sffamily \fontsize{8.8pt}{11pt}\selectfont \texttt{https://github.com/OpenWAM-Official/OpenWAM}}}
\model{\href{https://huggingface.co/OpenWAM}{\sffamily \fontsize{8.8pt}{11pt}\selectfont \texttt{https://huggingface.co/OpenWAM}}}

\begin{document}

\newcommand{\wamrule}{\par\noindent
  \textcolor{wamtitle!35}{\tikz{\draw[dashed,dash pattern=on 2.4pt off 2.2pt,%
    line width=0.5pt](0,0)--(\linewidth,0);}}\par}
\makeatletter
\patchcmd{\mymaketitle}{\tcbset{top=0.5cm}}{\tcbset{top=0.26cm}}{}{}
\patchcmd{\mymaketitle}{\tcbset{bottom=0.5cm}}{\tcbset{bottom=0.26cm}}{}{}
\patchcmd{\mymaketitle}{\vskip 0.5cm}{\vskip 0.15cm}{}{}
\patchcmd{\mymaketitle}{\abstractlist\par}{%
  {\setlength{\parskip}{0cm}%
   \vskip 0.07cm \wamrule \vskip 0.07cm
   {\centering\includegraphics[width=\linewidth,trim=1 10 1 14,clip]{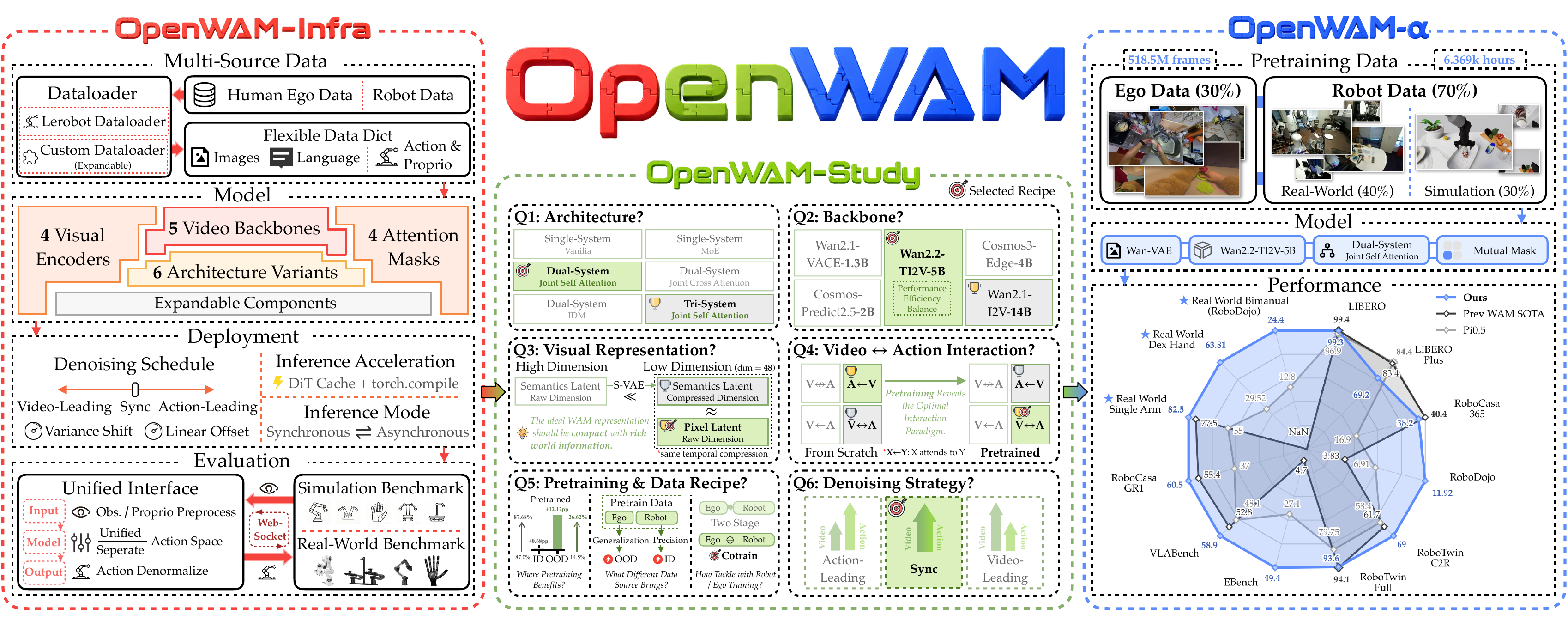}\par}%
   \vskip 0.02cm
   {\captionsetup{type=figure,skip=2pt}\caption{\textbf{Overview of \texttt{OpenWAM}.} \textbf{OpenWAM-Infra} (\emph{left}) factorizes world--action modeling into composable modules with unified training, deployment, and evaluation; \textbf{OpenWAM-Study} (\emph{middle}) resolves the design space through controlled questions and distills a pretraining recipe; \textbf{\openwamalpha{}} (\emph{right}), pretrained on 518.5M frames of egocentric and robot data, sustains top-tier performance from simulation to the real world.}\label{fig:teaser}}%
   \vskip 0.07cm \wamrule}%
}{}{}
\makeatother
\enlargethispage{0.60cm}
\vspace*{-0.98cm}
\maketitle
\vspace{0.30cm}
{\centering\sffamily\bfseries Abstract\par}
\vspace{0.05cm}
\abstractlist\par
\clearpage
{
  \hypersetup{linkcolor=wamaccent}
  \tableofcontents
}
\clearpage

\section{Introduction}

\begin{wamquote}
    \textquotedblleft Knowledge is the beginning of action, action is the completion of knowledge.\textquotedblright\par
    \vspace{0.5em}
    \raggedleft
    --- \textnormal{Yang-ming Wang}, \textquotedblleft Instructions for Practical Living\textquotedblright{} \textnormal{\citep{ymw}}\par
\end{wamquote}
\vspace{-0.6em}

Intelligence requires more than recognizing the world: an embodied agent must anticipate how the world changes and act to bring about desired changes. Modern vision and vision--language models have learned rich semantic representations from large-scale image--text data \citep{radford2021learning, caron2021emerging, tong2024cambrian}. Video generation models go one step further by learning to synthesize how visual worlds may evolve over time \citep{ho2022video, videoworldsimulators2024}. An embodied system, however, must learn not only \emph{what can happen in the world}, but also \emph{what actions it can take to make it happen}.

This distinction exposes a fundamental data asymmetry in embodied learning. Videos of the changing world are abundant, whereas robot trajectories with executable action labels remain comparatively scarce \citep{ye2026data}. Earlier approaches such as ACT \citep{zhao2023learning} and Diffusion Policy \citep{chi2025diffusion} largely learn visual regularities and control together solely from robot demonstrations. More recently, Vision--Language--Action (VLA) models \citep{brohan2023rt,kim2024openvla,black2024pi_0} instead inherit semantic and linguistic knowledge from pretrained vision--language models. World--Action Models (WAMs) \citep{pai2025mimic,ye2026world,li2026causal} offer a different warm start: they inherit a generative prior over visual dynamics from video generation pretraining and adapt it through embodied experience. Because video generation is explicitly trained to model temporal evolution, it provides a direct starting point for learning how actions interact with physical change. In this sense, a WAM \emph{inherits world knowledge from video generation priors, then learns how to take actions in this world through embodied experience}.

The promise of a WAM, however, lies not merely in initializing a policy with a video model or attaching an action head to a video generator. Its central hypothesis is that world prediction and action generation can be learned in \emph{synergy}: world modeling supplies structured knowledge of states, dynamics, and possible futures that can inform action, while action learning focuses the model on changes that matter for control. However, realizing this synergy is nontrivial. Useful knowledge may reside in different parts of an upstream video model; nominally joint world and action prediction may still lack an effective information path; and a design that fits one training domain may fail to retain its advantage across new scenes or embodiments. These challenges lead to three central questions: 
\begin{introquestions}
    \introquestion{1}{\textbf{What world knowledge should a WAM inherit?}}
    \introquestion{2}{\textbf{How can we create synergy between world and action learning?}}
    \introquestion{3}{\textbf{How can this synergy be scaled across domains?}}
\end{introquestions}

Answering these questions is difficult with existing monolithic systems, where the generative backbone, visual representation, model architecture, information flow, inference procedure, and composition of training data are often tightly coupled \citep{kim2026cosmos,bi2026motus,zhang2026native}. We therefore introduce \textbf{OpenWAM}, an open research stack for systematically developing World--Action Models. \textbf{OpenWAM-Infra} factorizes the WAM design space into modular components, while providing unified training, inference, deployment, and evaluation across domains and embodiments (\Cref{sec:infra}). This modularity turns world--action modeling from a collection of coupled implementation choices into a controlled experimental program.

Building on this framework, \textbf{OpenWAM-Study} investigates the design principles underlying World--Action Models (\Cref{sec:recipe}). Our study produces three main findings. First, upstream world knowledge transfers most effectively through a sufficiently capable generative backbone and a compact, information-rich visual latent space. Second, world--action synergy does not emerge from parameter count or joint prediction alone. It is best fostered with sufficient action-specific capacity, explicit world-to-action information flow, and a joint test-time denoising schedule. Third, embodied pretraining primarily improves out-of-domain generalization rather than in-domain fitting: human egocentric video broadens world coverage, robot trajectories provide executable action knowledge, and their joint training offers the strongest practical integration strategy in our experiments. Together, these findings suggest a practical recipe from \emph{inheriting world knowledge}, to \emph{enabling world--action synergy}, to \emph{scaling that synergy across domains}.

Finally, we compose the resulting design principles into \textbf{\openwamalpha{}}, an open pretrained World--Action Model (\Cref{sec:pretrain}). Pretrained on 518M frames ($\approx$6{,}400 hours) of egocentric human and robot data through a unified 80-D action space, \openwamalpha{} demonstrates that the recipe distilled from controlled settings remains effective when scaled across heterogeneous data, domains, and embodiments. It delivers consistently strong results on eight simulation benchmarks covering five embodiment categories, and preserves this standing in real-robot experiments on single-arm, bimanual, and dexterous-hand platforms. Beyond the scores themselves, these large-scale evaluations also distill further insights into the design and scaling behavior of embodied foundation models. To let the community reproduce and extend these findings, we release alongside the model the full stack: the infrastructure, evaluation protocols, pretrained weights, and data recipes.

In summary, our major contributions are as follows:
\begin{itemize}[leftmargin=1.5em]
    \item \textbf{OpenWAM-Infra: a Modular Infrastructure for World--Action Modeling.} It factorizes model, representation, training, inference, deployment, and evaluation choices, enabling controlled comparison across WAM designs and embodiments (\Cref{sec:infra}).
    \item \textbf{OpenWAM-Study: Design Principles for World--Action Synergy.} Through controlled studies of upstream priors, architectural capacity, information flow, denoising, and multi-domain pretraining, we identify how world knowledge can interact productively with action learning (\Cref{sec:recipe}).
    \item \textbf{\openwamalpha{}: a Pretrained World--Action Model.} It instantiates and scales the derived principles into an open model for evaluating generalization and efficiency across domains and embodiments (\Cref{sec:pretrain}).
\end{itemize}

\section{Related Work}

\paragraph{World--Action Models.}World models \citep{lecun2022path, ha2018world} learn predictive structure from observations and have long supported control through planning \citep{zhou2024dino, maes2026leworldmodel,huang2026nano}, model-based reinforcement learning \citep{hafner2019dream, m2023model}, and policy evaluation \citep{huang2025vid2world,wang2026interactive}. World--Action Models (WAMs) \citep{ye2026world,pai2025mimic,li2026causal} more directly connect this predictive capacity to executable behavior by serving as a policy model. Whereas VLA models \citep{brohan2023rt,kim2024openvla,black2024pi_0} primarily inherit semantic and linguistic knowledge from vision--language pretraining \citep{beyer2024paligemma, bai2025qwen3vltechnicalreport}, WAMs initialize from video-generative priors so that action learning begins with a model with rich visual dynamical priors \citep{wan2025wan, ali2025world}. Yet existing systems remain largely \textit{monolithic}, where changes in model architecture, training procedure, data recipe, and sampling schedule are often coupled. Consequently, it remains unclear which components transfer world knowledge, which interactions create world--action synergy, and which benefits persist across domains. \textbf{OpenWAM} exposes these coupled choices as controlled variables and organizes them around precisely these three questions.

\paragraph{Open Research Ecosystems for Generalist Robot Policy Learning.}Open models and codebases have made generalist robot learning increasingly accessible. OpenVLA \citep{kim2024openvla} established an open-weight pretrained baseline, while StarVLA \citep{community2026starvla} provides a modular and performant platform for varied design choices. StarVLA-$\alpha$ \citep{ye2026starvla} complements this breadth with a pretrained model of minimalist design, and XPolicyLab \citep{community2026xpolicylab} contributes a unified standard and open ecosystem for policy evaluation and deployment. These efforts have significantly reduced development complexity in the VLA research community; however, in the WAM community, such open research ecosystems remain largely absent. A modular system in this realm accompanied by a strong pretrained model would help democratize research, as well as serve as a principled foundation for understanding and scaling world--action model pretraining.

\paragraph{Towards a Scientific Understanding of Model Design.} A growing line of work treats model design as an empirical science \citep{allen2026physics,karras2022elucidating,mckinzie2024mm1,convnext,wen2026fantastic}: decomposing a complex system into controlled variables, testing the mechanisms behind observed gains, deriving a recipe, and validating whether it survives scale. In multimodal learning, Cambrian-1 \citep{tong2024cambrian} and Beyond Language Modeling \citep{tong2026beyond} systematically study visual representations, modality-specific capacity, data composition, and unified pretraining; Towards Physics of Multimodal Pretraining \citep{han2026towards} further isolates knowledge flow, synergy versus competition, and the timing of modality unification. In robot learning, analyses around Action Chunking \citep{simchowitz2025pitfalls, zhang2025action, lazzati2026does} and Generative Control Policies \citep{pan2026much} have substantially reshaped the community's understanding of these topics. At the data and system level, Large Behavior Models \citep{barreiros2026careful}, LBM co-train \citep{lin2026systematic}, StarVLA-$\alpha$ \citep{ye2026starvla}, and OpenHLM \citep{hu2026openhlm} similarly use controlled comparisons to study multitask transfer, heterogeneous supervision, action design, and embodiment interfaces. \textbf{OpenWAM} brings this methodology to WAMs: \textbf{OpenWAM-Infra} builds the substrate for controlled experiments, \textbf{OpenWAM-Study} turns them into controlled scientific questions about inheritance, synergy, and scaling, and \textbf{\openwamalpha{}} scales the resulting recipe under heterogeneous multi-domain pretraining.

\section{OpenWAM-Infra: A Modular Infrastructure for World--Action Modeling}
\label{sec:infra}
Most existing world--action models differ substantially in architecture and infrastructure implementation, with no shared standard; since each system is built around a single model design, its model, training, serving, and evaluation components are likewise organized idiosyncratically and are often tightly coupled. This brings two problems: 1) such codebases are difficult for users to extend or build upon, and 2) the coupling among components allows modules to interfere with one another, confounding the conclusions drawn from controlled comparisons. \textbf{OpenWAM-Infra} addresses both problems by factoring world--action modeling into four decoupled components with explicit interfaces: a \emph{composable model} assembled from interchangeable encoders, stream backbones, and visibility attention masks (\Cref{sec:infra_model}); a \emph{training runtime} that trains every such model with one trainer (\Cref{sec:infra_train}); a \emph{deployment runtime} that serves every resulting checkpoint through one policy server (\Cref{sec:infra_deploy}); and an \emph{evaluation protocol} through which every benchmark reaches that server (\Cref{sec:eval-protocol}). These components are either mutually independent or related by strict one-way dependencies, which keeps the codebase straightforward to extend, insulates modules from mutual interference, and further provides the substrate on which \textbf{OpenWAM-Study} (\Cref{sec:recipe}) conducts controlled experiments and \textbf{\openwamalpha{}} (\Cref{sec:pretrain}) is instantiated at scale.

\subsection{Composable Model}
\label{sec:infra_model}

\begin{figure}[htbp]
    \centering
    \includegraphics[width=\linewidth]{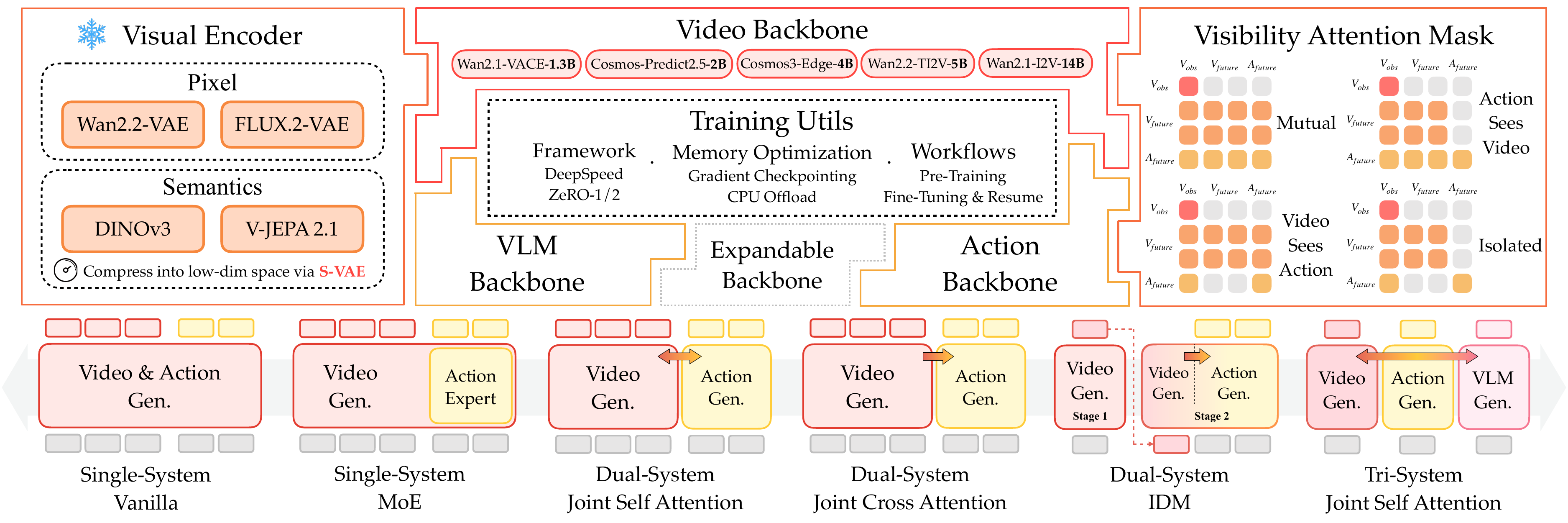}
    \caption{\textbf{OpenWAM Model Infra.} \emph{Top}: the three classes of interchangeable modules: the visual encoder $\mathcal{E}$ (left); the stream backbones $\mathcal{S}$ (middle); and the visibility attention mask $\mathcal{M}$ (right). The central Training Utils panel summarizes the utilities of the training runtime (\Cref{sec:infra_train}). \emph{Bottom}: the composition rule $C$ assembles the modules into six architecture variants across the Single-System, Dual-System, and Tri-System families.}
    \label{fig:openwam-infra}
\end{figure}

As shown in \Cref{fig:openwam-infra}, OpenWAM-Infra organizes a World--Action Model (WAM) as three classes of interchangeable modules that a composition rule $C$ assembles into a concrete \textbf{architecture}, written $C(\mathcal{E},\mathcal{S},\mathcal{M})$:
\begin{itemize}[leftmargin=1.5em]
    \item \textbf{Visual Encoder} $\mathcal{E}$: It maps observations into the latent
    sequences the world stream predicts;
    \item \textbf{Stream Backbones} $\mathcal{S}$: It processes the model's
    token streams: a world stream $\mathcal{W}$, an action stream $\mathcal{A}$, optionally an
    understanding stream $\mathcal{U}$, and any further streams a design may introduce;
    distinct streams may share a single backbone;
    \item \textbf{Visibility Attention Mask} $\mathcal{M}$: It specifies the attention relations among streams and tokens, i.e. which tokens may attend to
    which, both within and across streams.
\end{itemize}

\paragraph{Visual Encoder.}
The encoder $\mathcal{E}$ maps observations into the latent sequence the world stream predicts and is always kept frozen; different encoders capture different information in their latents and in turn induce different world-stream behavior. In most cases, a video backbone is accompanied by a natively matched encoder, in which case the pretrained parameters of the base DiT are reused directly, retaining the full benefit of pretraining. Beyond this default, OpenWAM-Infra additionally supports pluggable encoders: swapping $\mathcal{E}$ alters the latent representation to be predicted, and the base DiT can further be re-initialized to exclude the influence of pretrained parameters. Currently, OpenWAM-Infra provides two reconstructive encoders, trained on pixel-reconstruction objectives: Wan2.2-VAE~\citep{wan2025wan} and FLUX.2-VAE~\citep{bfl2025flux2}, and two representation encoders \citep{zheng2026diffusion}: DINOv3~\citep{simeoni2025dinov3} and V-JEPA 2.1~\citep{murlabadia2026vjepa2_1}, along with an optional S-VAE \citep{zhang2025both} module that compresses the latent dimension. These capabilities together support the study of visual representations in \Cref{sec:visual_representation_priors}.

\paragraph{Stream Backbones.}
The backbones in $\mathcal{S}$ are laid out in and around the central panel of \Cref{fig:openwam-infra}. The \textit{video backbone} ($\mathcal{W}$) predicts the temporal evolution of future world visual states, while the \textit{action backbone} ($\mathcal{A}$) predicts the upcoming action chunk~\citep{zhao2023learning}; together they form the world--action core of the model. The optional \textit{VLM backbone} ($\mathcal{U}$) supplements this core with semantic understanding of the current observation. Beyond these, the backbone roster is itself \textbf{expandable}: further backbones can be registered to execute any additional streams a design may introduce. Regardless of type, every backbone executes through one code contract that decomposes its forward pass into three stages:
\begin{itemize}[leftmargin=1.5em]
    \item \texttt{prepare}, invoked once before the stack, which embeds the
    inputs into the initial token state;
    \item \texttt{per-layer block step}, invoked once per layer, which
    advances this state through one transformer layer;
    \item \texttt{finalize}, invoked once after the stack, which maps the
    final state to the stream's prediction.
\end{itemize}
Each layer may further split its block step into a \texttt{pre-attention} half, which emits the layer's queries, keys, and values, and a \texttt{post-attention} half, which consumes the attention output, so that the attention between the two halves can be computed jointly across streams. \par For the video backbone, OpenWAM-Infra supports five pretrained video generation models with increasing model parameters, namely Wan2.1-VACE-1.3B, Cosmos-Predict2.5-2B, Cosmos3-Edge-4B, Wan2.2-TI2V-5B, and Wan2.1-I2V-14B~\citep{wan2025wan,ali2025world}. For VLM backbones, OpenWAM-Infra currently supports only the Qwen3-VL family~\citep{bai2025qwen3vltechnicalreport}. For the action backbone, OpenWAM-Infra offers two options: a separate set of parameters residing in ActionDiT, or a shared video backbone in which action tokens join the video token sequence and are processed jointly.

\paragraph{Visibility Attention Mask.}
The mask $\mathcal{M}$ governs the information flow with the mixed self-attention through which streams interact: it factorizes into intra-modality and cross-modality blocks, granting attention where tokens reinforce one another and withholding it where their mutual influence must be isolated. Over the video and action modalities, this factorization reads
\begin{equation}
\label{eq:mask-blocks}
\mathcal{M}=
\begin{pmatrix}
\mathcal{M}_{V\leftarrow V} & \mathcal{M}_{V\leftarrow A}\\
\mathcal{M}_{A\leftarrow V} & \mathcal{M}_{A\leftarrow A}
\end{pmatrix},
\end{equation}
where the block $\mathcal{M}_{X\leftarrow Y}$ specifies whether tokens of modality $X$ may attend to tokens of modality $Y$. OpenWAM-Infra fixes the two intra-modality blocks: $\mathcal{M}_{V\leftarrow V}$ adopts \emph{first-frame causal} attention, in which noisy frames attend to one another and to the clean first frame while the clean frame attends only to itself, shielding clean conditioning from noise; $\mathcal{M}_{A\leftarrow A}$ adopts \emph{bidirectional} attention, in which the noisy action tokens of a chunk are mutually visible so that the predicted actions inform one another. The two cross-modality blocks then define the four attention mask modes that OpenWAM-Infra supports, as drawn in the right panel of \Cref{fig:openwam-infra}: \emph{mutual} enables both $\mathcal{M}_{A\leftarrow V}$ and $\mathcal{M}_{V\leftarrow A}$, so the two modalities attend to each other; \emph{action-sees-video} enables only $\mathcal{M}_{A\leftarrow V}$, letting actions read the predicted world while leaving video generation undisturbed; \emph{video-sees-action} enables only $\mathcal{M}_{V\leftarrow A}$, the reverse; and \emph{isolated} disables both, denoising the two modalities independently. Building on this native support, \Cref{sec:training_time_flow} later compares these modes under controlled settings.

\paragraph{Architectures.}
The composition rule $C$ specifies where and how information crosses streams, and it does so purely through the execution contract above: it sequences the \textbf{prepare}, \textbf{per-layer block}, and \textbf{finalize} stages of the participating backbones and, when interaction must occur inside attention, splits the block step into its \textbf{pre-attention} and \textbf{post-attention} halves to substitute the attention computation itself, never modifying backbone internals. Composition rules therefore carry no parameters of their own; all learned capacity resides in the stream backbones. A concrete architecture is a choice $C(\mathcal{E},\mathcal{S},\mathcal{M})$; the designs currently supported fall into three families, laid out left to right in the bottom row of \Cref{fig:openwam-infra}.
\begin{itemize}[leftmargin=1.5em]
    \item \textbf{Single-System.} This family comprises only the video
    backbone: the action backbone takes the shared form and injects its tokens
    into the video sequence, so one transformer processes both modalities with
    most parameters shared; representative systems include Cosmos
    Policy~\citep{kim2026cosmos} and DreamZero~\citep{ye2026world}.
    OpenWAM-Infra provides two variants, differing in modality-specific
    capacity:
    \begin{itemize}[leftmargin=1.5em]
        \item \emph{Vanilla} processes video, action, and proprioceptive tokens
        as one sequence through the same attention and dense feed-forward
        blocks, providing no modality-specific capacity.
        \item \emph{MoE} retains the shared sequence and self-attention but
        hard-routes action tokens to a dedicated feed-forward expert while
        video tokens follow the default path, adding modality-specific
        capacity without separating the streams~\citep{mu2025comprehensive}.
    \end{itemize}
    \item \textbf{Dual-System.} This family comprises an independent video
    backbone and action backbone, the latter a dedicated ActionDiT: the two
    streams hold separate sets of parameters and are connected through self-
    or cross-attention; representative systems include
    Fast-WAM~\citep{yuan2026fast} and LingBot-VA~\citep{li2026causal}.
    OpenWAM-Infra provides three variants, differing in how the two streams
    communicate:
    \begin{itemize}[leftmargin=1.5em]
        \item \emph{Joint self-attention} merges the hidden states of the two
        streams into a joint sequence at designated bridge layers, allowing
        bidirectional token-level interaction before the states return to
        their streams.
        \item \emph{Joint cross-attention} lets the action stream query video
        features through video-to-action cross-attention at the bridge layers,
        trained end-to-end so that action-learning gradients update the video
        stream; optionally, gradients are detached at the video features to
        isolate action learning from video parameter updates.
        \item \emph{IDM} formulates the action module as an inverse-dynamics
        model conditioned on video features, trained with teacher-forced video
        states and run in two inference stages: the video trajectory is
        generated first and the actions are predicted from it.
    \end{itemize}
    \item \textbf{Tri-System.} This family extends the dual layout with a VLM
    backbone, in which a frozen vision--language model feeds a separate
    trainable understanding stream; representative systems include
    Motus~\citep{bi2026motus}.
    OpenWAM-Infra provides a single variant:
    \begin{itemize}[leftmargin=1.5em]
        \item \emph{Joint self-attention} extends the joint sequence to all
        three streams, which exchange information while retaining
        stream-specific parameters; the understanding stream joins as a
        read-only tail that the other streams may attend to while it attends
        only to itself.
    \end{itemize}
\end{itemize}
Within each architecture $C(\mathcal{E},\mathcal{S},\mathcal{M})$, every module (the visual encoder, the stream backbones, and the visibility attention mask) is instantiated from a registry, orthogonally to the composition rule: every combination is assembled from configuration, and neither the trainer nor the policy server is aware of which architecture is running.

\subsection{Training Runtime}
\label{sec:infra_train}

\paragraph{Training Formulation.}
OpenWAM-Infra trains every architecture under one trainer, against one sample contract and one joint flow-matching objective. The trainer never inspects architecture internals: it asks the selected architecture to prepare its own inputs and run its own forward pass, then optimizes the objective on the resulting predictions. Define a sample as $(\ell,\,\mathbf{o}_{1:T},\,\mathbf{a}_{1:H},\,\mathbf{q},\,\mathbf{m})$, a language instruction, a video window, an action chunk, an optional proprioceptive state, and a per-dimension validity mask. During input preparation, the architecture's visual encoder $\mathcal{E}$ encodes $\mathbf{o}_{1:T}$ into latents $\mathbf{z}$, and $\ell$, optionally joined by $\mathbf{q}$, becomes the context $\mathbf{c}$. Throughout the paper, $t=0$ denotes pure noise and $t=1$ clean data. Each stream is noised to its own timestep, $t_v$ for video and $t_a$ for actions, yielding the interpolants $\mathbf{z}^{t_v}=t_v\,\mathbf{z}+(1-t_v)\,\boldsymbol{\epsilon}_v$ and $\mathbf{a}^{t_a}=t_a\,\mathbf{a}+(1-t_a)\,\boldsymbol{\epsilon}_a$ with Gaussian noise $\boldsymbol{\epsilon}_v,\boldsymbol{\epsilon}_a$; one joint forward pass of the architecture $(\hat{\mathbf{v}}_z,\hat{\mathbf{v}}_a)=\mathbf{v}_\theta\big(\mathbf{z}^{t_v},\mathbf{a}^{t_a},t_v,t_a,\mathbf{c}\big)$ predicts both velocities, and the objective takes the form
\begin{equation}
\label{eq:wam-loss}
\mathcal{L}
=\lambda_v\,\mathbb{E}_{t_v,\epsilon_v}\!\Big[w(t_v)\,\big\lVert \hat{\mathbf{v}}_z-(\mathbf{z}-\boldsymbol{\epsilon}_v)\big\rVert_2^2\Big]
+\lambda_a\,\mathbb{E}_{t_a,\epsilon_a}\!\Big[w(t_a)\,\big\lVert \mathbf{m}\odot\big(\hat{\mathbf{v}}_a-(\mathbf{a}-\boldsymbol{\epsilon}_a)\big)\big\rVert_2^2\Big],
\end{equation}
where $\lambda_v,\lambda_a$ and $w(\cdot)$ weight the streams and the timesteps, while the validity mask $\mathbf{m}$ restricts the action term to the coordinates an embodiment actually populates, and clean conditioning frames are excluded from the video term. Because $t_v$ and $t_a$ are sampled \emph{independently}, training covers the entire $(t_v,t_a)$ noise plane; any inference schedule, whether it denoises the two streams synchronously at a shared timestep or asynchronously with one stream leading the other, traces a path through this plane and thus remains in-distribution.

\paragraph{Training Utilities.}
As shown in the Training Utils panel at the center of \Cref{fig:openwam-infra}, three core utilities support OpenWAM-Infra training:
\begin{enumerate}[leftmargin=1.5em]
    \item \textbf{Framework.} OpenWAM-Infra integrates DeepSpeed ZeRO (stage 1
    or 2) through Accelerate and supports mixed precision (bf16 by default),
    gradient accumulation, and gradient clipping; a single entry point scales from
    single-GPU runs to multi-node jobs.
    \item \textbf{Memory optimization.} To reduce memory consumption,
    OpenWAM-Infra provides gradient checkpointing, with optional CPU offload
    of the checkpointed activations, and optimizer-state offload to CPU.
    \item \textbf{Workflows.} OpenWAM-Infra supports three training workflows.
    \emph{Pretraining} starts a fresh run. \emph{Fine-tuning} starts a new
    run initialized from a previous checkpoint: the architecture is rebuilt
    from the checkpoint's own record, the new configuration is layered on top,
    and the identity of the modules the weights belong to is protected from
    override. \emph{Resume} continues the same run exactly: the full optimizer
    and scheduler state is restored,
    training re-enters the data stream at the recorded position, and the run
    refuses to continue if the dataset's normalization statistics diverge from
    those recorded with the run.
\end{enumerate}

\paragraph{Self-Contained Checkpoints.}
A self-contained checkpoint comprises three parts: the model weights, the fully resolved configuration with every module's reconstruction specification (and artifacts such as tokenizers) merged in, and the action-normalization statistics. Fine-tuning, resume, and deployment all rebuild the architecture from this record before loading parameters; at deployment, a missing normalization record is a hard error. An evaluation therefore cannot silently change the encoder, the action layout, or the scaling of a trained model, and the checkpoint is exactly what the deployment runtime serves.

\subsection{Deployment Runtime}
\label{sec:infra_deploy}

\begin{figure}[htbp]
    \centering
    \includegraphics[width=0.8\linewidth]{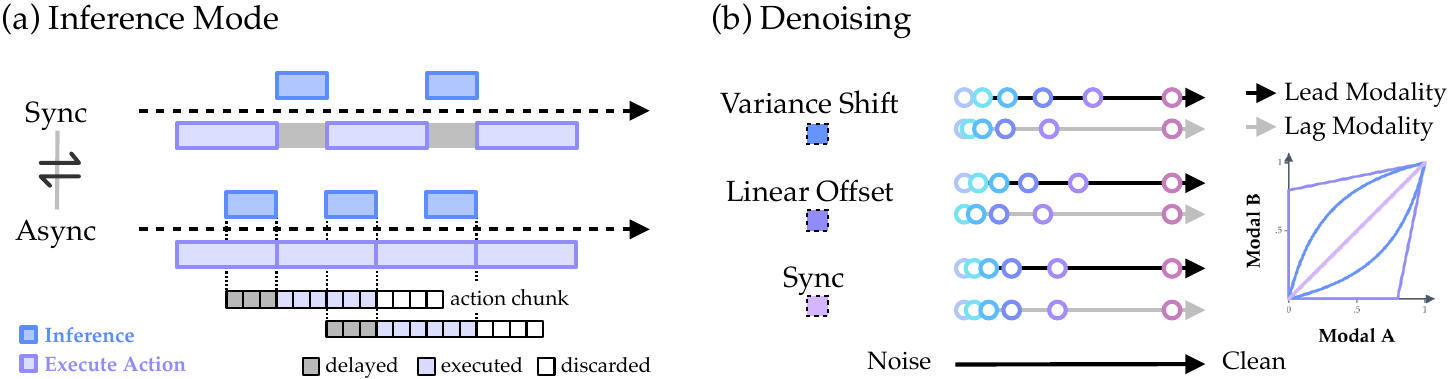}
    \caption{\textbf{Inference modes and denoising schedules of the deployment runtime.} (a) Illustration of the synchronous and asynchronous inference modes. (b) Illustration of the three denoising schedules (variance shift, linear offset, and sync); five denoising steps are drawn for illustration, and circles of the same color denote the timesteps that the two modalities reach at the same denoising step.}
    \label{fig:inference-denoise}
\end{figure}

Every checkpoint is served by one policy server, which rebuilds the architecture from its self-contained record and keeps two choices orthogonal: when inference runs (the inference mode) and how the two streams are denoised (the denoising schedule). \Cref{fig:inference-denoise} illustrates the two choices in panels (a) and (b), respectively.

\paragraph{Inference Modes.}
OpenWAM-Infra provides two inference modes over a common buffer mechanism (\Cref{fig:inference-denoise}(a)): each inference produces an action chunk that is buffered, and the server pops one action per request. Under \emph{synchronous} inference, the server blocks on a fresh inference whenever the buffer empties, so execution stalls for the inference latency. Under \emph{asynchronous} inference, let $H$ denote the length of the predicted chunk, $n\le H$ the inference horizon, and $d<n$ the lead threshold in steps, defaulting to $n/2$. Once only $d$ buffered actions remain, a single background worker prefetches the next chunk while those actions keep executing. The adopted chunk then splits, in order, into a \emph{delayed} prefix of $d$ actions, already covered by the previous buffer while inference ran and therefore skipped; an \emph{executed} window of the next $n$ actions, which becomes the new buffer; and a \emph{discarded} tail of the remaining $\max\{H-d-n,\,0\}$ actions. The two modes are indistinguishable to the client: every request is one observation in, one action out.

\paragraph{Denoising Schedules.}
A denoising schedule is a path $\tau=\{(t_v^i,t_a^i)\}_{i=0}^{N}$ through the joint noise plane, with $t=0$ pure noise and $t=1$ clean data as in \Cref{sec:infra_train}; \Cref{fig:inference-denoise}(b) draws the three schedules that OpenWAM-Infra supports. Under the \emph{sync} schedule, both streams advance in lockstep along the diagonal: each step performs one joint forward pass and a coupled Euler update in which each stream moves by its own timestep increment; the video latents $\mathbf{z}$ follow $\mathbf{z}^{t_v^{i+1}}=\mathbf{z}^{t_v^i}+(t_v^{i+1}-t_v^i)\,\hat{\mathbf{v}}_z$, and the action chunk $\mathbf{a}$ follows $\mathbf{a}^{t_a^{i+1}}=\mathbf{a}^{t_a^i}+(t_a^{i+1}-t_a^i)\,\hat{\mathbf{v}}_a$. Asynchronous schedules let one stream lead through two composable families~\citep{baade2026latent},
\begin{equation}
\label{eq:schedule-families}
f_{\alpha}(s)=\frac{\alpha s}{1+(\alpha-1)s},
\qquad
h_{o}(s)=\max\!\left\{\frac{s-o}{1-o},\,0\right\},
\end{equation}
where $s=i/N$ denotes global progress, the \emph{variance shift} curve $f_{\alpha}$ lifts the leading stream above the diagonal for $\alpha>1$ so that it reaches clean data earlier, and the \emph{linear offset} $h_{o}$ holds the lagging stream at pure noise until global progress exceeds $o$. Assigning the lead to the world stream or to the action stream yields the \emph{video-lead} and \emph{action-lead} regimes, and $(\alpha,o)=(1,0)$ recovers the synchronized diagonal exactly: synchronous serving is a special case rather than a separate code path, and every asynchronous run has an aligned baseline. Because training samples the two timesteps independently (\Cref{sec:infra_train}), every such path stays in-distribution. Independently of the schedule shape, each stream's timestep warp is a backbone property stored in the checkpoint and reused at inference, so the training and serving noise grids cannot drift.

\paragraph{Acceleration.}
OpenWAM-Infra provides four serving-side accelerations, each independently configurable.
{\setlength{\intextsep}{0pt}
\begin{wrapfigure}[13]{r}{0.58\textwidth}
    \centering
    \includegraphics[width=\linewidth]{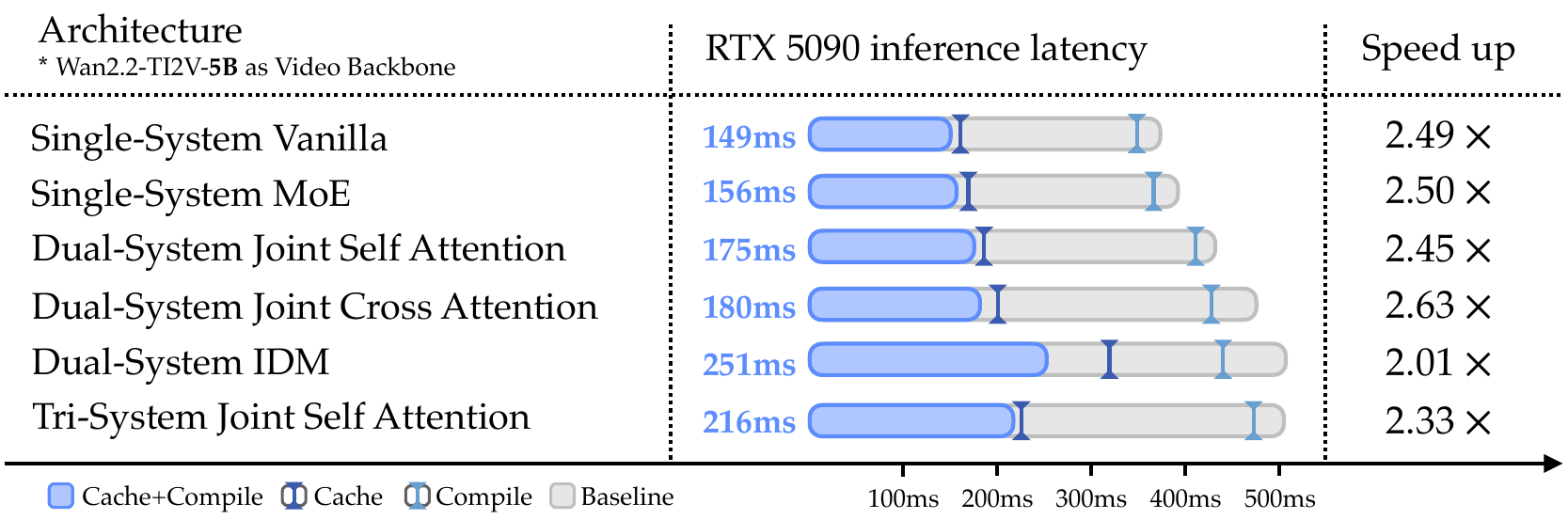}
    \caption{\textbf{Serving latency across architectures.} Inference latency with Wan2.2-TI2V-5B as the video backbone on an RTX 5090. The prompt-embedding cache and video-decode skip are enabled by default; the figure ablates compilation and the DiT velocity cache.}
    \label{fig:acceleration}
\end{wrapfigure}
\setlength{\parskip}{1.5pt}

\vspace{-10pt}
\noindent
\textbullet~\textbf{Prompt-embedding cache}: a server-lifetime cache maps each
prompt to its text-encoder embeddings, removing the text encoder from the
per-request path.

\noindent
\textbullet~\textbf{Video-decode skip}: control consumes actions rather than
pixels, so the serving path can skip VAE video decoding entirely.

\noindent
\textbullet~\textbf{Compilation}: each architecture registers a fixed-shape
\texttt{torch.compile} path for its inner joint denoising loop, replayed
under CUDA graphs to eliminate per-layer launch overhead; the first
request carries the compilation warmup.

\noindent
\textbullet~\textbf{DiT velocity cache}: when the recent velocity predictions of
\emph{both} streams are similarity-stable (cosine similarity above a
threshold), the next joint forward pass is skipped and the cached
velocities are integrated instead, with a bounded number of consecutive
skips, following the cross-step reuse of \citet{ye2026world}.\par}

\noindent
With Wan2.2-TI2V-5B as the video backbone, \Cref{fig:acceleration} illustrates the resulting inference speedups across the different architectures.

\subsection{Evaluation Protocol}
\label{sec:eval-protocol}

\begin{figure}[htbp]
    \centering
    \includegraphics[width=\linewidth]{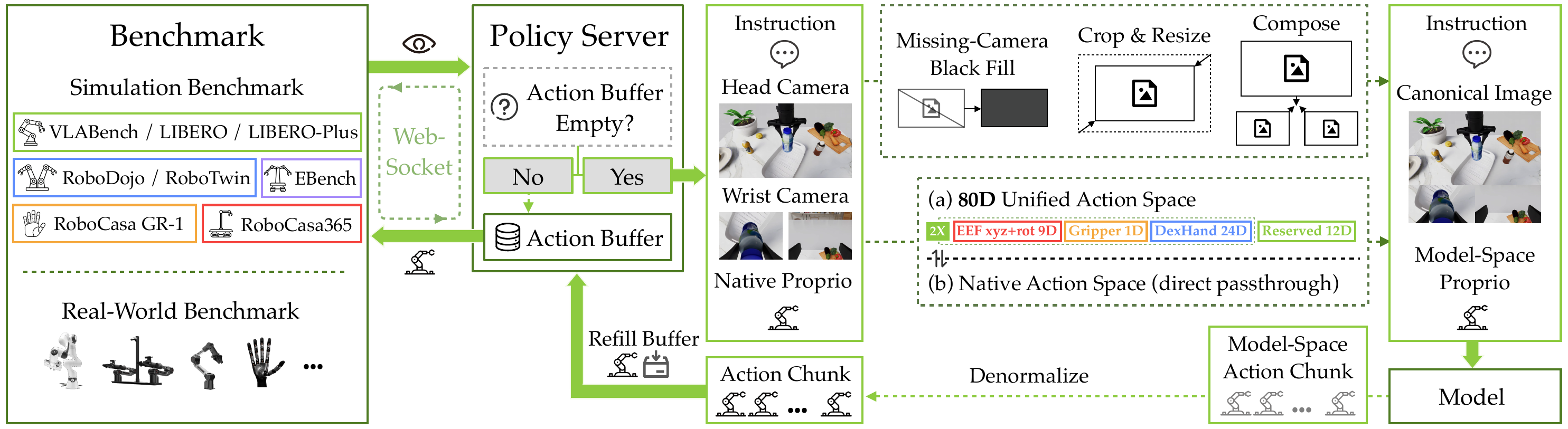}
    \caption{\textbf{Evaluation protocol of OpenWAM-Infra.} Benchmarks connect to the policy server as thin clients over WebSocket. The server canonicalizes each observation, maps the proprioceptive state into the model-side action representation (the 80-D unified action space or the benchmark's native action space), and denormalizes the predicted action chunk back to native physical units before returning actions.}
    \label{fig:eval-protocol}
\end{figure}

OpenWAM-Infra evaluates trained checkpoints through the policy server of \Cref{sec:infra_deploy}: each benchmark connects as a client, sends observations, and executes the actions returned by the server, as shown in \Cref{fig:eval-protocol}.

\paragraph{Client--Server Pipeline.}
Benchmarks reach the deployment runtime as thin clients over one persistent WebSocket connection and import nothing from the model or training stack. At control step $k$, the client sends an observation $(\mathbf{o}_k,\,\ell,\,\mathbf{q}_k)$: up to three camera views $\mathbf{o}_k$, with the head view required and the wrist views optional; the language instruction $\ell$; and optionally the raw robot state $\mathbf{q}_k$. The response is a single action $\mathbf{a}_k$ in the robot's native physical units. Every model-facing conversion runs server-side, driven by the self-contained checkpoint of \Cref{sec:infra_train}: as laid out in \Cref{fig:eval-protocol}, the views are cropped, resized, and composed into the canonical image layout the checkpoint was trained on, with missing cameras filled by black frames, and $\mathbf{q}_k$ is normalized and mapped into the model-side action representation defined below. Inference under the serving stack of \Cref{sec:infra_deploy} then yields a model-space action chunk $\hat{\mathbf{a}}_{1:H}$, which is mapped back and denormalized into native units before it refills the action buffer from which the server answers requests; normalized values therefore never reach a robot. Between episodes, a single reset request clears all per-episode executor state.

\paragraph{Benchmark Suite.}
OpenWAM-Infra currently integrates eight simulation benchmarks: LIBERO and LIBERO-Plus~\citep{liu2023libero,fei25libero-plus}, VLABench~\citep{zhang2024vlabench}, RoboTwin2.0~\citep{chen2025robotwin}, RoboDojo~\citep{chen2026robodojounifiedsimandrealbenchmark}, RoboCasa365~\citep{robocasa365}, RoboCasa-GR1~\citep{nvidia2025gr00tn1,robocasa2024}, and EBench~\citep{gao2026ebench}, together spanning single-arm and bimanual tabletop manipulation, dexterous-hand humanoid control, and mobile manipulation. Because the protocol exchanges only images, text, and action vectors, real-robot platforms connect through exactly the same interface as the simulators. Each bundled adapter reproduces the observation preprocessing of its benchmark's training reader, so evaluation-time views match the training distribution; integrating a new benchmark amounts to writing such an adapter, leaving the model and both runtimes untouched.

\paragraph{Action Space Definition.}
Actions cross this pipeline in one of two representations (\Cref{fig:eval-protocol}). By default, every benchmark keeps its \emph{native} action space --- RoboTwin2.0, for instance, is served in either a 14-D joint space or a 20-D bimanual end-effector space, and LIBERO in a 10-D end-effector space --- so a checkpoint trained on a single benchmark passes actions straight through. Training one model across embodiments, however, requires a single action head over bodies whose native layouts differ in both width and semantics. OpenWAM-Infra therefore also defines a \emph{unified action space} $\mathbf{u}\in\mathbb{R}^{80}$ with fixed slot semantics: two mirrored 34-D arm blocks, each comprising the end-effector position (3), a 6D rotation (6), the gripper (1), and a dexterous hand (24), followed by 12 reserved slots for embodiment-specific channels such as the mobile bases of EBench and RoboCasa365. Since the slot semantics are fixed, the structure that embodiments share lands on the same coordinates. Each dataset declares an index map $\pi$ from its native dimensions into these slots, with normalization applied \emph{before} scattering and inverted \emph{after} gathering,
\begin{equation}
\label{eq:unify-map}
\mathbf{u}=\mathrm{Scatter}_{\pi}\big(\mathrm{Norm}(\mathbf{a})\big),
\qquad
\mathbf{a}=\mathrm{Norm}^{-1}\big(\mathrm{Gather}_{\pi}(\mathbf{u})\big),
\end{equation}
and incoming proprioception traverses the same map in the forward direction. The validity mask $\mathbf{m}$ of \Cref{eq:wam-loss} marks exactly the mapped slots, so unmapped coordinates receive no gradient during training and remain on their analytic noise path at inference.

\clearpage

\section{OpenWAM-Study: Design Principles for World--Action Models}
\label{sec:recipe}

\paragraph{Overview of OpenWAM-Study.} Building on the substrate of \textbf{OpenWAM-Infra}, we systematically analyze design principles for world--action models through controlled experiments. In this section, we first study how WAMs should inherit upstream world priors in Section \ref{sec:inherit_world_prior}, then understand how to build synergy between the world priors and action learning in Section \ref{sec:world_action_synergy}, and finally test which design choices generalize to cross-domain embodied pretraining in Section \ref{sec:embodied_pt_study}. 


\paragraph{Evaluation Protocol.} In this section, we use \textbf{RoboTwin2.0}~\citep{chen2025robotwin}, a widely-adopted bi-manual manipulation benchmark spanning over 50 tasks, as our evaluation environment. In our experiments, we evaluate under two settings: (1) \textbf{In-Domain Performance (RoboTwin2.0-Full)}: Following \citet{bi2026motus}, we train our model with an entire multi-task data corpus of 2,500 demonstrations collected in clean scenes
and 25,000 demonstrations collected under heavy scene randomization, and evaluate under clean and randomized environments separately. (2) \textbf{Out-of-Domain Generalization (RoboTwin2.0-Clean2Random)}: Following \citet{yuan2026qwen}, we train our model on clean data only and evaluate under clean and randomized environments separately. Since the training mixture has never seen randomized scene configurations, it serves as a proxy for evaluating the models' generalization capabilities to novel scenes. We use success rate as our metric.

\subsection{Inheriting Upstream World Knowledge}
\label{sec:inherit_world_prior}

In this section, we focus on the following question:

\question{1}{What world knowledge should WAMs inherit, and how is it best inherited?}

World knowledge can be inherited largely through two channels: generative world priors and visual representation priors. Generative world priors refer to the visual and dynamical knowledge embedded in the video generation backbone, whereas visual representation priors refer to the representation space induced by vision encoders \citep{radford2021learning,caron2021emerging,wan2025wan}.

\subsubsection{Generative World Priors}

\begin{wrapfigure}{r}{0.39\textwidth}
    \vspace{-10pt}
    \centering
    \includegraphics[width=0.98\linewidth]{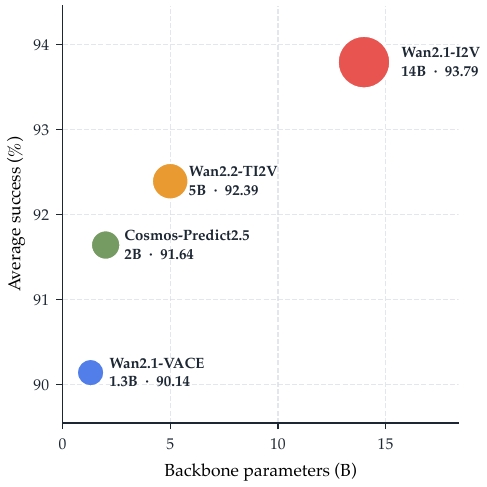}
    \caption{\textbf{WAM Performance with Different Video Backbone Size.} With increasing video generation backbone size, performance of the resulting WAM consistently improves.}
    \label{fig:generative-backbones}
    \vspace{-12pt}
\end{wrapfigure}

To understand whether and to what extent generative world priors facilitate WAM performance, we compare four video generation backbones with variable parameter counts: Wan2.1-VACE-1.3B \citep{wan2025wan}, Cosmos-Predict2.5-2B \citep{ali2025world}, Wan2.2-TI2V-5B \citep{wan2025wan}, and Wan2.1-I2V-14B \citep{wan2025wan}. We instantiate our World--Action Model with a Dual-System architecture consisting of a video generation module and an action generation module connected with joint self-attention, in which the video generation module parameters are copied from the pretrained video generation model. Results are evaluated on RoboTwin2.0-Full.

Across the four tested backbones, the average success rate of the resulting WAM improves consistently with video generation backbones of increasing capacity (\Cref{fig:generative-backbones}). Wan2.1-I2V-14B performs best, while Wan2.2-TI2V-5B trails it by only 1.40 points, even though the former has nearly \textbf{3x} the parameter count. Balancing performance against training and deployment efficiency across the four backbones, we ultimately adopt the 5B model as the default for the remaining controlled studies. Since these backbones also differ in architecture, pretraining data, and objective, this comparison establishes backbone choice as a consequential design variable without attributing the entire gain to parameter count alone. The 5B default also keeps subsequent action-side ablations tractable while holding the inherited world prior fixed.

\subsubsection{Visual Representation Priors}
\label{sec:visual_representation_priors}

\begin{figure}[t]
    \centering
    \includegraphics[width=0.90\linewidth]{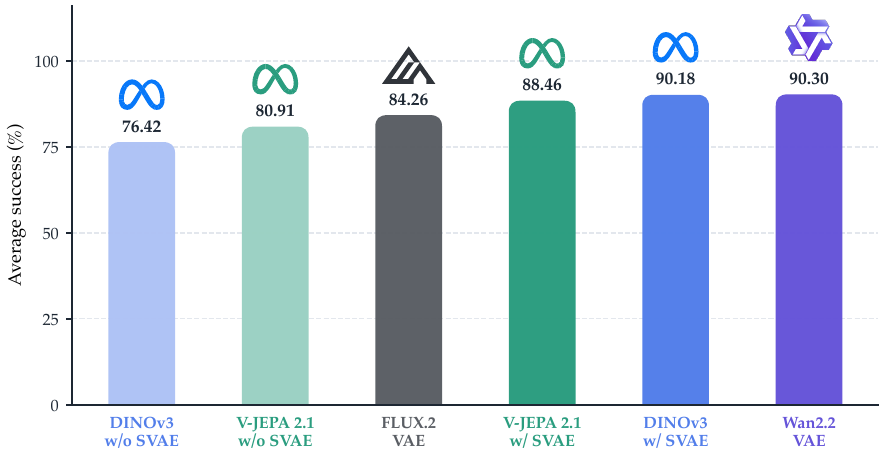}
    \caption{\textbf{Visual representation priors.} We consider building WAMs with both reconstructive and representation encoders, and include a variant of representation encoders with S-VAE \citep{zhang2025both}, an adapter that converts high-dimensional latents produced by representation encoders into low-dimensional vectors suitable for DiT processing.}
    \label{fig:visual-representations}
\end{figure}

Another important source of world knowledge comes from the latent representation space induced by vision encoders. Broadly speaking, these encoders fall into two categories: (1) \textbf{Reconstructive Encoders}: the objective of these encoders is compression, trained solely with pixel-reconstruction; (2) \textbf{Representation Encoders}: grounded by self-supervised or multimodal representation learning, these encoders learn semantically structured visual features that provide a basis for visual understanding. Motivated by recent advances in generative and world modeling with representation encoders \citep{zheng2026diffusion, singh2026improved, zhou2024dino, jha2026reconstruction,lyu2026lda}, we question the design of latent space in the context of world--action models.

In this set of experiments, we choose representative encoders from both categories. For Reconstructive Encoders, we use Wan2.2-VAE \citep{wan2025wan}, a state-of-the-art video encoder with a 4x temporal compression rate, and FLUX.2-VAE \citep{bfl2025flux2}, an advanced image encoder yielding highly performant image generation models built on its latent space. For Representation Encoders, we use DINOv3 \citep{simeoni2025dinov3}, the newest generation of DINO \citep{caron2021emerging}, a classical vision encoder learned through self-supervised learning; and V-JEPA 2.1 \citep{murlabadia2026vjepa2_1}, a dense feature encoder based on joint embedding predictive architectures \citep{lecun2022path}. To isolate the performance gain from visual representations alone, we inherit the model architecture of Wan2.2-TI2V-5B, but randomly initialize its model weights. For a fair comparison, we apply the same 4x temporal compression as Wan2.2-VAE to FLUX.2-VAE, DINOv3, and V-JEPA 2.1: since none of these encoders natively performs temporal compression, we impose the 4x downsampling by averaging the features of every four consecutive frames.

Since modern Diffusion Transformer architectures \citep{peebles2023scalable} are optimized mostly for reconstructive encoders, naive adoption of representation encoders, which produces high-dimensional latents (e.g., 768-D features for DINOv3 and 1024-D for V-JEPA 2.1), can lead to poor performance due to architectural incompatibility \citep{zheng2026diffusion}. Following \citet{jha2026reconstruction}, we include a variant for representation encoders, where we train an S-VAE \citep{zhang2025both} adapter that converts the high-dimensional features produced by representation encoders to lower-dimensional vectors (48-D in this experiment, matching the latent dimension of Wan2.2-VAE). 

As shown in \Cref{fig:visual-representations}, representation encoders can yield performance on par with or stronger than reconstructive encoders for world--action modeling with the help of S-VAEs. Specifically, while naive adoption of representation encoders yields worse performance than models trained with the reconstructive encoder FLUX.2-VAE, with dimension contraction using S-VAE, the representation encoders' contracted variants (DINOv3 w/ SVAE, V-JEPA 2.1 w/ SVAE) significantly outperform FLUX.2-VAE. DINOv3 w/ SVAE achieves nearly on-par performance with Wan2.2-VAE, and we attribute the remaining slim margin to two native advantages of Wan2.2-VAE: it is a reconstructive encoder specifically suited to the video backbone architecture, and its temporal compression is learned natively by the encoder rather than imposed through frame averaging.

Taken together, what determines the quality of a WAM latent space is not the categorical divide between reconstructive and representation encoders, but the operational properties of the latents themselves: compactness (in both the temporal and the token dimension) and rich world information. Priors in representation encoders can thus be inherited to build highly performant WAMs with the help of \textbf{temporal compression} and \textbf{dimension contraction}. We also encourage active research into building representation encoders with native temporal compression, which in turn may lead to even better prior inheritance.

\finding{1}{A WAM inherits upstream world knowledge most effectively through a sufficiently capable generative backbone and a compact, information-rich visual representation space. Reconstructive encoders are not the only option; representation encoders with dimension compression are also performant.}

\FloatBarrier

\subsection{Building Synergy between World and Action Learning}
\label{sec:world_action_synergy}
Inheriting the right priors is not enough; a world--action model needs to build synergy between world and action learning. This requires three decisions at different levels of the system: where action-specific capacity lives, which cross-modal information paths are available during training, and whether inference preserves the noise-state relationship on which those paths were learned.

\question{2}{How should inherited world knowledge interact with action learning?}

\subsubsection{Architectural Capacity}
\label{sec:architectural_capacity}

\begin{wraptable}{r}{0.64\textwidth}
    \vspace{-12pt}
    \centering
    \small
    \setlength{\tabcolsep}{4pt}
    \captionof{table}{\textbf{Architecture Ablation.} Averaged success rates (\%) on RoboTwin2.0-Full. Bold denotes best values.}
    \label{tab:architecture-ablation}
    \begin{tabular}{llccc}
        \toprule
        \multicolumn{2}{c}{Architecture} & \multicolumn{3}{c}{Success Rate (\%)} \\
        \cmidrule(lr){1-2}\cmidrule(lr){3-5}
        \multicolumn{1}{c}{System} & \multicolumn{1}{c}{Variant} & Clean & Randomized & Average \\
        \midrule
        \multirow{2}{*}{\textbf{Single-System}}
            & \cellcolor{black!4}\textbf{Vanilla}
            & \cellcolor{black!4}85.20
            & \cellcolor{black!4}85.80
            & \cellcolor{black!4}85.50 \\
            & \textbf{MoE} & 86.22 & 83.04 & 84.63 \\
        \cmidrule(lr){1-5}
        \multirow{4}{*}{\textbf{Dual-System}}
            & \cellcolor{black!4}\textbf{Joint Self-Attention}
            & \cellcolor{black!4}92.34
            & \cellcolor{black!4}\textbf{92.38}
            & \cellcolor{black!4}92.36 \\
            & \textbf{Joint Cross-Attention} & 87.86 & 88.64 & 88.25 \\
            & \cellcolor{black!4}\textbf{Detached Cross-Attention}
            & \cellcolor{black!4}92.06
            & \cellcolor{black!4}91.64
            & \cellcolor{black!4}91.85 \\
            & \textbf{IDM} & 87.76 & 88.14 & 87.95 \\
        \cmidrule(lr){1-5}
        \textbf{Tri-System}
            & \cellcolor{black!4}\textbf{Joint Self-Attention}
            & \cellcolor{black!4}\textbf{92.84}
            & \cellcolor{black!4}92.36
            & \cellcolor{black!4}\textbf{92.60} \\
        \bottomrule
    \end{tabular}
    \vspace{-8pt}
\end{wraptable}

A central question in world--action modeling is how much action-specific capacity a WAM requires and how strongly its video and action streams should be separated. The three architecture families of \Cref{sec:infra_model} span precisely this capacity axis, and we evaluate all six of their variants, instantiating joint cross-attention both end-to-end and with gradients detached at the video features, yielding seven baselines (\Cref{tab:architecture-ablation}).

\paragraph{Results.} As shown in \Cref{tab:architecture-ablation}, with increasing architecture capacity, performance from single- to dual- and tri-system continuously improves. Joint self-attention is the strongest dual-system variant, while the tri-system model achieves the best overall performance. Balancing performance with architectural complexity, and isolating the interaction between world knowledge and action learning from the potential influence of the VLM's understanding features, we therefore adopt dual-system joint self-attention for the remaining experiments, so that the subsequent findings reflect this interaction alone.

\subsubsection{Training-Time Information Flow}
\label{sec:training_time_flow}

The architectural comparison selects joint self-attention as the interface between the world and action streams, but joint attention alone does not specify which information flow creates the best synergy. We compare four information flow strategies at training time, controlled by attention masking: \textbf{Isolated}, with no cross-stream communication; \textbf{Video Sees Action}, which exposes action features to the world stream; \textbf{Action Sees Video}, which exposes world features to the action stream; and \textbf{Mutual}, which enables both directions.

\begin{center}
    \centering
    \begin{minipage}[t]{0.64\linewidth}
        \vspace{-3pt}
        \centering
        \includegraphics[width=\linewidth]{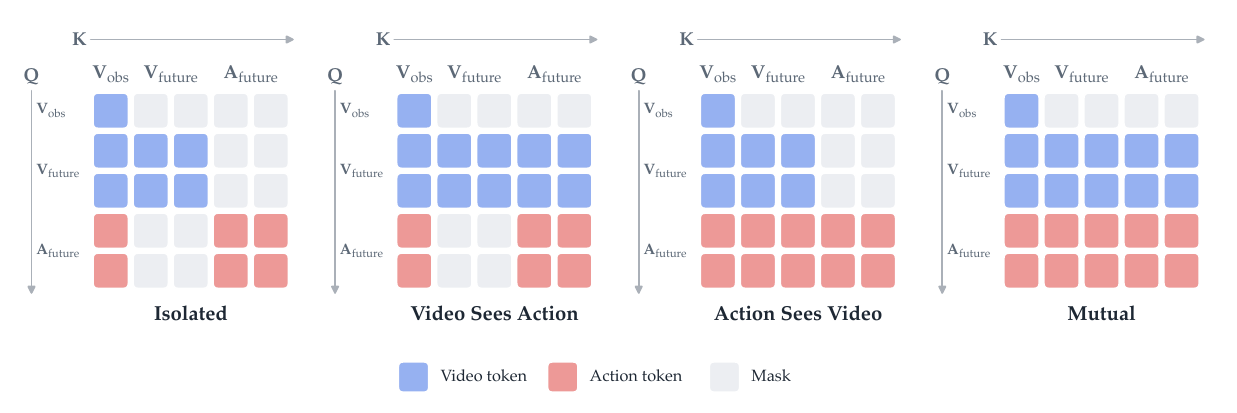}
        \vspace{-20pt}
        \captionof{figure}
        {\textbf{Attention Masking Strategies.} We control cross-modality information flow at training time via attention masking.}
        \label{fig:training-information-flow}
    \end{minipage}%
    \hfill
    \begin{minipage}[t]{0.34\linewidth}
        \vspace{0pt}
        \centering
        \captionof{table}{\textbf{Action learning requires access to world information.} Success rates (\%) on RoboTwin2.0-Full.}
        \label{tab:training-information-flow}
        \vspace{3pt}
        \scriptsize
        \setlength{\tabcolsep}{3.2pt}
        \renewcommand{\arraystretch}{1.16}
        \begin{tabular}{lccc}
            \toprule
            \multicolumn{1}{c}{\textbf{Mask}} & \multicolumn{3}{c}{\textbf{Success Rate (\%)}} \\
            \cmidrule(lr){1-1}\cmidrule(lr){2-4}
            & Clean & Random. & Average \\
            \midrule
            \rowcolor{black!4}
            \textbf{Isolated} & 88.08 & 86.74 & 87.41 \\
            \textbf{Video Sees Action} & 87.92 & 87.34 & 87.63 \\
            \rowcolor{black!4}
            \textbf{Action Sees Video} & \textbf{92.98} & \textbf{91.80} & \textbf{92.39} \\
            \textbf{Mutual} & 92.50 & \textbf{91.80} & 92.15 \\
            \bottomrule
        \end{tabular}
    \end{minipage}
\end{center}

The comparison separates cleanly according to whether the action stream can access world features (\Cref{fig:training-information-flow,tab:training-information-flow}). Isolated and video-sees-action masks underperform by roughly five points, whereas action-sees-video and mutual visibility both retain strong performance. World-to-action flow is therefore necessary in this setting. By contrast, adding the reverse action-to-world path changes the from-scratch result only marginally, leaving action-sees-video and mutual visibility as two viable masks to revisit after pretraining.

\subsubsection{Inference-Time Information Flow}
\label{sec:inference_time_flow}

\begin{wrapfigure}{r}{0.65\textwidth}
    \vspace{-24pt}
    \centering
    \includegraphics[width=0.99\linewidth]{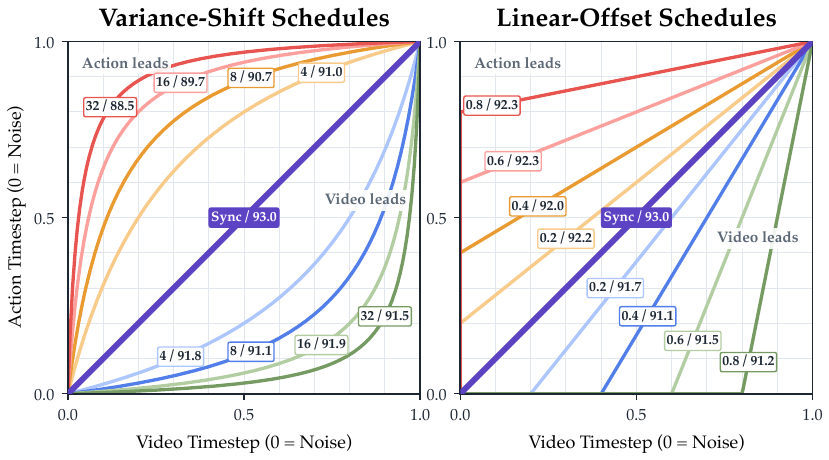}
    \vspace{-9pt}
    \caption{\textbf{Inference-time Information Flow via Denoising Schedule.} Each curve traces action denoising progress against video denoising progress. Curves above/below the diagonal denoise action/video first, respectively.}
    \label{fig:denoising-schedules}
    \vspace{-10pt}
\end{wrapfigure}

Training-time attention masking implements \textit{full masking}, while the inference-time denoising schedule provides a softer information gate, i.e. \textit{partial masking}. For this ablation, we fix mutual visibility and train the video and action streams with independently sampled noise levels, covering a two-dimensional space of joint noise states.

We instantiate the schedule abstraction of \Cref{sec:infra_deploy}: synchronized denoising follows the diagonal $(t_v^i,t_a^i)=(s_i,s_i)$, while the variance-shift and linear-offset families of \Cref{eq:schedule-families} let either stream lead~\citep{baade2026latent}. We evaluate $\alpha\in\{4,8,16,32\}$ and $o\in\{0.2,0.4,0.6,0.8\}$ in both leading directions.

\paragraph{Results.} Synchronized denoising performs best, and no asynchronous schedule improves performance, regardless of which stream leads or how relative progress is parameterized (\Cref{fig:denoising-schedules}). With variance-shift schedules, video-leading outperforms action-leading schedules, whereas with linear-offset schedules, action-leading outperforms video-leading schedules. This suggests that while explicit video-to-action information flow is necessary at training time, enforcing such priors through the inference-time denoising schedule does not yield gains, supporting the representation learning hypothesis of world--action modeling \citep{yuan2026fast} as opposed to an implicit planning-then-IDM schedule at test time \citep{ye2026world}.

\finding{2}{World--action synergy requires explicit world-to-action information flow during training, and synchronized joint denoising at inference. We carry forward dual-system joint self-attention with synchronized denoising and defer the close choice between one-way and mutual visibility to pretraining.}

\FloatBarrier

\subsection{Consolidating Knowledge Across Domains}
\label{sec:embodied_pt_study}
The preceding studies identify which world priors to inherit and how world and action streams should interact. However, strong single-domain performance alone does not establish transferable world--action knowledge: the model may simply fit the visual and action distribution of the target tasks, a distinction that cross-domain pretraining sharpens. Robot trajectories provide executable action supervision but limited visual coverage, whereas egocentric video offers broader visual diversity but lacks robot action labels. We therefore ask where pretraining gains arise, how these two sources should be combined, and whether the information-flow choice identified from scratch remains valid after pretraining.

\question{3}{How can world--action knowledge be consolidated and transferred across domains?}

\subsubsection{Problem Setup and Evaluation Protocol}

\paragraph{Controlled Transfer Protocol.} All runs keep the backbones, optimization budget, and inference procedure fixed --- the dual-system joint self-attention architecture with synchronized denoising selected above --- and vary only whether and how the model is pretrained with embodiment data; the information-flow mask is revisited in the final ablation. Two complementary protocols serve the evaluation: \textbf{RoboTwin2.0-Clean2Random} fine-tunes on Clean and evaluates Clean as in-domain (ID) and Randomized as out-of-domain (OOD), exposing transfer; \textbf{RoboTwin2.0-Full} fine-tunes on the full RoboTwin2.0 training set and reports the mean success rate over both conditions.

\paragraph{Embodied Pretraining Data Mixture.} We compare supervised fine-tuning from scratch with three pretraining strategies under an identical 600-hour data budget, drawing egocentric human video from EgoDex~\citep{hoque2025egodex} and real-robot manipulation trajectories from RoboCOIN~\citep{wu2025robocoin}. \emph{Robot-only} spends the full 600-hour budget on robot data; the two mixed variants combine 350 hours of egocentric data with 250 hours of robot data, either in two stages (\emph{ego then robot}) or jointly in one stage (\emph{ego + robot co-train}). All four variants then undergo identical downstream fine-tuning.

\subsubsection{Embodied Pretraining Primarily Expands OOD Generalization}

\begin{wrapfigure}{r}{0.64\textwidth}
    \vspace{-14pt}
    \centering
    \includegraphics[width=\linewidth]{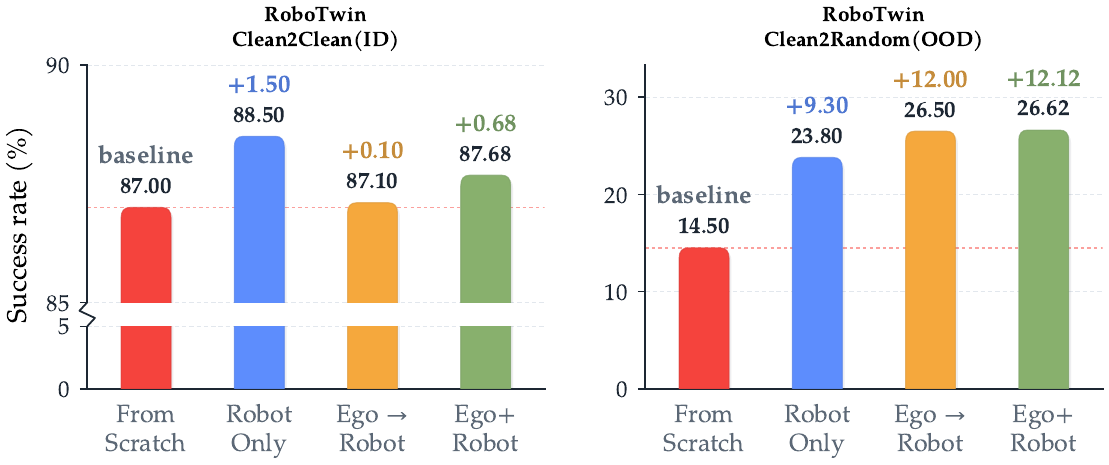}
    \caption{\textbf{Embodied Pretraining Primarily Improves OOD Generalization.} Success rates on RoboTwin2.0-Clean2Random, ordered from lower to higher performance within each evaluation setting.}
    \label{fig:pretraining-generalization}
    \vspace{-8pt}
\end{wrapfigure}

\paragraph{Pretraining Primarily Improves OOD Generalization.} As shown in \Cref{fig:pretraining-generalization}, embodied pretraining yields modest gains for in-domain performance, but yields strong performance gains in OOD evaluation. Embodied pretraining therefore contributes mainly knowledge that transfers beyond the downstream training distribution, rather than better fitting an already saturated ID benchmark.

\paragraph{Robot and Egocentric Data Contribute Different Strengths.} Robot-only pretraining yields the strongest ID performance, while both mixed strategies generalize better OOD. This trade-off is consistent with the insight of robot trajectories strengthening executable action grounding and egocentric video broadening the visual and interaction distribution. 

\paragraph{Absorbing Egocentric Videos: Sequential versus Co-Training.} Sequential training and one-stage co-training performance are nearly matched in both in-domain and OOD evaluation, indicating that using both sources matters more than their precise ordering. Co-training is marginally strongest overall and removes the extra curriculum transition, so we adopt it as the practical default.

\subsubsection{Pretraining Changes the Preferred Information Flow}

The from-scratch ablation establishes that the action stream must see the world stream, but leaves one-way and mutual visibility nearly tied. We repeat this comparison after cross-domain pretraining under both RoboTwin2.0-Clean2Random and RoboTwin2.0-Full.

\begin{figure}[!h]
    \centering
    \vspace{-6pt}
    \includegraphics[width=\linewidth]{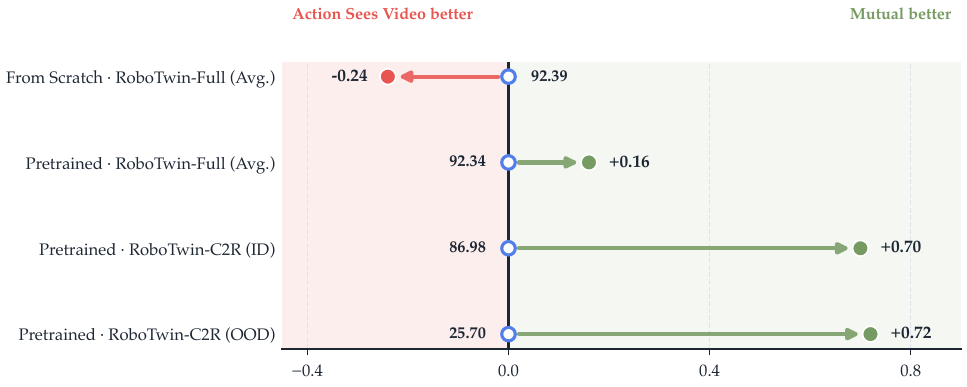}
    \captionof{figure}{\textbf{Pretraining Changes the Preferred Information Flow.} Central markers give the absolute success rate of Action Sees Video; arrows terminate at the matched Mutual result, with horizontal displacement reporting $\text{Mutual}-\text{Action Sees Video}$ in percentage points. Green and red denote gains and drops, respectively.}
    \label{fig:information-flow-scale}
    \vspace{-6pt}
\end{figure}

\paragraph{Mutual Visibility Becomes Preferable with Embodied Pretraining.} Without embodied pretraining, RoboTwin2.0-Full slightly favors one-way visibility; after pretraining, however, the same protocol favors Mutual. RoboTwin2.0-Clean2Random shows the same reversal in both ID and OOD, with comparable gains across the two splits (\Cref{fig:information-flow-scale}). The reversal is therefore neither an artifact of domain shift nor of the evaluation protocol: pretraining turns the world--action interaction into a genuinely bidirectional exchange, in which the predicted future frames provide visual guidance for action generation, while the predicted actions in turn inform the synthesis of the manipulator's motion in those frames. Without embodied pretraining, data scarcity likely prevents the two streams from reliably establishing such correspondences; the far more abundant pretraining data closes this gap, and Mutual accordingly realizes its advantage once embodied pretraining is in place. We carry Mutual into the final recipe.

\finding{3}{Embodied pretraining primarily expands OOD generalization. Robot trajectories preserve action grounding, egocentric video broadens transfer, and one-stage co-training integrates both effectively. At pretrained scale, mutual world--action visibility is consistently preferred.}

\vspace{-8pt}
\subsection{Concluding Remarks}

The three questions turn inherited world knowledge into a concrete model design: not a list of individually best hyperparameters, but a sequence in which each decision is tested under the conditions created by the previous one. The next section composes these defaults into \textbf{OpenWAM-}$\bm\alpha$ and asks whether they survive full-scale heterogeneous pretraining.

\noindent
\begin{minipage}{\linewidth}
    \centering
    \small
    \setlength{\tabcolsep}{5.5pt}
    \renewcommand{\arraystretch}{1.05}
    \captionsetup{skip=4pt}
    \captionof{table}{\textbf{The recipe accumulated by OpenWAM-Study.} Each evidence-backed choice becomes the default for \openwamalpha{}.}
    \label{tab:derived-recipe}
    \begin{tabular}{>{\raggedright\arraybackslash}m{0.16\linewidth}!{\color{black!20}\vrule width 0.45pt}>{\raggedright\arraybackslash}m{0.48\linewidth}!{\color{black!20}\vrule width 0.45pt}>{\raggedright\arraybackslash}m{0.27\linewidth}}
        \toprule[1.05pt]
        \rowcolor{black!4}
        \makebox[\linewidth][c]{\sffamily\bfseries Stage}
        & \makebox[\linewidth][c]{\sffamily\bfseries Findings}
        & \makebox[\linewidth][c]{\sffamily\bfseries Carried-forward default} \\
        \midrule
        \rowcolor{DiTblue5!6}
        \textcolor{DiTblue5}{\sffamily\bfseries Inherit} & Capable video backbones and compact representation latents transfer the strongest upstream priors. & Wan2.2-TI2V-5B; compact latent \\
        \addlinespace[1.6pt]
        \rowcolor{DiTyellow5!7}
        \textcolor{DiTyellow5}{\sffamily\bfseries Interact} & Dedicated action capacity and world-to-action visibility are necessary; synchronized denoising performs best. & Dual joint self-attention; synchronized denoising \\
        \addlinespace[1.6pt]
        \rowcolor{DiTgreen5!7}
        \textcolor{DiTgreen5}{\sffamily\bfseries Consolidate} & Embodied pretraining primarily improves OOD generalization and consistently favors mutual visibility. & One-stage ego + robot co-training; mutual visibility \\
        \bottomrule[1.05pt]
    \end{tabular}
\end{minipage}

\FloatBarrier

\clearpage

\section{\openwamalpha{}: From Principles to a Pretrained Model}
\label{sec:pretrain}

\textbf{Overview of \openwamalpha{}.} Motivated by the design principles and empirical insights uncovered through OpenWAM-Study, we instantiate these findings at scale in \openwamalpha{}, an open foundation world--action model for systematically investigating the capabilities and scaling behavior of world--action models across diverse robotic tasks.
In \Cref{sec:pretrain_phase_one}, we specify the final architecture, training recipe, and deployment scheme of \openwamalpha{} under the guidance of the insights established in \Cref{sec:recipe}. \Cref{sec:pretrain_phase_two} then details the pretraining data configuration together with the associated data curation and cleaning pipeline. Finally, \Cref{sec:pretrain_phase_three} evaluates \openwamalpha{} across a diverse set of simulation benchmarks, with analyses of its performance and the key empirical findings revealed by these evaluations, and \Cref{sec:realworld_eval} further evaluates it on real-world tasks.




\subsection{\openwamalpha{} Architecture, Training, and Deployment}
\label{sec:pretrain_phase_one}

\begin{figure}[h]
    \centering
    \includegraphics[width=\linewidth]{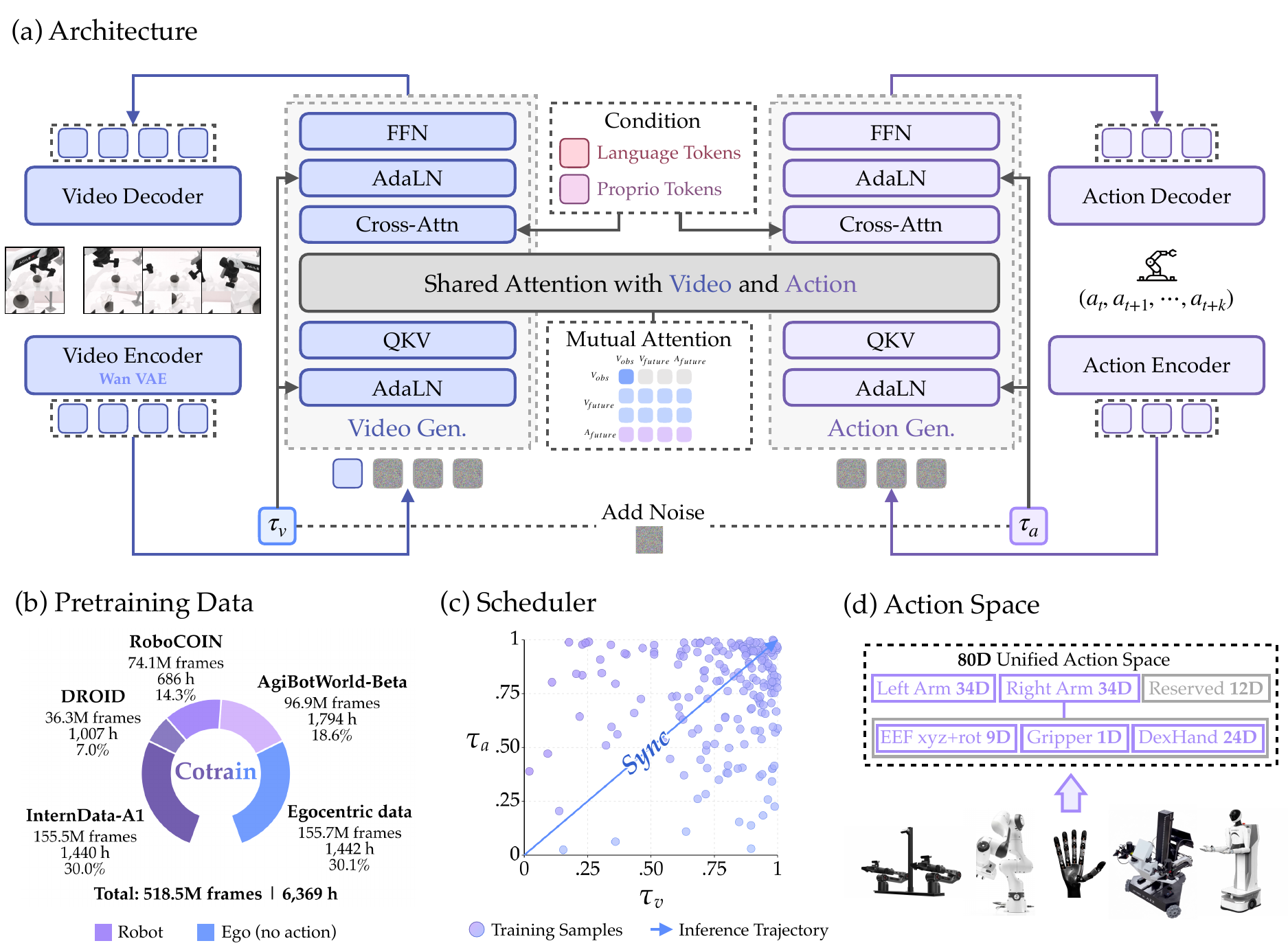}
    \caption{\textbf{Overview of \openwamalpha{}.} (a) The dual-system architecture: a video-generation DiT and an ActionDiT jointly denoise the future frames and the action chunk through shared attention under the mutual visibility mask, each conditioned on language and proprioception via cross-attention and carrying its own noise timestep. (b) The pretraining mixture: 518.5M frames (6{,}369 hours) of egocentric and robot data, co-trained in one stage. (c) Timestep sampling: training covers the full joint noise plane, while inference follows the synchronized diagonal. (d) The 80-D unified action space with fixed slot semantics shared across embodiments.}
    \label{fig:alpha-architecture}
\end{figure}

As shown in \Cref{fig:alpha-architecture}, the design principles distilled from \textbf{OpenWAM-Study} determine the configuration of \openwamalpha{} across its architecture, training, and deployment stages. The following paragraphs elaborate on each stage in turn.

\subsubsection{Architecture}

\textbf{\openwamalpha{}} adopts the architecture that \textbf{OpenWAM-Study} converges to (\Cref{tab:derived-recipe}), assembled from the modules of \Cref{sec:infra_model} as the composition $C(\mathcal{E},\mathcal{S},\mathcal{M})$: the frozen Wan2.2-VAE as the visual encoder $\mathcal{E}$; the pretrained Wan2.2-TI2V-5B DiT~\citep{wan2025wan} executing the world stream and a dedicated 1B-parameter ActionDiT executing the action stream, coupled through joint self-attention, as the stream backbones $\mathcal{S}=\{\mathcal{W},\mathcal{A}\}$; and the \emph{mutual} mode with first-frame-causal intra-video attention as the visibility mask $\mathcal{M}$. Conditioned on the current observation $\mathbf{o}_1$ (all camera views tiled into one canvas), the language instruction $\ell$, and the proprioceptive state $\mathbf{q}\in\mathbb{R}^{80}$ expressed in the unified action space, \openwamalpha{} jointly denoises the future frames $\mathbf{o}_{2:T}$ of a $T$-frame video window $\mathbf{o}_{1:T}$ in latent space, together with a continuous action chunk $\mathbf{a}=\mathbf{a}_{1:H}\in\mathbb{R}^{H\times80}$~\citep{zhao2023learning} (\Cref{fig:alpha-architecture}a).

\textbf{Tokenization and Context.} The frozen Wan2.2-VAE encodes $\mathbf{o}_{1:T}$ causally -- the first frame alone, subsequent frames in groups of four -- into $T'=1+(T-1)/4$ latent frames $\mathbf{z}$, so the first latent frame remains a clean anchor of the present. A $(1,2,2)$ patch embedding flattens the latent video into world-stream tokens of width 3072 carrying 3D RoPE over the (frame, height, width) grid; each of the $H$ noised action steps is linearly embedded into one action-stream token of width 1024 carrying 1D RoPE over the chunk index. The frozen umT5 encoder maps $\ell$ into a 4096-dimensional context, a linear projection appends $\mathbf{q}$ as one additional context token, and both streams consume the resulting context $\mathbf{c}$ through their own per-block cross-attention.

\textbf{Stream Bridging and Prediction.} All 30 paired layers of the two backbones act as bridge layers, where stream-owned projections map the two residual widths into a shared attention space of 24 heads $\times$ 128 dimensions; under the configured $\mathcal{M}$, the two streams read each other freely while the clean first-frame rows attend to neither noised future frames nor actions. AdaLN injects each stream's own denoising timestep -- $t_v$ token-wise into the world stream, with first-frame tokens pinned to the clean endpoint $t_v=1$, and $t_a$ into the action stream -- and linear heads decode both final states into velocities, realizing the joint forward pass of \Cref{sec:infra_train} at any joint noise state $(t_v,t_a)$; how training and inference each cover this plane is specified in \Cref{sec:alpha_training,sec:alpha_deployment}.

\subsubsection{Training}
\label{sec:alpha_training}

\textbf{Co-Training Setup.} Following the one-stage co-training strategy identified in \Cref{sec:embodied_pt_study}, \openwamalpha{} is trained end to end in a single stage on the ego + robot mixture of \Cref{sec:pretrain_phase_two}. The video DiT (initialized from the pretrained Wan2.2-TI2V-5B weights), the ActionDiT, and the proprioception encoder are all updated; only the umT5 text encoder and the Wan2.2-VAE remain frozen.

\textbf{Unified Action Supervision.} Heterogeneous embodiments meet in the unified action space of \Cref{sec:eval-protocol}: an 80-dimensional vector with fixed slot semantics, comprising two mirrored 34-D arm blocks -- end-effector position (3), 6D rotation (6), gripper (1), and dexterous hand (24) -- followed by 12 slots reserved for embodiment-specific channels (\Cref{fig:alpha-architecture}d). Each dataset scatters its native action and state into these slots, the validity mask $\mathbf{m}$ of \Cref{eq:wam-loss} confines action supervision to the populated coordinates, and the robot state at the start of the window supplies the proprioception token in $\mathbf{c}$.

\textbf{Objective and Timestep Sampling.} \openwamalpha{} is trained with the joint flow-matching objective of \Cref{eq:wam-loss}~\citep{black2024pi_0}, instantiated with $\lambda_v=\lambda_a=1$ and a bell-shaped timestep weight $w(\cdot)$ peaked at intermediate noise levels. The per-stream timesteps are drawn independently as $t_v=1-f_{\rho}(u_v)$ and $t_a=1-f_{\rho}(u_a)$ with $u_v,u_a\sim\mathcal{U}[0,1]$, where the timestep warp $f_{\rho}$ of \Cref{eq:schedule-families} is applied identically to both streams with $\rho=5$ to bias sampling toward high noise (\Cref{fig:alpha-architecture}c).

\subsubsection{Deployment}
\label{sec:alpha_deployment}

\textbf{Synchronized Denoising.} At test time, \openwamalpha{} follows the synchronized schedule selected in \Cref{sec:inference_time_flow}: both streams advance in lockstep along the diagonal of the joint noise plane, realized on the training warp as $t^i=1-f_{\rho}(1-i/N)$ with $\rho=5$ and $N=10$ steps, so neither stream leads. Denoising starts from Gaussian noise -- with the first latent frame clamped to the encoding of the current observation and re-pinned after every step -- and each step performs one joint forward pass and the coupled Euler update of \Cref{sec:infra_deploy}, so the action chunk is refined against progressively cleaner world features, and vice versa, at all 30 bridge layers. The finished chunk returns to each robot's native action space through the inverse map of \Cref{eq:unify-map}.

\textbf{Inference Mode.} All benchmark evaluations that follow, in simulation and in the real world alike, use synchronous inference: whenever the action buffer empties, execution pauses until the model predicts a fresh chunk from the current observation, and the robot then executes that chunk exactly as predicted. The reported results therefore reflect the model's own performance in the most direct way.

\textbf{Inference Acceleration.} \openwamalpha{} is served through the acceleration stack of \Cref{sec:infra_deploy}: the joint denoising loop is compiled into a fixed-shape graph replayed under CUDA graphs, stable velocity predictions are reused across adjacent steps, prompt embeddings are cached, and no VAE decode runs in the control path. The $N$-step loop completes in roughly $170$\,ms per chunk on an RTX 5090, well within real-time control budgets.

\FloatBarrier

\subsection{Multi-Domain Pretraining Data and Curation}
\label{sec:pretrain_phase_two}

\openwamalpha{} is pretrained on multi-domain data drawn from five sources spanning three data types: egocentric human data, real-world robot data, and synthetic robot data. From a raw pool of 1.33B frames ($\approx$14{,}300 hours), we construct a training set of 518M frames ($\approx$6{,}400 hours) through curation and per-source subsampling (\Cref{tab:pretrain-mixture}). This section describes how the mixture is composed (\Cref{sec:data_mixture}) and how each source is cleaned (\Cref{sec:data_curation}). We further provide the pretraining-stage hyperparameter configuration and other relevant details in \Cref{app:training_details}.

\medskip
\noindent\begin{minipage}{\linewidth}
    \centering
    \small
    \setlength{\tabcolsep}{5.0pt}
    \renewcommand{\arraystretch}{1.24}
    \captionof{table}{\textbf{The \openwamalpha{} pretraining data.} \emph{\#Emb.} counts each source's distinct embodiments. \emph{Task Coverage} marks the manipulation settings each source spans. \emph{Full} reports each source's raw size before processing, while \emph{Curated + Sampled} reports the data actually used for training, after cleaning (\Cref{sec:data_curation}) and per-source whole-episode subsampling (\Cref{sec:data_mixture}); \emph{Share} is each source's actual per-epoch sample share under proportional sampling.}
    \label{tab:pretrain-mixture}
    \resizebox{\linewidth}{!}{%
    \begin{tabular}{llcc cccc rr rrr}
        \toprule[1.05pt]
        & & & & \multicolumn{4}{c}{\sffamily\bfseries Task Coverage} & \multicolumn{2}{c}{\sffamily\bfseries Full} & \multicolumn{3}{c@{}}{\sffamily\bfseries Curated + Sampled} \\
        \cmidrule(lr){5-8}\cmidrule(lr){9-10}\cmidrule(l){11-13}
        \multicolumn{1}{@{}l}{\sffamily\bfseries Source} & {\sffamily\bfseries Type} & {\sffamily\bfseries \#Emb.} & {\sffamily\bfseries FPS} & Single & Bimanual & Mobile & Dexterous & Frames (M) & Hours & Frames (M) & Hours & Share (\%) \\
        \midrule
        \rowcolor{black!4}
        Egocentric data (ours) & Human video & 1 & 30 & \multicolumn{4}{c}{\emph{in-the-wild human manipulation}} & 744.9 & 6{,}897 & 155.7 & 1{,}442 & 30.1 \\
        AgiBotWorld-Beta~\citep{bu2025agibot} & Real robot & 1 & 15 & & $\checkmark$ & $\checkmark$ & $\checkmark$ & 124.5 & 2{,}306 & 96.9 & 1{,}794 & 18.6 \\
        \rowcolor{black!4}
        RoboCOIN~\citep{wu2025robocoin} & Real robot & 15 & 30 & & $\checkmark$ & $\checkmark$ & $\checkmark$ & 104.5 & 956 & 74.1 & 686 & 14.3 \\
        DROID~\citep{khazatsky2024droid} & Real robot & 1 & 10 & $\checkmark$ & & & & 46.3 & 1{,}285 & 36.3 & 1{,}007 & 7.0 \\
        \rowcolor{black!4}
        InternData-A1~\citep{tian2025interndata} & Simulation & 4 & 30 & $\checkmark$ & $\checkmark$ & & & 313.7 & 2{,}904 & 155.5 & 1{,}440 & 30.0 \\
        \midrule
        \textbf{Total} & & 21 robot + human & & \multicolumn{4}{c}{} & \textbf{1{,}333.9} & \textbf{14{,}348} & \textbf{518.5} & \textbf{6{,}369} & \textbf{100.0} \\
        \bottomrule[1.05pt]
    \end{tabular}%
    }
\end{minipage}
\medskip

\subsubsection{Pretraining Data Mixture}
\label{sec:data_mixture}

\textbf{Data Composition.} Following the co-training recipe of \Cref{sec:embodied_pt_study}, the mixture combines three complementary data types. Egocentric human data comes from a dataset we carefully constructed for manipulation-centric world modeling -- 71.6K long-form first-person recordings of 0.25--6 minutes each, covering 3{,}006 everyday manipulation tasks; it supplies broad visual and interaction diversity but carries no robot action labels, so its action and proprioception channels remain fully masked and it supervises only the world stream. Real-world robot data (AgiBotWorld-Beta~\citep{bu2025agibot}, RoboCOIN~\citep{wu2025robocoin}, and DROID~\citep{khazatsky2024droid}) grounds the action stream with executable trajectories across 17 physical platforms, while also providing the most faithful visual observations of robots interacting with the physical world. Synthetic robot data (InternData-A1~\citep{tian2025interndata}) further broadens the coverage of robot data, encompassing a more comprehensive range of single-arm and bimanual manipulation skills under diverse environmental variations.

\textbf{Data Budget and Sampling.} Considering the compute resources and time cost of pretraining, each source is subsampled under a per-source hour budget. The budgets are derived from frame-based targets -- the egocentric and synthetic sources each contribute 30\% of the total training frames, and the remaining 40\% is divided among the three real-robot sources in proportion to their curated valid-frame counts -- so that sources with different native frame rates are balanced by the quantity of data the model actually consumes. Within each source, the hour budget is water-filled across its constituent sub-datasets, and whole episodes are subsampled from the curated pool under a fixed seed until the budget is met. Training then draws samples proportionally to the actual per-source counts, so every retained sample is visited exactly once per epoch.

\subsubsection{Data Curation}
\label{sec:data_curation}

Aggregating data across different sources and embodiments introduces heterogeneous defects in both the visual and the signal channel. Our cleaning protocol is informed by the data-cleaning pipeline of Qwen-RobotManip~\citep{yuan2026qwen}, supplemented with rules for the failure modes we observe in the collected sources, and operates at two levels: vision-level cleaning shared by all sources, and signal-level cleaning specific to robot data.

\textbf{Vision-Level Cleaning.} All sources first pass a uniform visual-quality screen that removes undecodable video, frozen or duplicated frames, black, white, and solid-color frames, over- and under-exposure and exposure flicker, blurred frames, and abrupt visual jumps. Egocentric data further exhibits one failure mode of its own: segments in which the hands leave the field of view carry no manipulation signal and are removed; recordings with empty or invalid language annotations are likewise discarded.

\textbf{Signal-Level Cleaning.} Robot data additionally carries state and action channels, which are cleaned in four steps:
\begin{enumerate}
    \item \textbf{Signal integrity.} An episode is discarded outright when its recorded end-effector state fails to track the commanded actions (amplitude ratio $\geq 3\times$ with per-axis correlation $<0.5$), or when video--signal misalignment affects more than 2\% of frames.
    \item \textbf{State-first idle detection.} The state channel serves as the primary criterion for idle footage: leading and trailing segments whose state is static -- detected with per-robot motion thresholds calibrated from the p99.5 of single-frame deltas, and confirmed when average end-effector translation and geodesic rotation rates fall below 2\,cm/s and 5\,$^\circ$/s -- are trimmed, whereas mid-episode pauses are never cut, since cutting them would splice temporally non-adjacent frames. State discontinuities such as jerk and spike outliers are additionally screened with robust median--MAD thresholds.
    \item \textbf{Visual cross-checking.} When the state does move, it is cross-checked against the visuals: apparent state motion under which every camera view remains visually static is attributed to sensor jitter and trimmed as well, whereas a genuinely moving arm observed by a frozen camera marks a capture defect and the episode is removed.
    \item \textbf{Episode-level deletion.} An episode that loses more than 70\% of its frames to the steps above, or whose video is frozen for 90\% or more of its length, is dropped entirely.
\end{enumerate}

\subsection{Simulation Benchmark Evaluation}
\label{sec:pretrain_phase_three}

Starting from the pretrained \openwamalpha{} base model, we conduct supervised fine-tuning and evaluation on the eight simulation benchmarks integrated in OpenWAM-Infra (\Cref{sec:eval-protocol}), spanning five embodiment categories:
\begin{itemize}[leftmargin=1.5em]
    \item \textbf{Single-arm}: LIBERO~\citep{liu2023libero}, LIBERO-Plus~\citep{fei25libero-plus}, and VLABench~\citep{zhang2024vlabench};
    \item \textbf{Bimanual}: RoboTwin2.0~\citep{chen2025robotwin} and RoboDojo~\citep{chen2026robodojounifiedsimandrealbenchmark};
    \item \textbf{Mobile single-arm}: RoboCasa365~\citep{robocasa365};
    \item \textbf{Mobile bimanual}: EBench~\citep{gao2026ebench};
    \item \textbf{Dexterous-hand}: RoboCasa-GR1~\citep{nvidia2025gr00tn1,robocasa2024}.
\end{itemize}
For RoboTwin2.0, we evaluate two variants. \textbf{RoboTwin2.0-Full} fine-tunes on the mixture of clean and randomized data and then evaluates under both conditions, probing the model's in-distribution (ID) capability; \textbf{RoboTwin2.0-Clean2Random} fine-tunes on clean data only and evaluates under both conditions, probing out-of-distribution (OOD) generalization.

\begin{figure}[!htb]
    \centering
    \includegraphics[width=\linewidth]{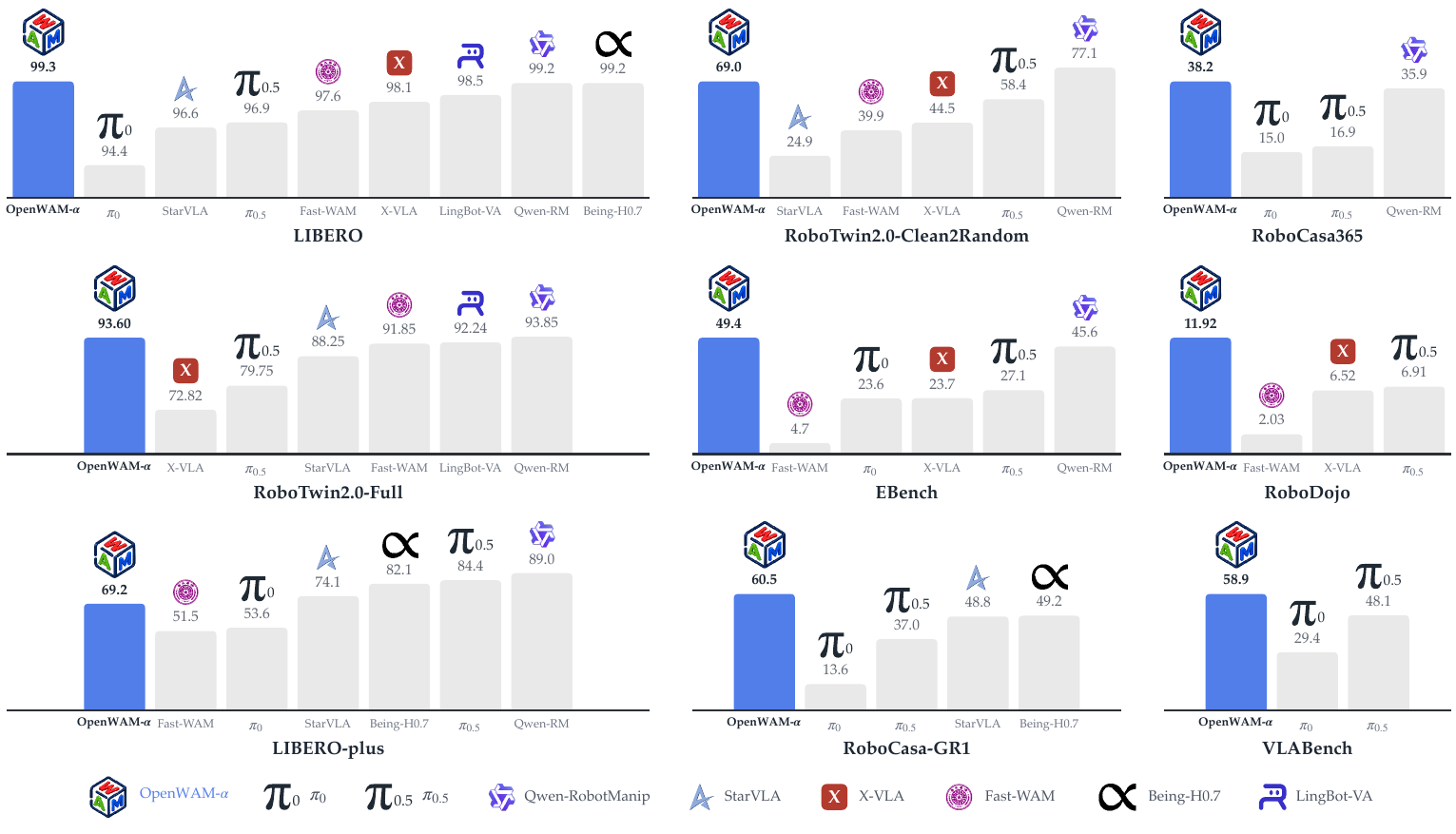}
    \caption{\textbf{Score comparison of \openwamalpha{} against representative VLA and WAM baselines across the simulation benchmarks.} Within each panel, baselines are ordered by score, and every bar is labeled with its actual value.}
    \label{fig:alpha-score-comparison}
\end{figure}

\begin{wrapfigure}{r}{0.5\linewidth}
    \centering
    \vspace{-15mm}
    \includegraphics[width=\linewidth]{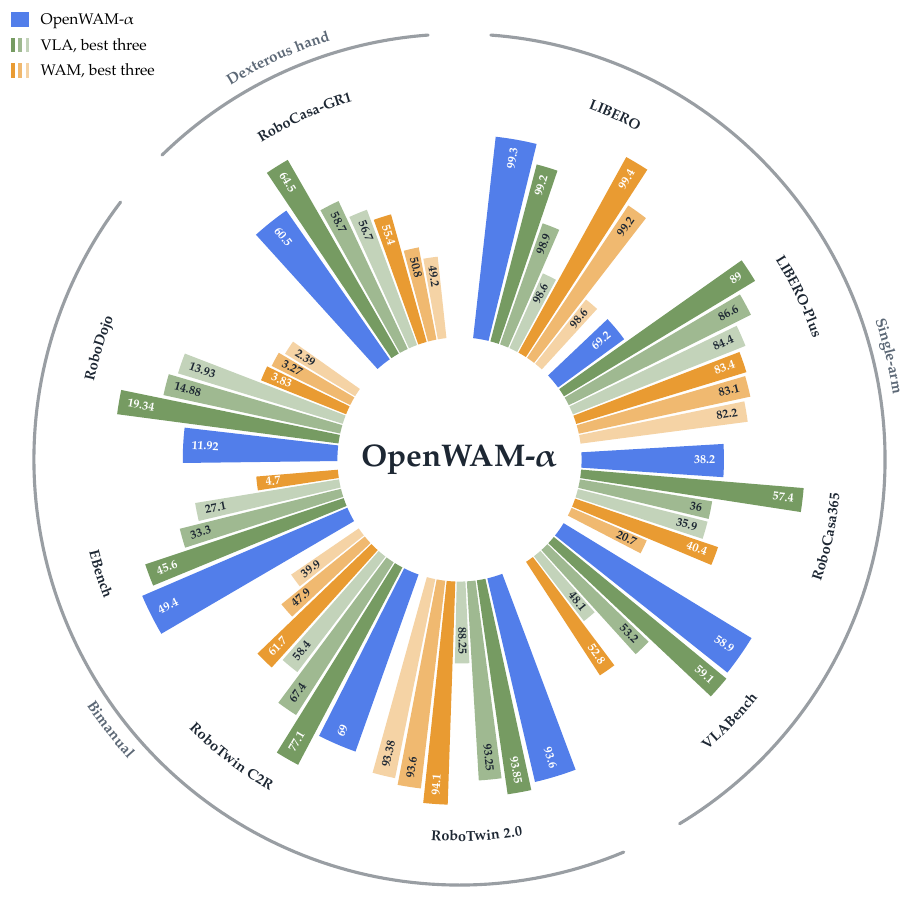}
    \caption{\textbf{\openwamalpha{} against the best of each family, per benchmark},
    grouped by embodiment.}
    \label{fig:vla-vs-wam-radial}
    \vspace{-3em}
\end{wrapfigure}

\Cref{fig:alpha-score-comparison} summarizes the scores of \openwamalpha{} alongside representative VLA and WAM baselines on each benchmark, with every bar labeled by its actual score, giving a clear account of where the model stands. \Cref{fig:vla-vs-wam-radial} complements this view by pitting \openwamalpha{} against the strongest VLAs and WAMs on every leaderboard, grouped by embodiment, so that the two paradigms can be compared directly. The detailed per-benchmark training and evaluation configurations, including hyperparameter settings and evaluation details, are provided in \Cref{app:sft_configuration}. The per-benchmark tables behind both figures are reported in \Cref{app:simulation_tables}. Through these scores, we seek to answer the two questions at the heart of this evaluation: \textbf{(1) how does \openwamalpha{} perform, and (2) between VLA and WAM, which paradigm prevails?} The following two subsections address them in turn.

\subsubsection{How Does \openwamalpha{} Perform?}
\label{sec:sim_eval_alpha}

Across the majority of the benchmarks, \openwamalpha{} delivers excellent performance:
\begin{itemize}[leftmargin=1.5em]
    \item On the single-arm benchmarks \textbf{LIBERO} and \textbf{VLABench}, the bimanual benchmark \textbf{RoboTwin2.0-Full}, the mobile single-arm benchmark \textbf{RoboCasa365}, and the dexterous-hand benchmark \textbf{RoboCasa-GR1}, \openwamalpha{} sits firmly in the top tier, within a marginal gap of the best model.
    \item On the mobile bimanual benchmark \textbf{EBench}, \openwamalpha{} sets the state of the art, leading the runner-up \textsc{Qwen-RobotManip} by roughly 4 points in both SR and Score.
    \item On the bimanual benchmarks \textbf{RoboTwin2.0-Clean2Random} and \textbf{RoboDojo}, a gap to the best models (which are VLAs) remains, yet \openwamalpha{} is the strongest WAM on both leaderboards, ahead of the other WAMs by a clear margin.
\end{itemize}

The unexpected exception is the single-arm benchmark \textbf{LIBERO-Plus}, where the scores of \openwamalpha{} fall markedly below its standing elsewhere, as shown in \Cref{tab:liberoplus}. The per-perturbation breakdown of \textbf{LIBERO-Plus} is telling: the losses concentrate under the \emph{camera} and \emph{noise} perturbations, with visible deficits under the \emph{background} and \emph{layout} perturbations as well. On the very same leaderboard, however, ABot-M0.5, ImageWAM, and Being-H0.7 --- all WAMs themselves --- perform strongly, with scores approaching the state of the art. We therefore compare \openwamalpha{} against these models along two axes, pretraining data and architecture, to identify the underlying causes.

\begin{wrapfigure}{r}{0.45\textwidth}
    \centering
    \vspace{-1.1em}
    \includegraphics[width=\linewidth]{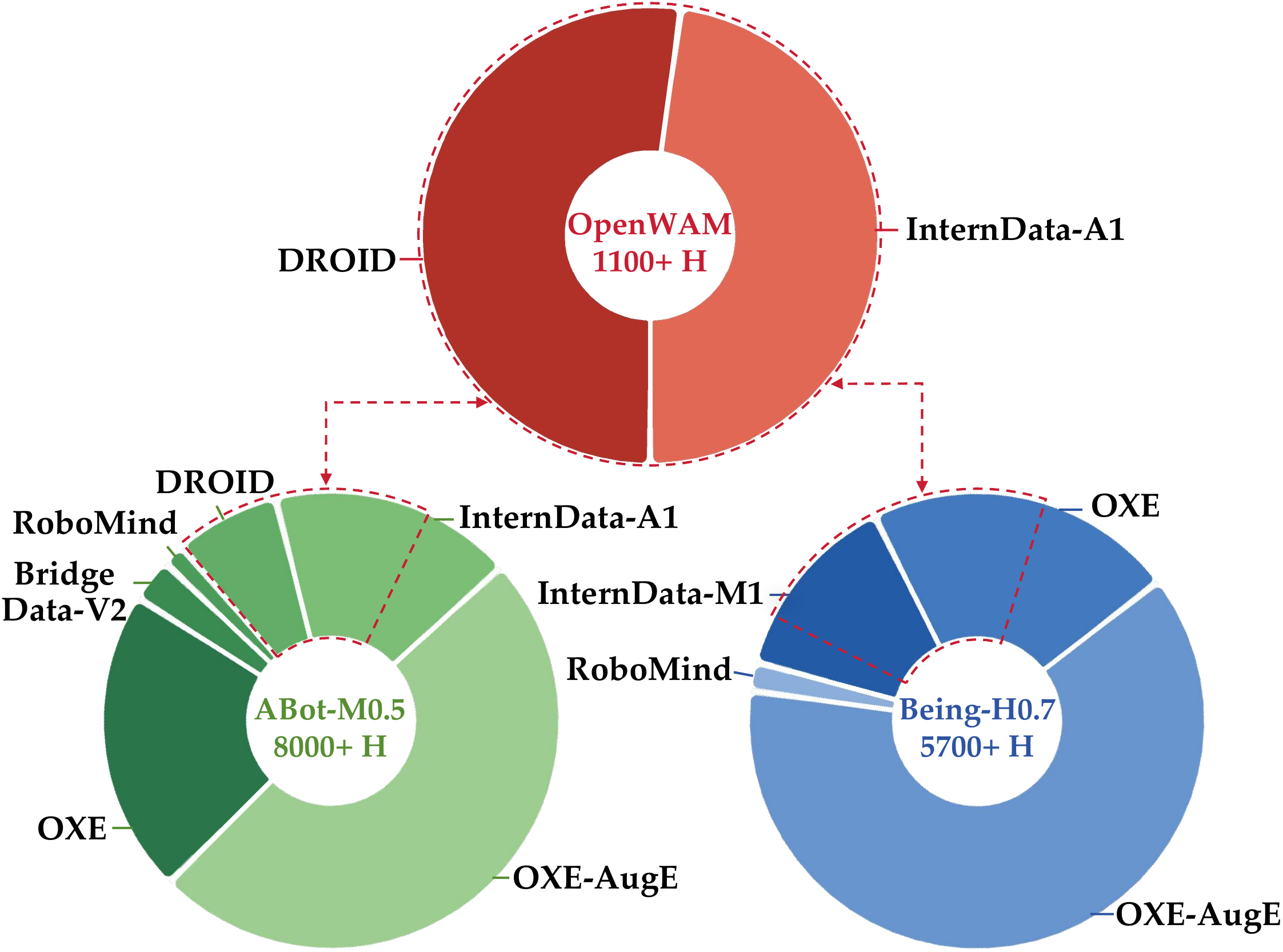}
    \caption{\textbf{Single-arm pretraining data of ABot-M0.5, Being-H0.7, and \openwamalpha{}.} The dashed lines indicate that the single-arm data of \openwamalpha{} amounts to only a small fraction of what ABot-M0.5 and Being-H0.7 consume.}
    \label{fig:singlearm-data}
\end{wrapfigure}

\textbf{The Data Perspective.} \Cref{fig:singlearm-data} contrasts the single-arm portion of the pretraining data of ABot-M0.5 and Being-H0.7 with that of \openwamalpha{}. Both baselines pretrain on far larger and more varied single-arm collections, spanning diverse embodiments, scenes, and camera viewpoints, so during pretraining they have already seen visual information and world knowledge close to the LIBERO-Plus test scenes --- in viewpoint and noise as much as in background and layout. In contrast, the single-arm data of \openwamalpha{} (\Cref{tab:pretrain-mixture}) is far smaller in both volume and variety, drawing on only two sources: DROID, collected on a fixed single-arm platform with fixed camera viewpoints, and the single-arm portion of the synthetic InternData-A1. With such limited single-arm coverage, the model receives far less single-arm world knowledge, and its single-arm generalization suffers accordingly on the OOD perturbations of LIBERO-Plus. The converse also holds: the \openwamalpha{} mixture is rich in egocentric, bimanual, and dexterous-hand data, and the model is correspondingly strong on the bimanual and dexterous-hand benchmarks.

\begin{table}[t]
\centering
\scriptsize
\setlength{\tabcolsep}{5pt}
\renewcommand{\arraystretch}{1.15}
\captionsetup{justification=centering,singlelinecheck=true}
\caption{\textbf{Evaluation Results on LIBERO-Plus.} Bold denotes best values, underline second best.}
\label{tab:liberoplus}
\resizebox{\linewidth}{!}{%
\begin{tabular}{lcccccccc}
\toprule
 & \textbf{Camera} & \textbf{Robot} & \textbf{Language} & \textbf{Light} & \textbf{Background} & \textbf{Noise} & \textbf{Layout} & \textbf{Avg} \\
\midrule
\rowcolor{black!8}
\multicolumn{9}{c}{\textit{\textbf{VLA}}} \\
\midrule
\textbf{\boldmath $\pi_0$}~\citep{black2024pi_0} & 13.8 & 6.0 & 58.8 & 85.0 & 81.4 & 79.0 & 68.9 & 53.6 \\
\rowcolor{black!4}
\textbf{OpenVLA-OFT}~\citep{kim2025fine} & 56.4 & 31.9 & 79.5 & 88.7 & 93.3 & 75.8 & 74.2 & 69.6 \\
\textbf{StarVLA}~\citep{community2026starvla} & 52.5 & 49.8 & 88.5 & 95.7 & 95.7 & 73.0 & 76.9 & 74.1 \\
\rowcolor{black!4}
\textbf{ABot-M0}~\citep{yang2026abot} & 60.4 & 67.9 & 86.4 & 96.2 & 91.6 & 86.4 & 82.6 & 80.5 \\
\textbf{\boldmath $\pi_{0.5}$}~\citep{intelligence2025pi_} & 78.4 & 73.6 & 80.8 & 96.2 & 94.1 & 89.0 & 84.5 & 84.4 \\
\rowcolor{black!4}
\textbf{ACoT-VLA}~\citep{zhong2026acot} & 72.6 & \underline{82.6} & 87.5 & 97.7 & \underline{96.5} & 87.8 & \underline{88.1} & \underline{86.6} \\
\textbf{\textsc{Qwen-RobotManip}}~\citep{yuan2026qwen} & \textbf{87.2} & 75.5 & 85.6 & 96.6 & \textbf{97.7} & \textbf{97.7} & 87.3 & \textbf{89.0} \\
\midrule
\rowcolor{black!8}
\multicolumn{9}{c}{\textit{\textbf{WAM}}} \\
\midrule
\textbf{Fast-WAM}~\citep{yuan2026fast} & 16.4 & 44.5 & 68.9 & 78.2 & 53.7 & 37.7 & 60.7 & 51.5 \\
\rowcolor{black!4}
\textbf{Being-H0.7}~\citep{luo2026beingh07} & \underline{82.0} & 59.0 & 82.8 & \underline{97.8} & 90.0 & 93.5 & \textbf{88.5} & 82.1 \\
\textbf{Cosmos-Policy}~\citep{kim2026cosmos} & 75.8 & 63.3 & 81.7 & 96.5 & 88.9 & 92.7 & 82.2 & 82.2 \\
\rowcolor{black!4}
\textbf{ImageWAM}~\citep{zhang2026imagewam} & 80.8 & 50.3 & \textbf{91.4} & \textbf{98.1} & 85.5 & \underline{93.8} & 80.5 & 83.1 \\
\textbf{ABot-M0.5}~\citep{chen2026abot} & 70.5 & \textbf{87.4} & \underline{88.6} & 94.0 & 89.7 & 75.5 & 85.2 & 83.4 \\
\midrule
\textbf{\openwamalpha{}} & 33.8 & 76.1 & 88.0 & 97.0 & 87.1 & 39.8 & 77.5 & 69.2 \\
\bottomrule
\end{tabular}%
}
\end{table}

\textbf{The Architecture Perspective.} Fast-WAM, ABot-M0.5, and \openwamalpha{} share one core prediction target --- a pixel-level video of the future over a temporal horizon, with the intermediate latents produced by the reconstructive Wan2.2-VAE. On LIBERO-Plus, all three exhibit the same signature: the scores under the \emph{camera} and \emph{noise} perturbations fall clearly below those under the other perturbations, and only ABot-M0.5, backed by its pretraining data, recovers much of the loss relative to Fast-WAM and \openwamalpha{}. Pixel-level information is evidently acutely sensitive to camera and noise perturbations --- an inherent limitation of pixel-level prediction architectures that only large-scale pretraining can compensate. Two further baselines corroborate this reading. ImageWAM, although not pretrained, remains conspicuously strong on LIBERO-Plus: its prediction target is also a pixel-level latent, but it predicts only a single future frame of the current observation --- closer to an edit than a rollout --- so no error accumulates across frames and the impact of future-pixel prediction on the camera and noise scores shrinks accordingly. Being-H0.7, in turn, encodes observations with V-JEPA 2.1: its intermediate latents remain temporal (several frames are encoded jointly), yet they are semantic-level features rather than pixel reconstructions, which makes the model markedly more robust to the camera and noise perturbations. Unlike LIBERO-Plus, the OOD designs of the other benchmarks impose no deliberate camera or noise disturbance, so \openwamalpha{} remains highly competitive there; on LIBERO-Plus, the camera and noise disturbances compound the single-arm data deficit above, and the performance of \openwamalpha{} inevitably degrades.

\takeaway{1}{How well an embodied model generalizes on a benchmark is ultimately determined by whether its pretraining mixture contains data close to the benchmark's test conditions, in both embodiment and environment. Extending \Cref{sec:embodied_pt_study}, the decisive ingredient of an embodied foundation model remains large-scale, diverse, scene-rich robot manipulation data, which at once covers the broad range of test scenarios a model may later encounter and supplies precise action / visual information grounded in the embodiment --- the most direct route to stronger generalization.}

\takeaway{2}{Driven by large-scale data, WAMs that predict the future in a pixel latent space can achieve excellent performance, yet their robustness to visual disturbance is inherently limited. Echoing \Cref{sec:visual_representation_priors}, a representation that is robust to environmental variation, information-rich, and sufficiently compact is still needed to push WAM performance further.}

\subsubsection{VLA versus WAM: Which Paradigm Prevails?}
\label{sec:sim_eval_paradigm}

\begin{figure}[t]
    \centering
    \includegraphics[width=\linewidth]{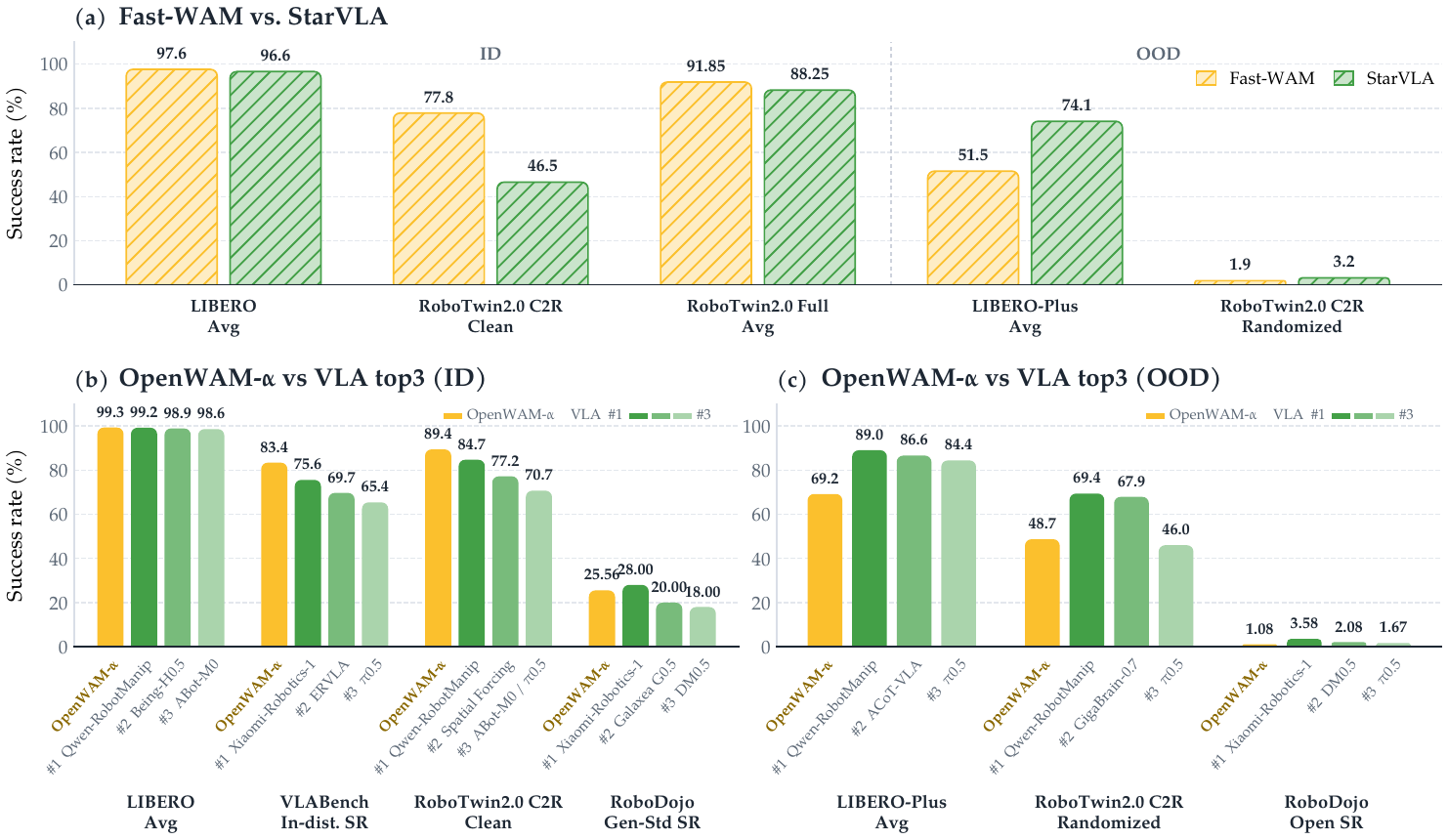}
    \caption{\textbf{ID and OOD comparisons between the two paradigms.} (a) Fast-WAM versus StarVLA, two models without embodied pretraining, on ID and OOD splits. (b) \openwamalpha{} versus the three strongest VLAs on ID splits. (c) \openwamalpha{} versus the three strongest VLAs on OOD splits.}
    \label{fig:id-ood-benchmarks}
    \vspace{-4mm}
\end{figure}

Across all benchmarks (\Cref{fig:vla-vs-wam-radial}), the VLA and WAM groups show no substantial gap in overall success rate, and each side places standout models at the top of leaderboards: ABot-M0.5 and \openwamalpha{} among WAMs, Xiaomi-Robotics-1 and \textsc{Qwen-RobotManip} among VLAs. Since these models differ in pretraining data, architecture, and training configuration alike, neither paradigm can be declared superior outright. The fine-grained scores, however, reveal a consistent pattern: each paradigm holds an advantage region of its own (\Cref{fig:id-ood-benchmarks}).

\textbf{In Distribution, WAMs Fit Better.} The cleanest comparison is between StarVLA and Fast-WAM, two models without embodied pretraining (\Cref{fig:id-ood-benchmarks}a): on LIBERO, the Clean split of RoboTwin2.0-Clean2Random, and RoboTwin2.0-Full, the WAM attains visibly higher scores on these ID tasks, fitting the training distribution more effectively than its VLA counterpart. The pretrained models tell the same story from both directions (\Cref{fig:id-ood-benchmarks}b): \openwamalpha{} leads the three strongest VLAs on LIBERO, the In-dist.\ split of VLABench, and the Clean split of RoboTwin2.0-Clean2Random, and stays within a small gap of the best on the Gen-Std split of RoboDojo --- even though the pretraining data of these VLAs exceeds ours. This advantage traces back to the video-latent supervision in WAM training: whereas a VLA is optimized purely against action supervision, with no intermediate latent target, the video-latent term injects an additional source of information into parameter optimization, allowing a WAM to fit the training data more closely.


\textbf{Out of Distribution, VLAs Generalize Better.} The same StarVLA--Fast-WAM comparison reverses out of distribution (\Cref{fig:id-ood-benchmarks}a): on LIBERO-Plus the VLA leads by a wide margin, and even on the Randomized split of RoboTwin2.0-Clean2Random, where both models collapse, the ordering still favors the VLA. The pretrained models mirror the reversal (\Cref{fig:id-ood-benchmarks}c): on LIBERO-Plus, the Randomized split of RoboTwin2.0-Clean2Random, and the Open split of RoboDojo, \openwamalpha{} trails the state-of-the-art VLAs by an evident margin. The mechanism is the flip side of the ID advantage: long-horizon future prediction adds a supervision signal that helps fitting, but under distribution shift the same long horizon means heavier error accumulation --- a burden the action-only VLA never carries --- so WAMs perform visibly below VLAs in unseen evaluation environments.

Crucially, both deficits are remediable by data. Whether it is the ID fitting deficit of VLAs or the OOD generalization deficit of WAMs, sufficiently rich pretraining data --- covering complex environmental variation and carrying precise action annotation --- lets either paradigm draw on the inherited priors to achieve both strong fitting and strong generalization at test time. Data therefore remains the first priority of model development. At the same time, VLAs and WAMs are both end-to-end models built on the same core information flow, from observation to action; how to combine the complementary strengths of the two paradigms, and thereby push the capability boundary of end-to-end models further, remains a question well worth pursuing.

\takeaway{3}{Neither paradigm prevails outright: video-latent supervision gives WAMs the edge in in-distribution fitting, while VLAs generalize better out of distribution --- and either deficit can be compensated by sufficiently large and diverse pretraining data. Combining the complementary strengths of the two end-to-end paradigms is a promising route to push the capability boundary further.}

\subsection{Real-Robot Evaluation}
\label{sec:realworld_eval}

\begin{figure}[!t]
    \centering
    \includegraphics[width=\linewidth]{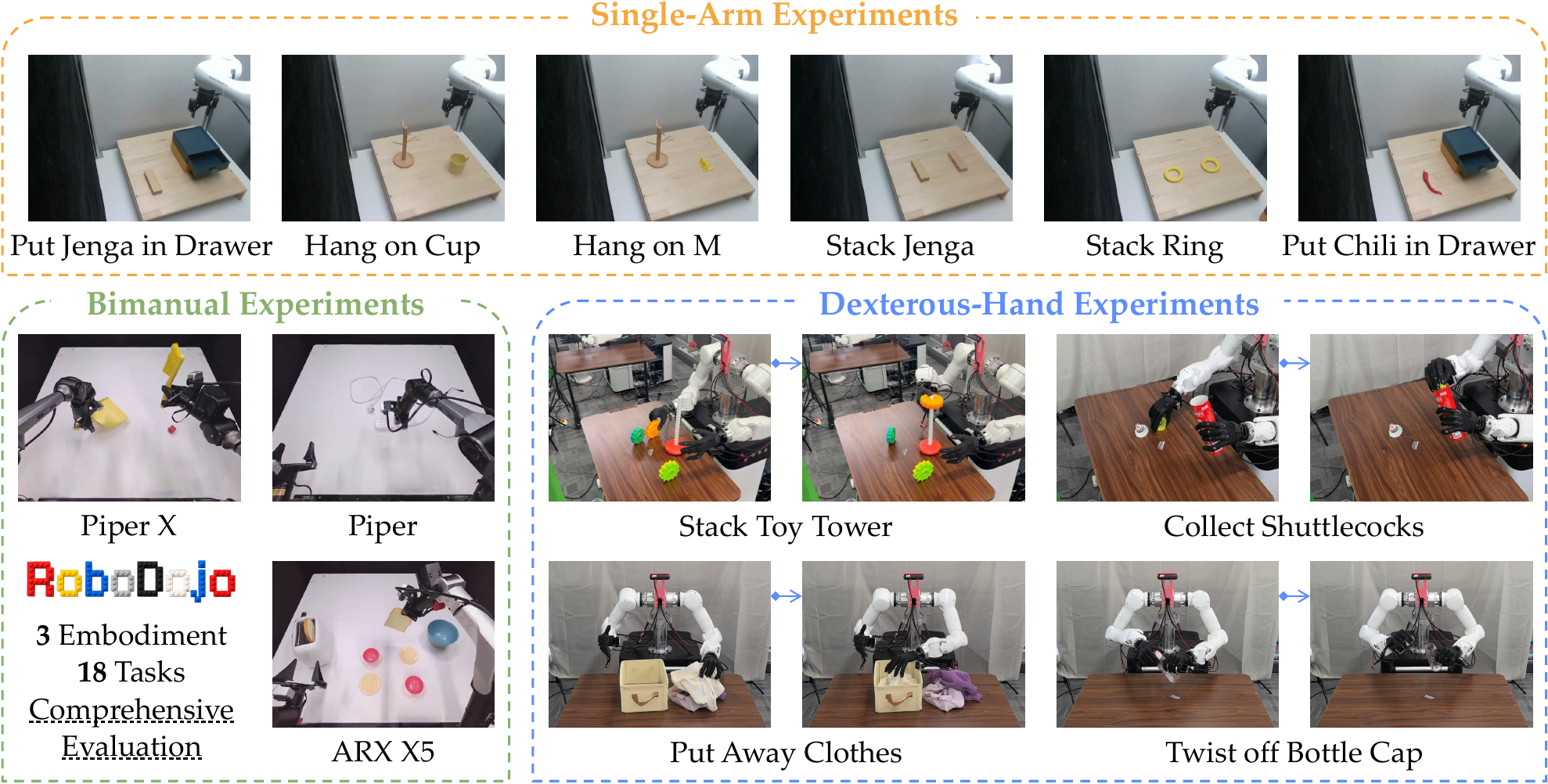}
    \caption{\textbf{Real-robot experimental setups across three embodiments.} \emph{Top}: the six single-arm tasks on the Franka-Research-3 platform. \emph{Bottom left}: the three bimanual embodiments of the RoboDojo real-world track (Piper X, Piper, and ARX X5), covering 18 tasks in total. \emph{Bottom right}: the four dexterous-hand tasks on the Wuji-hand and Tianji-arm platform, each illustrated by key intermediate stages of its execution.}
    \label{fig:realworld-setup}
\end{figure}

To further examine \openwamalpha{} beyond simulation and validate both its general capability and its generalization, we conduct comprehensive real-robot evaluations across three embodiments --- single-arm, bimanual, and dexterous-hand --- with the experimental setups shown in \Cref{fig:realworld-setup}. Specifically:
\begin{itemize}[leftmargin=1.5em]
    \item \textbf{Single-arm experiments} are conducted on the Franka-Research-3 platform and cover three task families --- stacking, pick-and-place, and hanging --- probing the model's basic and fine-grained manipulation capabilities. Performance is measured by task success rate (SR).
    \item \textbf{Bimanual experiments} are conducted on the official RoboDojo real-robot platform, spanning three embodiments (ARX X5, Piper, and Piper X); following the RoboDojo task taxonomy, the evaluation covers generalization, precision, long-horizon, memory, and open tasks, assessing the model comprehensively. Performance is measured by SR and Progress Score.
    \item \textbf{Dexterous-hand experiments} are conducted on a platform pairing the Wuji dexterous hand with the Tianji robotic arm --- an embodiment and action space absent from the \openwamalpha{} pretraining mixture --- and cover bimanual-interactive, long-horizon, and fine manipulation tasks, probing how well the model adapts and generalizes to unseen embodiments and unseen action dimensions. Performance is measured by SR and Progress Score.
\end{itemize}

The task setups and evaluation protocols of each embodiment are documented in \Cref{app:realworld_protocol}, and the SFT configurations used for these experiments are provided in \Cref{app:sft_configuration}.

\Cref{tab:realworld}, \Cref{tab:robodojo_real_world_benchmark}, and \Cref{tab:realworld-dex} report the detailed scores of the single-arm, RoboDojo bimanual, and dexterous-hand experiments, respectively.

\begin{table}[!h]
\centering
\scriptsize
\setlength{\tabcolsep}{2.9pt}
\renewcommand{\arraystretch}{1.15}
\captionsetup{justification=centering,singlelinecheck=true}
\caption{\textbf{Evaluation Results on Single-Arm Real-Robot Tasks.} Bold denotes best values, underline second best.}
\label{tab:realworld}
\resizebox{\linewidth}{!}{%
\begin{tabular}{lccccccc}
\toprule
 & \textbf{\makecell{Stack\\Jenga}} & \textbf{\makecell{Stack\\Ring}} & \textbf{\makecell{Put Chili\\in Drawer}} & \textbf{\makecell{Put Jenga\\in Drawer}} & \textbf{\makecell{Hang\\on M}} & \textbf{\makecell{Hang\\on Cup}} & \textbf{Avg} \\
\midrule
\textbf{\boldmath $\pi_{0.5}$}~\citep{intelligence2025pi_} & 13/20 (65\%) & 7/20 (35\%) & 14/20 (70\%) & 15/20 (75\%) & 7/20 (35\%) & 10/20 (50\%) & 66/120 (55.0\%) \\
\rowcolor{black!4}
\textbf{LingBot-VA}~\citep{li2026causal} & \underline{16/20 (80\%)} & \textbf{15/20 (75\%)} & \underline{17/20 (85\%)} & \underline{17/20 (85\%)} & \textbf{13/20 (65\%)} & \underline{15/20 (75\%)} & \underline{93/120 (77.5\%)} \\
\midrule
\textbf{\openwamalpha{}} & \textbf{17/20 (85\%)} & \underline{12/20 (60\%)} & \textbf{20/20 (100\%)} & \textbf{20/20 (100\%)} & \textbf{13/20 (65\%)} & \textbf{17/20 (85\%)} & \textbf{99/120 (82.5\%)} \\
\bottomrule
\end{tabular}%
}
\end{table}

On the single-arm platform (\Cref{tab:realworld}), \openwamalpha{} clearly leads both LingBot-VA, a representative WAM, and $\pi_{0.5}$, a representative VLA, on the majority of tasks, and attains the best average success rate, providing initial evidence of its general and fine-grained manipulation capabilities. We then turn to RoboDojo-Real (\Cref{tab:robodojo_real_world_benchmark}), a real-robot benchmark that comprehensively evaluates generalist manipulation policies across three bimanual embodiments and a wide range of task dimensions. \openwamalpha{} tops the leaderboard: it remains consistently strong across the embodiments and their tasks, and reaches the state of the art on the great majority of them, corroborating the generality and robustness of the model in the real world.

\providecommand{\best}[1]{\textbf{#1}}
\providecommand{\secondbest}[1]{\underline{#1}}

\begin{table*}[!t]
\centering
\scriptsize
\setlength{\tabcolsep}{4pt}
\renewcommand{\arraystretch}{1.15}
\captionsetup{justification=centering,singlelinecheck=true}
\caption{\textbf{Evaluation Results on the Bimanual RoboDojo Real-World Track.} Each cell reports Score / SR (\%). Bold denotes best values, underline second best. Rows are shaded by embodiment.}
\label{tab:robodojo_real_world_benchmark}
\providecommand{\res}[2]{#1 / #2}
\providecommand{\na}{--}
\providecommand{\cellA}{\cellcolor{black!5}}
\providecommand{\cellB}{\cellcolor{black!10}}
\resizebox{\textwidth}{!}{%
\begin{tabular}{llcccccccc}
\toprule
\textbf{Policy} & \textbf{Embodiment} & \textbf{Task 1} & \textbf{Task 2} & \textbf{Task 3} & \textbf{Task 4} & \textbf{Task 5} & \textbf{Task 6} & \textbf{Emb. Avg.} & \textbf{Overall Avg.} \\
\midrule

& ARX X5
& \res{0.0}{0.0}
& \res{0.0}{0.0}
& \res{0.0}{0.0}
& \res{2.0}{0.0}
& \res{20.7}{10.0}
& \res{18.0}{0.0}
& \res{6.8}{1.7}
& \\
& \cellA Piper
& \cellA\res{0.0}{0.0}
& \cellA\res{5.3}{0.0}
& \cellA\res{0.0}{0.0}
& \cellA\res{0.0}{0.0}
& \cellA\res{53.0}{50.0}
& \cellA\res{37.0}{0.0}
& \cellA\res{15.9}{8.3}
& \\
\multirow{-3}{*}{\textbf{X-VLA}~\citep{zheng2026x}}
& \cellB Piper X
& \cellB\res{0.0}{0.0}
& \cellB\res{0.0}{0.0}
& \cellB\res{0.0}{0.0}
& \cellB\res{0.0}{0.0}
& \cellB\res{0.0}{0.0}
& \cellB\res{0.7}{0.0}
& \cellB\res{0.1}{0.0}
& \multirow{-3}{*}{\res{7.6}{3.3}} \\
\midrule

& ARX X5
& \res{24.0}{\best{20.0}}
& \res{0.0}{0.0}
& \res{3.0}{0.0}
& \res{4.0}{0.0}
& \res{40.0}{\secondbest{20.0}}
& \res{19.0}{\secondbest{10.0}}
& \res{15.0}{8.3}
& \\
& \cellA Piper
& \cellA\res{0.0}{0.0}
& \cellA\res{0.0}{0.0}
& \cellA\res{0.0}{0.0}
& \cellA\res{0.0}{0.0}
& \cellA\res{23.0}{20.0}
& \cellA\res{22.7}{0.0}
& \cellA\res{7.6}{3.3}
& \\
\multirow{-3}{*}{\textbf{Xiaomi-Robotics-0}~\citep{cai2026xiaomi}}
& \cellB Piper X
& \cellB\res{0.0}{0.0}
& \cellB\res{0.7}{0.0}
& \cellB\res{\secondbest{4.0}}{0.0}
& \cellB\res{0.0}{0.0}
& \cellB\res{0.0}{0.0}
& \cellB\res{2.7}{0.0}
& \cellB\res{1.2}{0.0}
& \multirow{-3}{*}{\res{7.9}{3.9}} \\
\midrule

& ARX X5
& \res{1.0}{0.0}
& \res{0.0}{0.0}
& \res{3.0}{0.0}
& \res{6.0}{0.0}
& \res{0.0}{0.0}
& \res{10.3}{0.0}
& \res{3.4}{0.0}
& \\
& \cellA Piper
& \cellA\res{0.0}{0.0}
& \cellA\res{\secondbest{32.7}}{\secondbest{10.0}}
& \cellA\res{\secondbest{13.3}}{\secondbest{10.0}}
& \cellA\res{0.0}{0.0}
& \cellA\res{56.0}{50.0}
& \cellA\res{30.0}{10.0}
& \cellA\res{22.0}{13.3}
& \\
\multirow{-3}{*}{\textbf{GalaxeaVLA (G0)}~\citep{jiang2025galaxea}}
& \cellB Piper X
& \cellB\res{0.0}{0.0}
& \cellB\res{0.0}{0.0}
& \cellB\res{0.0}{0.0}
& \cellB\res{4.0}{0.0}
& \cellB\res{0.0}{0.0}
& \cellB\res{6.0}{0.0}
& \cellB\res{1.7}{0.0}
& \multirow{-3}{*}{\res{9.0}{4.4}} \\
\midrule

& ARX X5
& \res{0.0}{0.0}
& \res{0.0}{0.0}
& \res{0.0}{0.0}
& \res{0.0}{0.0}
& \res{\secondbest{48.0}}{\secondbest{20.0}}
& \res{12.0}{0.0}
& \res{10.0}{3.3}
& \\
& \cellA Piper
& \cellA\res{0.0}{0.0}
& \cellA\res{7.3}{0.0}
& \cellA\res{0.0}{0.0}
& \cellA\res{0.0}{0.0}
& \cellA\res{\secondbest{73.0}}{\secondbest{70.0}}
& \cellA\res{59.0}{\secondbest{40.0}}
& \cellA\res{23.2}{18.3}
& \\
\multirow{-3}{*}{\textbf{InternVLA-A1}~\citep{cai2026internvla}}
& \cellB Piper X
& \cellB\res{0.0}{0.0}
& \cellB\res{0.0}{0.0}
& \cellB\res{\secondbest{4.0}}{0.0}
& \cellB\res{2.0}{0.0}
& \cellB\res{0.0}{0.0}
& \cellB\res{\secondbest{10.0}}{0.0}
& \cellB\res{2.7}{0.0}
& \multirow{-3}{*}{\res{12.0}{7.2}} \\
\midrule

& ARX X5
& \res{\secondbest{24.6}}{\best{20.0}}
& \res{\best{1.8}}{0.0}
& \res{\best{25.8}}{\best{10.0}}
& \res{\secondbest{47.0}}{\best{20.0}}
& \res{40.0}{\secondbest{20.0}}
& \res{\secondbest{26.8}}{\secondbest{10.0}}
& \res{\secondbest{27.7}}{\secondbest{13.3}}
& \\
& \cellA Piper
& \cellA\res{\best{10.0}}{\best{10.0}}
& \cellA\res{28.0}{0.0}
& \cellA\res{10.0}{\secondbest{10.0}}
& \cellA\res{0.0}{0.0}
& \cellA\res{72.0}{60.0}
& \cellA\res{\secondbest{72.0}}{\best{50.0}}
& \cellA\res{\secondbest{32.0}}{\secondbest{21.7}}
& \\
\multirow{-3}{*}{\textbf{\boldmath $\pi_{0.5}$}~\citep{intelligence2025pi_}}
& \cellB Piper X
& \cellB\res{0.0}{0.0}
& \cellB\res{\secondbest{14.8}}{\best{10.0}}
& \cellB\res{0.0}{0.0}
& \cellB\res{\secondbest{29.5}}{\secondbest{10.0}}
& \cellB\res{\secondbest{7.5}}{0.0}
& \cellB\res{3.0}{0.0}
& \cellB\res{\secondbest{9.1}}{\secondbest{3.3}}
& \multirow{-3}{*}{\res{\secondbest{22.9}}{\secondbest{12.8}}} \\
\midrule

& ARX X5
& \res{\best{31.0}}{0.0}
& \res{0.0}{0.0}
& \res{\secondbest{13.0}}{0.0}
& \res{\best{52.0}}{\best{20.0}}
& \res{\best{100.0}}{\best{100.0}}
& \res{\best{38.0}}{\best{20.0}}
& \res{\best{39.0}}{\best{23.3}}
& \\
& \cellA Piper
& \cellA\res{\secondbest{8.3}}{0.0}
& \cellA\res{\best{60.0}}{\best{40.0}}
& \cellA\res{\best{36.7}}{\best{30.0}}
& \cellA\res{0.0}{0.0}
& \cellA\res{\best{100.0}}{\best{100.0}}
& \cellA\res{\best{75.0}}{\best{50.0}}
& \cellA\res{\best{46.7}}{\best{36.7}}
& \\
\multirow{-3}{*}{\textbf{\openwamalpha{}}}
& \cellB Piper X
& \cellB\res{0.0}{0.0}
& \cellB\res{\best{28.0}}{\best{10.0}}
& \cellB\res{\best{18.0}}{0.0}
& \cellB\res{\best{52.3}}{\best{20.0}}
& \cellB\res{\best{40.0}}{\best{40.0}}
& \cellB\res{\best{24.7}}{\best{10.0}}
& \cellB\res{\best{27.2}}{\best{13.3}}
& \multirow{-3}{*}{\res{\best{37.6}}{\best{24.4}}} \\
\bottomrule
\end{tabular}%
}
\begin{minipage}{0.98\textwidth}
\vspace{0.5em}
\footnotesize
\textit{Task order.}
\textbf{ARX X5}: \texttt{cover\_blocks}, \texttt{make\_bread}, \texttt{make\_food}, \texttt{pack\_and\_pour\_fruit}, \texttt{store\_in\_safe}, \texttt{insert\_tubes}.
\textbf{Piper}: \texttt{stack\_and\_cover\_blocks}, \texttt{fill\_pen\_holder}, \texttt{put\_objects\_into\_basket}, \texttt{insert\_charger}, \texttt{stack\_bowls}, \texttt{stand\_up\_bottles}.
\textbf{Piper X}: \texttt{classify\_objects}, \texttt{disassemble\_LEGO}, \texttt{hang\_mugs}, \texttt{pack\_objects\_into\_backpack}, \texttt{sweep\_blocks}, \texttt{cap\_pen}.
\end{minipage}
\end{table*}

\begin{table*}[htbp]
\centering
\scriptsize
\setlength{\tabcolsep}{4pt}
\renewcommand{\arraystretch}{1.15}
\captionsetup{justification=centering,singlelinecheck=true}
\caption{\textbf{Evaluation Results on Dexterous-Hand Real-Robot Tasks.} Each cell reports Score / SR (\%); the OOD column aggregates all variation settings of its task. Bold denotes best values.}
\label{tab:realworld-dex}

\resizebox{\linewidth}{!}{%
\begin{tabular}{lcccc}
\toprule
 & \multicolumn{2}{c}{\textbf{Stack Toy Tower}} & \multicolumn{2}{c}{\textbf{Collect Shuttlecocks}} \\
\cmidrule(lr){2-3}\cmidrule(lr){4-5}
\textbf{Method} & \textbf{ID} & \textbf{OOD} & \textbf{ID} & \textbf{OOD} \\
\midrule
$\pi_{0.5}$
& 16/30 (53.3) / 1/10 (10\%)
& 12/30 (40.0) / 1/10 (10\%)
& 11/35 (31.4) / 2/10 (20\%)
& 17/54 (31.5) / 3/15 (20\%) \\
\rowcolor{black!4}
\openwamalpha{}
& \textbf{21/30 (70.0) / 4/10 (40\%)}
& \textbf{17/30 (56.7) / 3/10 (30\%)}
& \textbf{29/35 (82.9) / 6/10 (60\%)}
& \textbf{30/57 (52.6) / 4/15 (26.7\%)} \\
\bottomrule
\end{tabular}%
}

\vspace{0.6em}

\resizebox{\linewidth}{!}{%
\begin{tabular}{lcccc}
\toprule
 & \multicolumn{2}{c}{\textbf{Put Away Clothes}} & \multicolumn{2}{c}{\textbf{Twist off Bottle Cap}} \\
\cmidrule(lr){2-3}\cmidrule(lr){4-5}
\textbf{Method} & \textbf{ID} & \textbf{OOD} & \textbf{ID} & \textbf{OOD} \\
\midrule
$\pi_{0.5}$
& 26/30 (86.7) / 6/10 (60\%)
& 45/60 (75.0) / 9/20 (45\%)
& 8/10 (80.0) / 3/10 (30\%)
& 13/20 (65.0) / 6/20 (30\%) \\
\rowcolor{black!4}
\openwamalpha{}
& \textbf{30/30 (100.0) / 10/10 (100\%)}
& \textbf{55/60 (91.7) / 17/20 (85\%)}
& \textbf{10/10 (100.0) / 7/10 (70\%)}
& \textbf{18/20 (90.0) / 16/20 (80\%)} \\
\bottomrule
\end{tabular}%
}

\end{table*}

To further probe the extensibility and generalization of \openwamalpha{}, we fine-tune the pretrained model on four dexterous manipulation tasks built on the Wuji-hand and Tianji-arm platform and test it under both in-domain and out-of-domain setups (\Cref{tab:realworld-dex}). Neither the platform nor its action space --- a 9-D end-effector pose combined with 21 dexterous-hand degrees of freedom --- ever appears in the \openwamalpha{} pretraining mixture; nevertheless, \openwamalpha{} outperforms $\pi_{0.5}$ by a clear margin across all tasks and setups, demonstrating that the model adapts reliably and stably to an entirely unseen embodiment.

Together, these experiments assess \openwamalpha{} across three embodiment types and a broad spectrum of real-world manipulation tasks. The results show that \openwamalpha{} performs strongly in every setting and stands on par with today's leading models, indicating that it can serve as a strong baseline for further development and comparison by the community.

\section{Conclusions}

This work introduced \textbf{OpenWAM}, an open research stack that turns world--action modeling from a set of tightly coupled implementation choices into a controlled experimental program. \textbf{OpenWAM-Infra} factorizes the WAM design space into composable modules assembled into three architecture families, served by a single trainer, policy server, and evaluation protocol spanning eight simulation benchmarks and real robots. On this substrate, \textbf{OpenWAM-Study} examined what world knowledge a WAM should inherit, how world and action learning create synergy, and how that synergy consolidates across domains, distilling the answers into a concrete recipe. \textbf{\openwamalpha{}} then instantiated this recipe at scale on egocentric human and robot data through a unified action space, delivering consistently strong results across the simulation benchmarks and real-robot experiments on single-arm, bimanual, and dexterous-hand platforms.

We release the full stack, including the infrastructure, evaluation protocols, pretrained weights, and data recipes, as a shared and reproducible foundation for world--action research, with \openwamalpha{} serving as a strong baseline for further development and comparison. Looking ahead, these findings point to larger and more diverse embodied data with precise action annotation, visual representations that are both compact and robust to environmental variation, and end-to-end designs that combine the complementary strengths of WAMs and VLAs as the most promising directions for WAMs. A more detailed discussion of limitations and future work is provided in \Cref{sec:limitations}.

\section*{Acknowledgements}
We thank Nilaksh and Chuning Zhu for their helpful discussions. We thank Wuji Technology for providing compute resources, which are crucial for the completion of this project.

\phantomsection
\addcontentsline{toc}{section}{References}

\bibliography{paper}
\bibliographystyle{arxiv-numbered}

\clearpage
\appendix

\addtocontents{toc}{\protect\setcounter{tocdepth}{1}}

\phantomsection
\addcontentsline{toc}{section}{Appendix}

\section*{Appendix}\label{app:appendix}

This appendix provides supplementary analyses and implementation details supporting the main paper:
\begin{itemize}[noitemsep,topsep=0pt,parsep=0pt,partopsep=0pt,leftmargin=1.5em]
    \item \S\ref{sec:limitations} discusses limitations of our work and directions for future research.
    \item \S\ref{app:training_details} documents the pretraining configuration and the dataset-specific SFT configurations used during post-training.
    \item \S\ref{app:realworld_protocol} details the task setups and evaluation protocols of the real-world experiments.
    \item \S\ref{app:simulation_tables} reports the full per-benchmark simulation scores behind \Cref{fig:alpha-score-comparison}.
\end{itemize}

\section{Limitations and Future Work}
\label{sec:limitations}

While OpenWAM provides a fully-open, systematic exploration towards world--action model pretraining, it has several limitations and opens up interesting future directions worth exploring.
\begin{enumerate}[noitemsep,topsep=2pt,parsep=2pt,partopsep=0pt,leftmargin=1.5em]
    \item \textbf{Training phases.} We mostly focus on the embodied pretraining phase of world--action modeling. Post-training and adaptation methods can lead to significant improvements for embodied foundation models, and the empirical recipe as well as underlying mechanisms for these methods remain open questions.
    \item \textbf{Architecture.} Across the six architecture variants currently supported by OpenWAM, we mostly explore modality fusion through cross-modality attention or hard-routed MoE. Drawing experience from the Unified Multimodal Model (UMM) community, we encourage future work to explore more native modality-fusion techniques, such as tokenization-phase early fusion and soft-routed MoE.
    \item \textbf{Pretraining data mixture.} We did not include UMI-style (e.g., UMI \citep{chi2024universal}, DexUMI \citep{xu2025dexumi}) collected data. In theory, UMI-style data offers task and scene diversity comparable to human egocentric videos, which is a crucial component for out-of-domain generalization capabilities. Co-training with data that contain robot-executable actions but are diverse in scene and task level, which can either be collected through UMI-style interfaces or post-processing pipelines, may offer a more data-efficient path towards autonomous embodied machine intelligence. In addition, we look forward to further breakthroughs in simulation for embodied AI: simulation can natively generate robot manipulation data with diverse scenes and realistic motion trajectories, unconstrained by the time and labor costs of the physical world, and thus holds unbounded potential for scaling robot data by orders of magnitude.
    \item \textbf{Visual encoder.} Weighing the compression of candidate encoders in both the temporal and the token dimension, we ultimately adopt Wan2.2-VAE as the final encoder of OpenWAM --- a choice that reflects the best trade-off currently available rather than an optimal solution: our evaluations reveal that pixel-reconstruction encoders such as Wan2.2-VAE are not sufficiently robust to viewpoint, noise, and scene variations. A latent representation that is compact while carrying sufficient environment information is still needed to push WAM performance further, and merits deeper exploration.
\end{enumerate}

\section{Training Details}
\label{app:training_details}

This section documents the optimization and data-loading configurations used to train and adapt \openwamalpha{}. We organize the details into two stages: multi-domain pretraining and dataset-specific supervised fine-tuning (SFT) during post-training.

\subsection{Pretraining Configuration}
\label{app:pretraining_configuration}

The key hyperparameters used for multi-domain pretraining are summarized in \Cref{tab:pretraining-configuration}. Pretraining uses 16 nodes with eight NVIDIA H200 GPUs per node (128 GPUs in total) and takes approximately seven days. With 24 clips per GPU and no gradient accumulation, the global batch size is 3{,}072 clips per optimizer step. The source-level data budgets and realized mixture proportions are reported separately in \Cref{tab:pretrain-mixture}.

\medskip
\noindent\begin{minipage}{\linewidth}
    \centering
    \small
    \setlength{\tabcolsep}{8pt}
    \renewcommand{\arraystretch}{1.08}
    \captionof{table}{\textbf{Pretraining configuration for \openwamalpha{}.}}
    \label{tab:pretraining-configuration}
    \begin{tabular}{@{}>{\raggedright\arraybackslash}p{0.42\linewidth}!{\vrule width 0.5pt}>{\centering\arraybackslash}p{0.48\linewidth}@{}}
        \toprule[1.05pt]
        \sffamily\bfseries Configuration & \sffamily\bfseries Value \\
        \midrule
        Compute & 16 nodes (128 NVIDIA H200 GPUs) \\
           Training time & $\approx 7$ days \\
        Optimizer & AdamW \\
        Batch size & 3{,}072 (24 per GPU) \\
        Learning rate & $1\times10^{-4}$ \\
        LR schedule & Cosine; 5\% warmup; minimum ratio $0.01$ \\
        Weight decay & $0.01$ \\
        Optimizer momentum & $\beta_1,\beta_2=0.9,0.95$ \\
        Training iterations & 155{,}862 (1 epoch) \\
        Gradient clipping & Global norm $1.0$ \\
        Model precision & bfloat16 \\
        Distributed training & DeepSpeed ZeRO Stage~2 \\
        \midrule
        Input clip & 33 frames; video stride 4; window stride 1 \\
        Image resolution & $384\times320$ \\
        Multi-view input & Enabled \\
        Image augmentation & \texttt{ColorJitter}(0.2, 0.2, 0.2, 0.0) \\
        Flow shifts & Video/action: $5.0/5.0$ \\
        Loss weights & $\lambda_v=1.0$, $\lambda_a=1.0$ \\
        Unified control space & 80-D action; 80-D proprioceptive state \\
        \bottomrule[1.05pt]
    \end{tabular}
\end{minipage}
\medskip

\subsection{Dataset-Specific SFT Configuration}
\label{app:sft_configuration}

All downstream models are initialized from the same pretrained \openwamalpha{} checkpoint. Supervised fine-tuning keeps the pretraining configuration of \Cref{tab:pretraining-configuration} unchanged and differs only in the three benchmark-dependent settings summarized in \Cref{tab:sft-configuration}: the global batch size, the number of training epochs or steps, and whether image augmentation is applied. Training length is given in epochs over the fine-tuning set, with the corresponding number of optimizer steps in parentheses, or directly in optimizer steps where no epoch-based schedule was used. Image augmentation, where enabled, is the same \texttt{ColorJitter}(0.2, 0.2, 0.2, 0.0) used in pretraining. LIBERO-Plus is evaluated with the LIBERO checkpoint without further fine-tuning.

\medskip
\noindent\begin{minipage}{\linewidth}
    \centering
    \small
    \setlength{\tabcolsep}{5pt}
    \renewcommand{\arraystretch}{1.08}
    \captionof{table}{\textbf{Dataset-specific SFT configuration for \openwamalpha{}.} Settings not listed are identical to pretraining (\Cref{tab:pretraining-configuration}); ``--'' denotes no image augmentation.}
    \label{tab:sft-configuration}
    \begin{tabular}{@{}l c c c@{}}
        \toprule[1.05pt]
        \sffamily\bfseries Benchmark & \sffamily\bfseries Batch size & \sffamily\bfseries Training Epochs / Steps & \sffamily\bfseries Augmentation \\
        \midrule
        \rowcolor{black!8}
        \multicolumn{4}{c}{\textit{\textbf{Simulation benchmarks}}} \\
        \midrule
        LIBERO & 256 & 10 epochs (10{,}690 steps) & -- \\
        VLABench & 196 & 6k steps & ColorJitter \\
        RoboTwin2.0-Full & 256 & 5 epochs (118{,}655 steps) & -- \\
        RoboTwin2.0-Clean2Random & 256 & 5 epochs (10{,}740 steps) & ColorJitter \\
        RoboDojo & 256 & 60k steps & ColorJitter \\
        RoboCasa365 & 1{,}024 & 60k steps & ColorJitter \\
        EBench & 256 & 100k steps & ColorJitter \\
        RoboCasa-GR1 & 256 & 100k steps & ColorJitter \\
        \midrule
        \rowcolor{black!8}
        \multicolumn{4}{c}{\textit{\textbf{Real-robot experiments}}} \\
        \midrule
        Single-arm (Franka-Research-3) & 256 & 10 epochs (9{,}860 steps) & -- \\
        Bimanual (RoboDojo real-world track) & 256 & 30k steps & ColorJitter \\
        Dexterous hand (Wuji + Tianji) & 256 & 5 epochs (10{,}925 steps) & -- \\
        \bottomrule[1.05pt]
    \end{tabular}
\end{minipage}
\medskip

\section{Real-World Evaluation Protocols}
\label{app:realworld_protocol}

This section details the real-world evaluation of \Cref{sec:realworld_eval}: for each embodiment, we document the task setup and the corresponding evaluation protocol.

\subsection{Single-Arm Real-Robot Experiments}
\label{app:realworld_singlearm}

\begin{figure}[H]
\centering
\includegraphics[width=\linewidth]{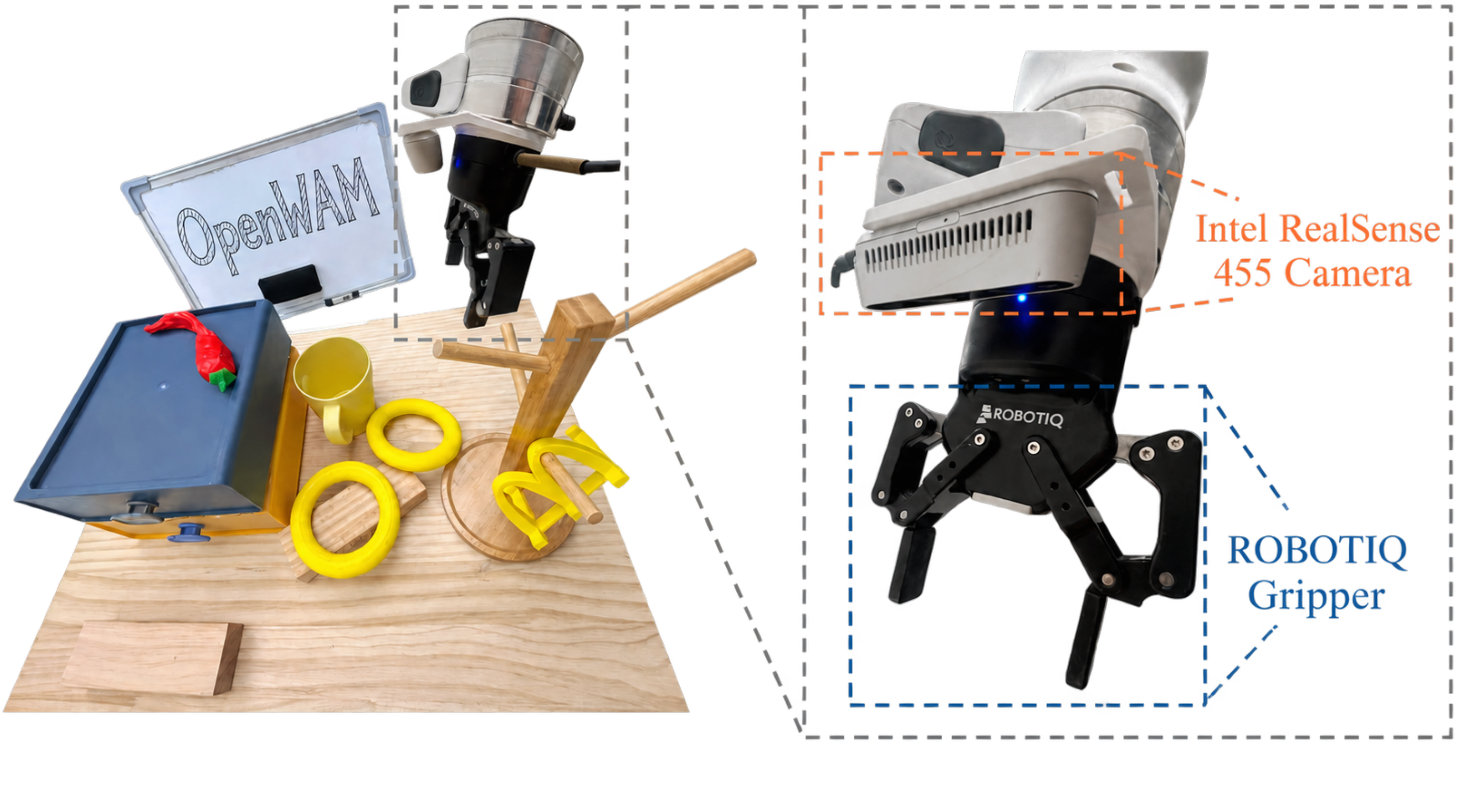}
\caption{Single-arm real-robot setup on the Franka-Research-3 platform. The
workspace contains the drawer, stacking objects, and hanging fixtures
used across the six tasks, while the robot is equipped with an Intel
RealSense camera and a Robotiq parallel gripper for closed-loop execution.}
\label{fig:singlearm-realworld-setup}
\end{figure}

\paragraph{Task Setup.}
We evaluate single-arm policies on the Franka-Research-3 platform using six
real-world tabletop tasks, covering stacking, hanging, and drawer
manipulation. The tasks use a Franka-Research-3 arm with a parallel
gripper and RGB observation cameras, as shown in
\Cref{fig:singlearm-realworld-setup}. Each task is specified by a natural-language
instruction and instantiated with a fixed physical scene: stacking tasks
place two target objects on the tabletop, hanging tasks place the object
and shelf in the workspace, and drawer tasks place the object on the
table next to an upper drawer.

\begin{table}[H]
\centering
\scriptsize
\setlength{\tabcolsep}{4pt}
\renewcommand{\arraystretch}{1.15}
\caption{Task instructions used in the single-arm real-robot evaluation.}
\label{tab:singlearm_task_setup}
\begin{tabular}{lp{0.72\linewidth}}
\toprule
\textbf{Task} & \textbf{Instruction} \\
\midrule
Stack Ring &
Pick the yellow ring on the left side, stack it on the other ring. \\
Stack Jenga &
Pick the jenga on the left side, stack it on the other jenga. \\
Hang on Cup &
Pick the cup on the table, hang it on the shelf. \\
Hang on M &
Pick the M-shaped object on the table, hang it on the shelf. \\
Put Chili in Drawer &
Pick the chili on the table, put it into the drawer, then push the upper drawer closed. \\
Put Jenga in Drawer &
Pick up the jenga block on the table, put it into the drawer, then push the upper drawer closed. \\
\bottomrule
\end{tabular}
\end{table}

\begin{figure}[H]
\centering
\includegraphics[width=1\textwidth]{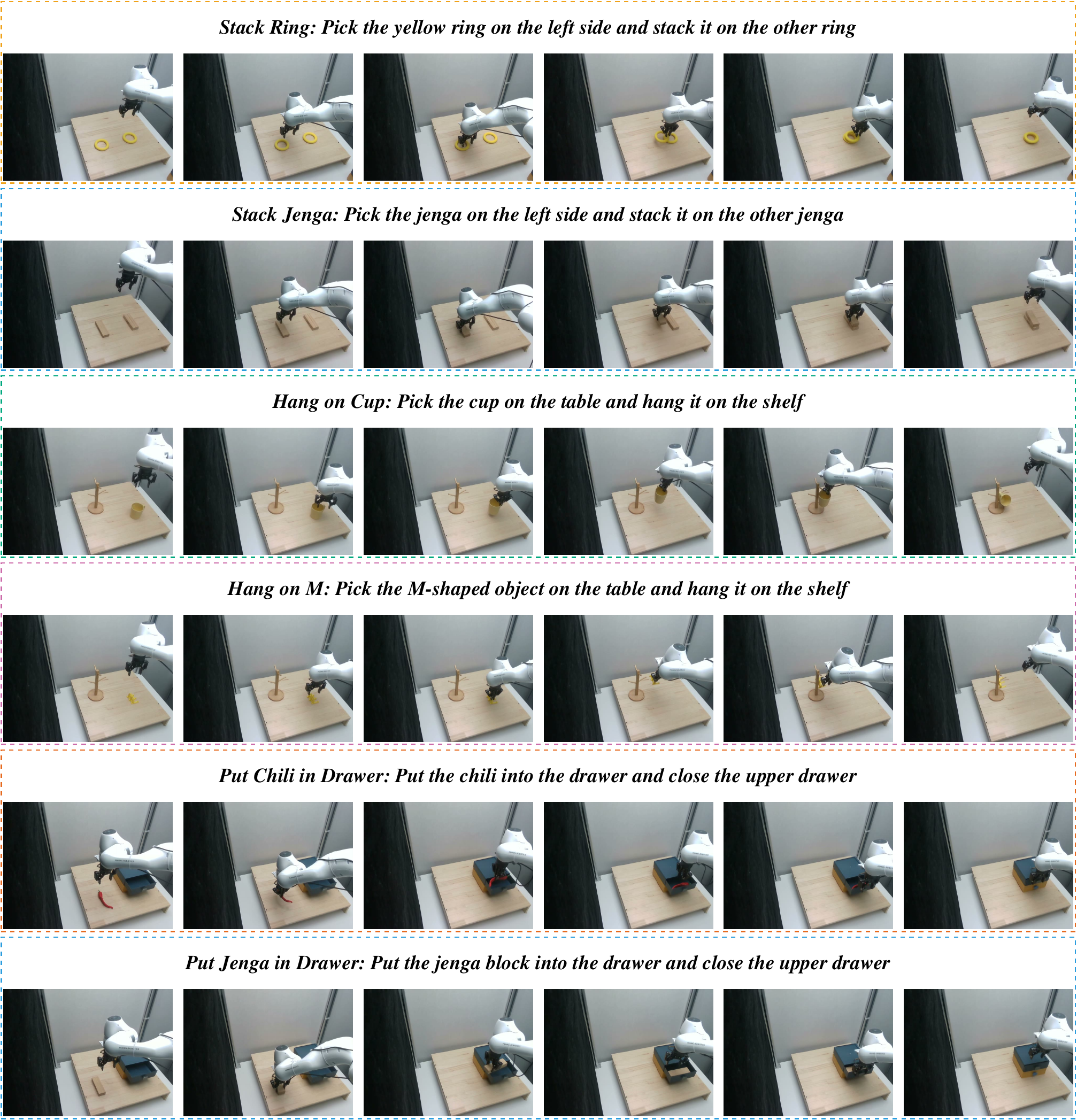}
\caption{Single-arm task execution sequences on the Franka-Research-3 platform.
Each row shows one task, with six frames uniformly sampled from the
corresponding left-view rollout video.}
\label{fig:singlearm-task-sequence}
\end{figure}

The corresponding execution sequences are visualized in
\Cref{fig:singlearm-task-sequence}, where each row contains six frames
uniformly sampled from one rollout video of the task.

\paragraph{Evaluation Protocol.}
For each single-arm task, we fine-tune the policy with 100 task-specific
real-robot demonstrations. After fine-tuning, each task is evaluated over
20 independent real-world trials, and we report the success rate as the
number of successful trials out of 20 in \Cref{tab:realworld}. A stacking
trial succeeds only if the left object is picked and stably stacked on
the target object. A hanging trial succeeds only if the object is picked
and remains hanging on the shelf. A drawer trial succeeds only if the
object is placed inside the drawer and the upper drawer is pushed closed.

\subsection{Dexterous-Hand Real-Robot Experiments}
\label{app:realworld_dexhand}

\paragraph{Task Setup.}
We evaluate dexterous-hand policies on four real-world manipulation
tasks: Stack Toy Tower, Collect Shuttlecocks, Twist off Bottle Cap, and Put Away Clothes.
Each task is specified by a natural-language instruction that defines the
desired manipulation objective. The task instructions are summarized in
\Cref{tab:dex_task_setup}.

\begin{table}[H]
\centering
\scriptsize
\setlength{\tabcolsep}{4pt}
\renewcommand{\arraystretch}{1.15}
\caption{Task instructions used in the real-world evaluation.}
\label{tab:dex_task_setup}
\begin{tabular}{lp{0.68\linewidth}}
\toprule
\textbf{Task} & \textbf{Instruction} \\
\midrule
Stack Toy Tower &
Stack the discs onto the tower pole in order from largest to smallest. \\

Collect Shuttlecocks &
Put all the shuttlecocks into the shuttlecock tube. \\

Twist off Bottle Cap &
Twist off the bottle cap. \\

Put Away Clothes &
Pick up the clothes from the pile on the table and put them into the basket. \\
\bottomrule
\end{tabular}
\end{table}

\begin{figure}[H]
\centering
\includegraphics[width=\linewidth]{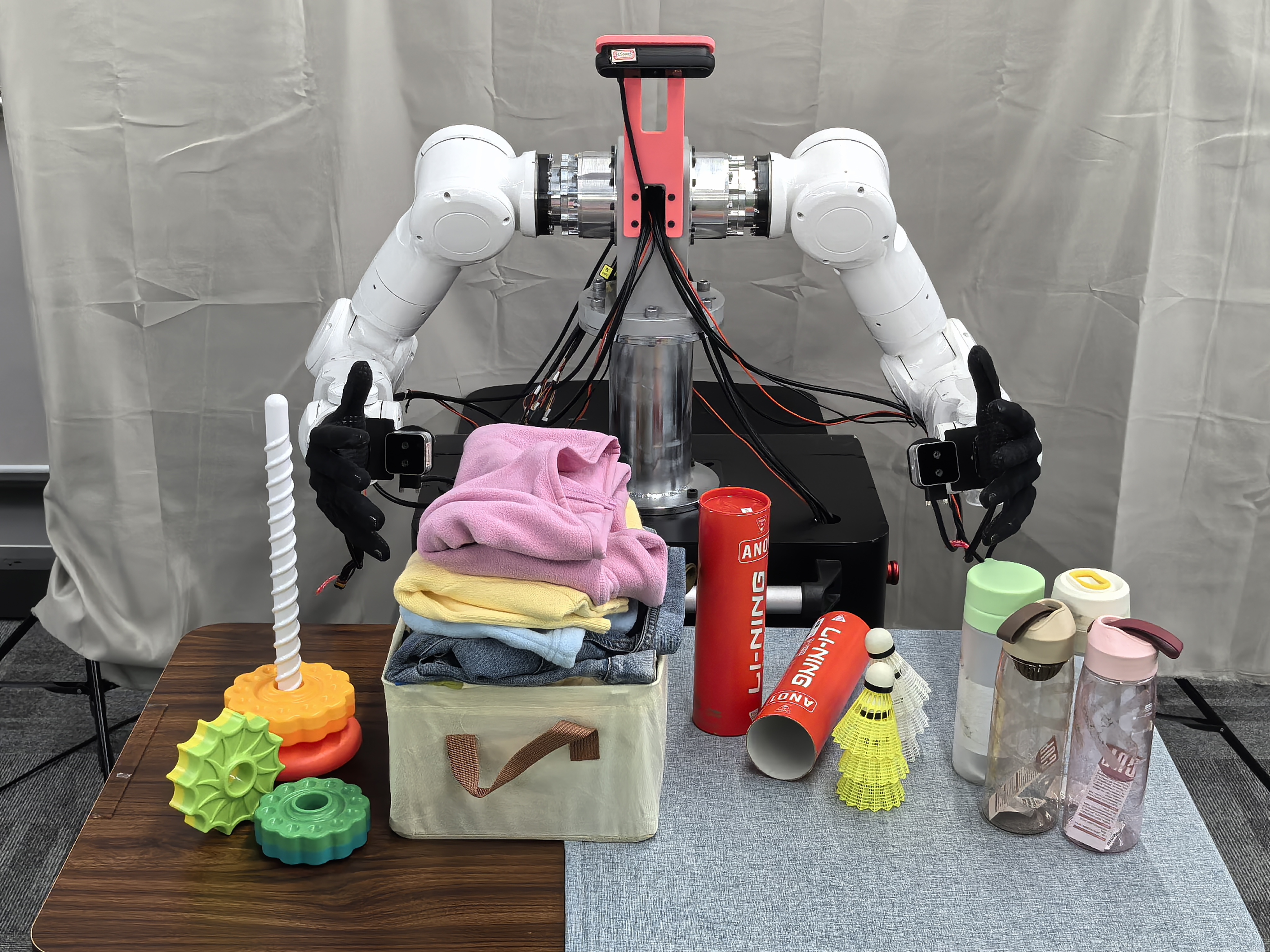}
\caption{Real-world experimental setup. The figure shows the physical
dexterous-hand platform, the workspace, the observation camera, and
representative objects for the four manipulation tasks.}
\label{fig:real_robot_setup}
\end{figure}

All experiments are conducted on a physical dexterous-hand platform.
The robot receives the task instruction and executes the manipulation
autonomously in the corresponding scene. \Cref{fig:real_robot_setup}
shows the physical robot, dexterous hand, workspace, camera viewpoint,
and representative objects used in the evaluation.

\begin{figure*}[t]
\centering
\includegraphics[width=\textwidth]{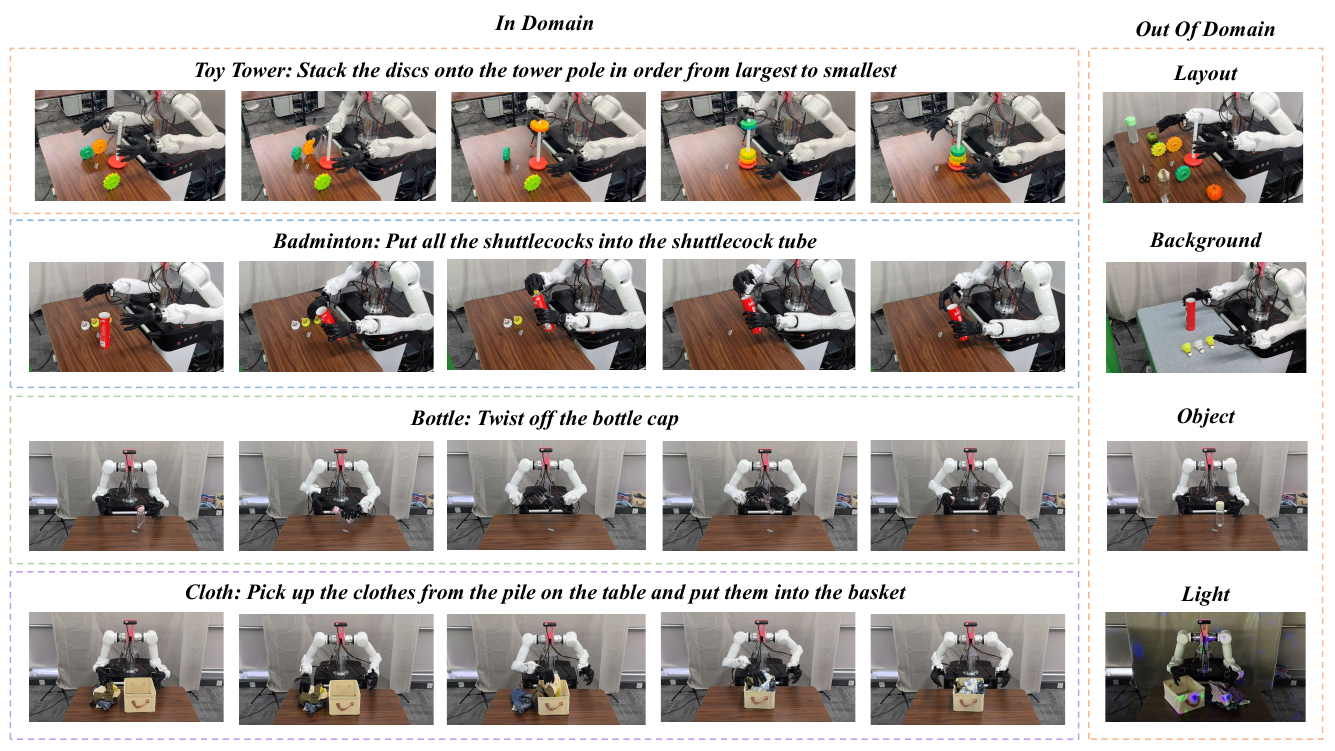}
\caption{Task execution sequences and evaluation conditions. Each row
illustrates the main manipulation stages of one task, while the columns
show the corresponding ID scene and OOD variations. OOD conditions
include changes in object identity, object layout, illumination, and
background appearance.}
\label{fig:task_sequence_ood}
\end{figure*}

For each task, we evaluate the policy under an in-distribution (ID)
condition and several out-of-distribution (OOD) conditions. The OOD
conditions modify one factor at a time, including object identity, object
layout, illumination, or background appearance, while preserving the
task instruction and the overall manipulation objective. The task
execution sequence and the corresponding ID/OOD configurations are
illustrated in \Cref{fig:task_sequence_ood}.

\paragraph{Evaluation Protocol.}
Each task is evaluated over multiple independent trials under both ID and
OOD conditions. We report two complementary metrics: the progress score
(Score) and the final success rate (SR).

\paragraph{Final Success Rate.}
A trial is counted as a final success only when the complete task objective
is achieved. The final success rate is computed as
\[
S_{\mathrm{final}}
=
\frac{N_{\mathrm{success}}}{N_{\mathrm{trial}}}
\times 100\%,
\]
where $N_{\mathrm{success}}$ is the number of trials satisfying the
complete task criterion and $N_{\mathrm{trial}}$ is the total number of
valid trials.

\paragraph{Progress Score.}
The progress score measures the fraction of required manipulation elements
that are successfully completed, regardless of whether the final task
state is achieved. For trial $i$, let $n_i$ denote the number of
successfully completed elements and $m_i$ denote the total number of
elements present in that trial. The progress score is computed as
\[
S_{\mathrm{process}}
=
\frac{\sum_i n_i}{\sum_i m_i}
\times 100\%.
\]

\paragraph{Task-Specific Criteria.}

\begin{itemize}
    \item \textbf{Collect Shuttlecocks.}
    Each trial contains two to four shuttlecocks on the tabletop.
    A trial is counted as a final success only when all shuttlecocks are
    placed into the shuttlecock tube. The progress score is the fraction
    of shuttlecocks successfully placed into the tube.

    \item \textbf{Stack Toy Tower.}
    Each trial contains three discs. A trial is counted as a final
    success only when all three discs are successfully inserted onto the
    tower pole in descending order of size. The progress score is the
    fraction of discs successfully inserted.

    \item \textbf{Put Away Clothes.}
    Each trial contains three to four pieces of clothing. A trial is
    counted as a final success only when all pieces of clothing are placed
    into the basket. The progress score is the fraction of clothing items
    successfully placed into the basket.

    \item \textbf{Twist off Bottle Cap.}
    A trial is counted as a final success when the bottle cap is fully
    twisted off and the robot maintains a stable grasp of the bottle or
    cap. The progress score records whether the cap is successfully
    twisted off, regardless of whether the final stable grasp is achieved.
\end{itemize}

\subsection{Bimanual Real-Robot Experiments}
\label{app:realworld_bimanual}

We conduct a comprehensive evaluation on all 18 tasks of RoboDojo-Real, using the official real-robot evaluation platform provided by the RoboDojo team and covering its three bimanual embodiments: ARX X5, Piper, and Piper X. The evaluation strictly follows the unified protocol defined by the official evaluation team; detailed documentation of the platform and its task specifications is available on the RoboDojo website (\url{https://robodojo-benchmark.com/}) and in the accompanying official documentation (\url{https://robodojo-benchmark.com/doc/real-tasks/}).

\section{Per-Benchmark Simulation Results}
\label{app:simulation_tables}

The tables below report the full per-benchmark scores summarized in \Cref{fig:alpha-score-comparison}, except for LIBERO-Plus, whose scores are already reported in \Cref{tab:liberoplus} of the main text. Within each table the baselines are grouped as VLA or WAM; bold denotes the best value and underline the second best.

\begin{table}[H]
\centering
\small
\setlength{\tabcolsep}{7pt}
\renewcommand{\arraystretch}{1.15}
\captionsetup{justification=centering,singlelinecheck=true}
\caption{\textbf{Evaluation Results on LIBERO.} Bold denotes best values, underline second best.}
\label{tab:libero}
\begin{tabular}{lccccc}
\toprule
 & \textbf{Spatial} & \textbf{Object} & \textbf{Goal} & \textbf{Long} & \textbf{Avg} \\
\midrule
\rowcolor{black!8}
\multicolumn{6}{c}{\textit{\textbf{VLA}}} \\
\midrule
\textbf{OpenVLA}~\citep{kim2024openvla} & 84.7 & 88.4 & 79.2 & 53.7 & 76.5 \\
\rowcolor{black!4}
\textbf{\boldmath $\pi_0$}~\citep{black2024pi_0} & 98.0 & 96.8 & 94.4 & 88.4 & 94.4 \\
\textbf{StarVLA}~\citep{community2026starvla} & 97.8 & 98.6 & 96.2 & 93.8 & 96.6 \\
\rowcolor{black!4}
\textbf{\boldmath $\pi_{0.5}$}~\citep{intelligence2025pi_} & 98.8 & 98.2 & 98.0 & 92.4 & 96.9 \\
\textbf{GR00T-N1.6}~\citep{nvidia2025gr00tn1} & 97.7 & 98.5 & 97.5 & 94.4 & 97.0 \\
\rowcolor{black!4}
\textbf{OpenVLA-OFT}~\citep{kim2025fine} & 97.6 & 98.4 & 97.9 & 94.5 & 97.1 \\
\textbf{X-VLA}~\citep{zheng2026x} & 98.2 & 98.6 & 97.8 & 97.6 & 98.1 \\
\rowcolor{black!4}
\textbf{ABot-M0}~\citep{yang2026abot} & 98.8 & \underline{99.8} & 99.0 & 96.6 & 98.6 \\
\textbf{Being-H0.5}~\citep{luo2026being} & 99.2 & 99.6 & \underline{99.4} & 97.4 & 98.9 \\
\rowcolor{black!4}
\textbf{\textsc{Qwen-RobotManip}}~\citep{yuan2026qwen} & -- & -- & -- & -- & 99.2 \\
\midrule
\rowcolor{black!8}
\multicolumn{6}{c}{\textit{\textbf{WAM}}} \\
\midrule
\textbf{Fast-WAM}~\citep{yuan2026fast} & 98.2 & \textbf{100.0} & 97.0 & 95.2 & 97.6 \\
\rowcolor{black!4}
\textbf{Motus}~\citep{bi2026motus} & 96.8 & \underline{99.8} & 96.6 & 97.6 & 97.7 \\
\textbf{ImageWAM}~\citep{zhang2026imagewam} & 97.2 & 99.2 & 98.8 & \underline{98.4} & 98.4 \\
\rowcolor{black!4}
\textbf{LingBot-VA}~\citep{li2026causal} & 98.5 & 99.6 & 97.2 & \textbf{98.5} & 98.5 \\
\textbf{DiT4DiT}~\citep{ma2026dit4dit} & -- & -- & -- & -- & 98.6 \\
\rowcolor{black!4}
\textbf{Being-H0.7}~\citep{luo2026beingh07} & -- & -- & -- & -- & 99.2 \\
\textbf{ABot-M0.5}~\citep{chen2026abot} & \textbf{100.0} & \underline{99.8} & \underline{99.4} & \underline{98.4} & \textbf{99.4} \\
\midrule
\textbf{\openwamalpha{}} & \underline{99.6} & 99.6 & \textbf{99.8} & 98.2 & \underline{99.3} \\
\bottomrule
\end{tabular}
\end{table}

\begin{table}[H]
\centering
\scriptsize
\setlength{\tabcolsep}{2.6pt}
\renewcommand{\arraystretch}{1.15}
\captionsetup{justification=centering,singlelinecheck=true}
\caption{\textbf{Evaluation Results on VLABench.} Bold denotes best values, underline second best.}
\label{tab:vlabench}
\resizebox{\linewidth}{!}{%
\begin{tabular}{lccc@{\hspace{5pt}}ccc@{\hspace{5pt}}ccc@{\hspace{5pt}}ccc@{\hspace{5pt}}ccc@{\hspace{5pt}}ccc}
\toprule
 & \multicolumn{3}{c}{\textbf{In-dist.}} & \multicolumn{3}{c}{\textbf{Category}} & \multicolumn{3}{c}{\textbf{Commonsense}} & \multicolumn{3}{c}{\textbf{Instruction}} & \multicolumn{3}{c}{\textbf{Texture}} & \multicolumn{3}{c}{\textbf{Avg}} \\
\cmidrule(lr){2-4}\cmidrule(lr){5-7}\cmidrule(lr){8-10}\cmidrule(lr){11-13}\cmidrule(lr){14-16}\cmidrule(lr){17-19}
 & \textbf{SR} & \textbf{PS} & \textbf{IS} & \textbf{SR} & \textbf{PS} & \textbf{IS} & \textbf{SR} & \textbf{PS} & \textbf{IS} & \textbf{SR} & \textbf{PS} & \textbf{IS} & \textbf{SR} & \textbf{PS} & \textbf{IS} & \textbf{SR} & \textbf{PS} & \textbf{IS} \\
\midrule
\rowcolor{black!8}
\multicolumn{19}{c}{\textit{\textbf{VLA}}} \\
\midrule
\textbf{\boldmath $\pi_0$}~\citep{black2024pi_0} & 47.0 & 62.7 & 67.8 & 21.2 & 33.6 & 44.0 & 29.1 & 43.0 & 54.9 & 17.3 & 38.7 & 58.0 & 32.2 & 42.5 & 50.6 & 29.4 & 44.1 & 55.0 \\
\rowcolor{black!4}
\textbf{LoHo-Manip}~\citep{liu2026lohomanip} & 54.0 & -- & -- & 23.0 & -- & -- & 36.0 & -- & -- & 42.0 & -- & -- & 39.0 & -- & -- & 39.0 & -- & -- \\
\textbf{ACoT-VLA}~\citep{zhong2026acot} & -- & 66.1 & 79.8 & -- & 38.9 & \underline{54.1} & -- & 37.8 & 52.3 & -- & 39.6 & 56.8 & -- & 54.6 & 74.6 & -- & 47.4 & 63.5 \\
\rowcolor{black!4}
\textbf{\boldmath $\pi_{0.5}$}~\citep{intelligence2025pi_} & 65.4 & 77.8 & 80.4 & 38.2 & 49.7 & 52.0 & 43.9 & 57.3 & \underline{60.0} & 48.2 & 64.2 & 67.0 & 44.9 & 62.3 & 65.0 & 48.1 & 62.3 & 64.9 \\
\textbf{ERVLA}~\citep{sun2026revisiting} & 69.7 & 81.1 & \underline{84.2} & \underline{47.0} & \underline{61.0} & \textbf{66.4} & 44.0 & 55.0 & 57.2 & \underline{58.0} & \underline{70.2} & \underline{73.8} & 47.4 & 62.3 & 70.6 & 53.2 & 65.9 & \underline{70.4} \\
\rowcolor{black!4}
\textbf{Xiaomi-Robotics-1}~\citep{team2026xiaomi} & 75.6 & 85.0 & 79.8 & \textbf{53.0} & \textbf{66.6} & \textbf{66.4} & 48.4 & 58.3 & 58.2 & 55.8 & 66.8 & 70.2 & \textbf{62.6} & \textbf{74.9} & \underline{74.8} & \textbf{59.1} & \textbf{70.3} & 69.9 \\
\midrule
\rowcolor{black!8}
\multicolumn{19}{c}{\textit{\textbf{WAM}}} \\
\midrule
\textbf{Bridge-WA}~\citep{bai2026bridge} & \underline{78.0} & \underline{85.8} & \textbf{85.0} & 23.0 & 28.8 & 39.0 & \underline{51.1} & \underline{64.4} & \textbf{74.2} & \textbf{67.0} & \textbf{80.3} & \textbf{82.0} & 45.0 & 60.3 & \textbf{76.0} & 52.8 & 64.0 & \textbf{71.2} \\
\midrule
\textbf{\openwamalpha{}} & \textbf{83.4} & \textbf{87.9} & 75.2 & 38.1 & 45.9 & 45.9 & \textbf{58.0} & \textbf{64.8} & 57.9 & 53.8 & 64.5 & 66.5 & \underline{61.4} & \underline{72.9} & 71.6 & \underline{58.9} & \underline{67.2} & 63.5 \\
\bottomrule
\end{tabular}%
}
\end{table}

\begin{table}[H]
\centering
\small
\setlength{\tabcolsep}{9pt}
\renewcommand{\arraystretch}{1.15}
\captionsetup{justification=centering,singlelinecheck=true}
\caption{\textbf{Evaluation Results on RoboTwin2.0-Clean2Random.} Bold denotes best values, underline second best.}
\label{tab:clean2rand}
\begin{tabular}{lccc}
\toprule
 & \textbf{Clean} & \textbf{Randomized} & \textbf{Avg} \\
\midrule
\rowcolor{black!8}
\multicolumn{4}{c}{\textit{\textbf{VLA}}} \\
\midrule
\textbf{StarVLA}~\citep{community2026starvla} & 46.5 & 3.2 & 24.9 \\
\rowcolor{black!4}
\textbf{GR00T-N1.7}~\citep{nvidia2025gr00tn1} & 43.6 & 20.7 & 32.2 \\
\textbf{X-VLA}~\citep{zheng2026x} & 68.0 & 20.9 & 44.5 \\
\rowcolor{black!4}
\textbf{Spatial Forcing}~\citep{li2026spatial} & 77.2 & 26.7 & 52.0 \\
\textbf{ABot-M0}~\citep{yang2026abot} & 70.7 & 36.0 & 53.4 \\
\rowcolor{black!4}
\textbf{\boldmath $\pi_{0.5}$}~\citep{intelligence2025pi_} & 70.7 & 46.0 & 58.4 \\
\textbf{GigaBrain-0.7}~\citep{team2026gigabrain} & 66.8 & \underline{67.9} & 67.4 \\
\rowcolor{black!4}
\textbf{\textsc{Qwen-RobotManip}}~\citep{yuan2026qwen} & \underline{84.7} & \textbf{69.4} & \textbf{77.1} \\
\midrule
\rowcolor{black!8}
\multicolumn{4}{c}{\textit{\textbf{WAM}}} \\
\midrule
\textbf{AHA-WAM}~\citep{cai2026aha} & 64.3 & 3.2 & 33.8 \\
\rowcolor{black!4}
\textbf{Fast-WAM}~\citep{yuan2026fast} & 77.8 & 1.9 & 39.9 \\
\textbf{X-WAM}~\citep{guo2026unified} & 70.0 & 25.8 & 47.9 \\
\rowcolor{black!4}
\textbf{4D-WAM}~\citep{yang20264dwaminfusingspatiotemporalawareness} & 81.5 & 41.8 & 61.7 \\
\midrule
\textbf{\openwamalpha{}} & \textbf{89.4} & 48.7 & \underline{69.0} \\
\bottomrule
\end{tabular}
\end{table}

\begin{table}[H]
\centering
\small
\setlength{\tabcolsep}{9pt}
\renewcommand{\arraystretch}{1.15}
\captionsetup{justification=centering,singlelinecheck=true}
\caption{\textbf{Evaluation Results on RoboTwin2.0-Full.} Bold denotes best values, underline second best.}
\label{tab:robotwin}
\begin{tabular}{lccc}
\toprule
 & \textbf{Clean} & \textbf{Randomized} & \textbf{Avg} \\
\midrule
\rowcolor{black!8}
\multicolumn{4}{c}{\textit{\textbf{VLA}}} \\
\midrule
\textbf{X-VLA}~\citep{zheng2026x} & 72.80 & 72.84 & 72.82 \\
\rowcolor{black!4}
\textbf{\boldmath $\pi_{0.5}$}~\citep{intelligence2025pi_} & 82.70 & 76.80 & 79.75 \\
\textbf{ABot-M0}~\citep{yang2026abot} & 86.06 & 85.08 & 85.57 \\
\rowcolor{black!4}
\textbf{Qwen-VLA}~\citep{wang2026qwen} & 86.10 & 87.20 & 86.65 \\
\textbf{StarVLA}~\citep{community2026starvla} & 88.18 & 88.32 & 88.25 \\
\rowcolor{black!4}
\textbf{Galaxea G0.5}~\citep{liu2026g0} & 93.70 & 92.80 & 93.25 \\
\textbf{\textsc{Qwen-RobotManip}}~\citep{yuan2026qwen} & 93.70 & \underline{94.00} & \underline{93.85} \\
\midrule
\rowcolor{black!8}
\multicolumn{4}{c}{\textit{\textbf{WAM}}} \\
\midrule
\textbf{Motus}~\citep{bi2026motus} & 88.66 & 87.02 & 87.84 \\
\rowcolor{black!4}
\textbf{Fast-WAM}~\citep{yuan2026fast} & 91.90 & 91.80 & 91.85 \\
\textbf{LingBot-VA}~\citep{li2026causal} & 92.93 & 91.55 & 92.24 \\
\rowcolor{black!4}
\textbf{ImageWAM}~\citep{zhang2026imagewam} & 93.20 & 93.56 & 93.38 \\
\textbf{LingBot-VA 2.0}~\citep{zhang2026native} & \underline{93.80} & 93.40 & 93.60 \\
\rowcolor{black!4}
\textbf{ABot-M0.5}~\citep{chen2026abot} & \textbf{94.00} & \textbf{94.20} & \textbf{94.10} \\
\midrule
\textbf{\openwamalpha{}} & 93.74 & 93.46 & 93.60 \\
\bottomrule
\end{tabular}
\end{table}

\begin{table}[H]
\centering
\scriptsize
\setlength{\tabcolsep}{2.6pt}
\renewcommand{\arraystretch}{1.15}
\captionsetup{justification=centering,singlelinecheck=true}
\caption{\textbf{Evaluation Results on RoboDojo.} Bold denotes best values, underline second best.}
\label{tab:robodojo}
\begin{tabular}{lcc@{\hspace{5pt}}cc@{\hspace{5pt}}cc@{\hspace{5pt}}cc@{\hspace{5pt}}cc@{\hspace{5pt}}cc@{\hspace{5pt}}cc}
\toprule
 & \multicolumn{2}{c}{\textbf{Gen-Std}} & \multicolumn{2}{c}{\textbf{Gen-Rand}} & \multicolumn{2}{c}{\textbf{Precision}} & \multicolumn{2}{c}{\textbf{Long-Horizon}} & \multicolumn{2}{c}{\textbf{Memory}} & \multicolumn{2}{c}{\textbf{Open}} & \multicolumn{2}{c}{\textbf{Avg}} \\
\cmidrule(lr){2-3}\cmidrule(lr){4-5}\cmidrule(lr){6-7}\cmidrule(lr){8-9}\cmidrule(lr){10-11}\cmidrule(lr){12-13}\cmidrule(lr){14-15}
 & \textbf{SR} & \textbf{Score} & \textbf{SR} & \textbf{Score} & \textbf{SR} & \textbf{Score} & \textbf{SR} & \textbf{Score} & \textbf{SR} & \textbf{Score} & \textbf{SR} & \textbf{Score} & \textbf{SR} & \textbf{Score} \\
\midrule
\rowcolor{black!8}
\multicolumn{15}{c}{\textit{\textbf{VLA}}} \\
\midrule
\textbf{\boldmath StarVLA-$\alpha$}~\citep{ye2026starvla} & 5.00 & 7.54 & 0.00 & 0.33 & 4.33 & 9.90 & 6.50 & 14.15 & 2.44 & 3.34 & 0.58 & 0.68 & 3.24 & 6.40 \\
\rowcolor{black!4}
\textbf{X-VLA}~\citep{zheng2026x} & 12.00 & 17.90 & 1.00 & 3.04 & 12.00 & 18.32 & 9.75 & 16.53 & 3.56 & 4.76 & 0.50 & 0.55 & 6.52 & 10.13 \\
\textbf{\boldmath $\pi_{0.5}$}~\citep{intelligence2025pi_} & 15.00 & 20.93 & 1.00 & 5.82 & 5.50 & 12.40 & 14.67 & 23.54 & 4.56 & 5.78 & 1.67 & 1.98 & 6.91 & 11.41 \\
\rowcolor{black!4}
\textbf{Spatial Forcing}~\citep{li2026spatial} & 15.00 & 21.25 & 4.00 & 6.98 & 10.58 & 17.33 & 14.58 & 23.26 & 4.11 & 5.43 & 1.58 & 1.78 & 8.04 & 12.38 \\
\textbf{Hy-Embodied-0.5-VLA}~\citep{zhang2026hy} & 17.00 & 21.98 & 0.00 & 1.57 & 8.00 & 13.81 & 14.92 & 25.74 & \underline{12.11} & \underline{13.37} & 0.58 & 0.65 & 8.80 & 13.07 \\
\rowcolor{black!4}
\textbf{Xiaomi-Robotics-1}~\citep{team2026xiaomi} & \textbf{28.00} & \textbf{35.65} & \textbf{6.00} & \textbf{11.44} & \underline{18.83} & \underline{26.69} & 23.67 & \underline{38.39} & 6.56 & 7.81 & \textbf{3.58} & \textbf{3.94} & 13.93 & 20.07 \\
\textbf{Galaxea G0.5}~\citep{liu2026g0} & 20.00 & 26.74 & \textbf{6.00} & \underline{11.16} & \textbf{20.42} & \textbf{28.25} & \textbf{32.25} & \textbf{44.12} & 7.33 & 8.61 & 1.58 & 1.73 & \underline{14.88} & \underline{20.23} \\
\rowcolor{black!4}
\textbf{DM0.5}~\citep{dexmal2026dm05} & 18.00 & 23.49 & 4.00 & 8.06 & 16.75 & 24.82 & 19.50 & 33.70 & \textbf{47.44} & \textbf{47.74} & \underline{2.08} & \underline{2.43} & \textbf{19.34} & \textbf{24.90} \\
\midrule
\rowcolor{black!8}
\multicolumn{15}{c}{\textit{\textbf{WAM}}} \\
\midrule
\textbf{Fast-WAM}~\citep{yuan2026fast} & 2.00 & 4.33 & 0.00 & 0.34 & 0.00 & 1.96 & 5.17 & 9.14 & 3.44 & 3.55 & 0.42 & 0.42 & 2.03 & 3.48 \\
\rowcolor{black!4}
\textbf{AHA-WAM}~\citep{cai2026aha} & 6.00 & 10.32 & 0.00 & 1.26 & 2.42 & 5.86 & 2.67 & 8.61 & 2.78 & 2.97 & 0.83 & 0.88 & 2.39 & 4.82 \\
\textbf{GigaWorld-Policy}~\citep{ye2026gigaworld} & 6.00 & 10.28 & 0.00 & 0.41 & 1.83 & 6.15 & 8.92 & 15.51 & 2.22 & 3.46 & 0.50 & 0.54 & 3.27 & 6.20 \\
\rowcolor{black!4}
\textbf{X-WAM}~\citep{guo2026unified} & 5.00 & 11.24 & 1.00 & 3.54 & 1.83 & 6.72 & 9.08 & 17.47 & 4.67 & 6.32 & 0.25 & 0.57 & 3.83 & 7.69 \\
\midrule
\textbf{\openwamalpha{}} & \underline{25.56} & \underline{33.16} & \underline{4.11} & 8.26 & 9.25 & 18.45 & \underline{25.33} & 34.93 & 9.11 & 10.41 & 1.08 & 1.41 & 11.92 & 17.18 \\
\bottomrule
\end{tabular}
\end{table}

\begin{table}[H]
\centering
\small
\setlength{\tabcolsep}{5pt}
\renewcommand{\arraystretch}{1.15}
\captionsetup{justification=centering,singlelinecheck=true}
\caption{\textbf{Evaluation Results on EBench.} Bold denotes best values, underline second best.}
\label{tab:ebench}
\begin{tabular}{lcc@{\hspace{5pt}}cc@{\hspace{5pt}}cc@{\hspace{5pt}}cc}
\toprule
 & \multicolumn{2}{c}{\textbf{Table Top}} & \multicolumn{2}{c}{\textbf{Simple PnP}} & \multicolumn{2}{c}{\textbf{Long Horizon}} & \multicolumn{2}{c}{\textbf{Overall}} \\
\cmidrule(lr){2-3}\cmidrule(lr){4-5}\cmidrule(lr){6-7}\cmidrule(lr){8-9}
 & \textbf{SR} & \textbf{Score} & \textbf{SR} & \textbf{Score} & \textbf{SR} & \textbf{Score} & \textbf{SR} & \textbf{Score} \\
\midrule
\rowcolor{black!8}
\multicolumn{9}{c}{\textit{\textbf{VLA}}} \\
\midrule
\textbf{StarVLA-OFT}~\citep{community2026starvla} & -- & -- & -- & -- & -- & -- & 0.0 & 0.2 \\
\rowcolor{black!4}
\textbf{\boldmath $\pi_0$}~\citep{black2024pi_0} & 15.7 & 30.0 & 35.0 & 39.0 & 17.0 & 41.0 & 23.6 & 37.0 \\
\textbf{X-VLA}~\citep{zheng2026x} & 8.6 & 24.0 & 50.0 & 54.0 & 6.2 & 25.0 & 23.7 & 36.0 \\
\rowcolor{black!4}
\textbf{InternVLA-A1}~\citep{cai2026internvla} & 4.3 & 11.0 & 43.0 & 47.0 & 17.9 & 46.0 & 23.9 & 36.0 \\
\textbf{\boldmath $\pi_{0.5}$}~\citep{intelligence2025pi_} & 12.9 & 32.0 & 45.0 & 50.0 & 18.1 & 39.0 & 27.1 & 41.0 \\
\rowcolor{black!4}
\textbf{GigaBrain-0.7}~\citep{team2026gigabrain} & -- & -- & -- & -- & -- & -- & 33.3 & 46.0 \\
\textbf{\textsc{Qwen-RobotManip}}~\citep{yuan2026qwen} & \textbf{50.0} & \textbf{70.0} & \underline{56.5} & \underline{60.0} & \underline{29.9} & \underline{55.0} & \underline{45.6} & \underline{60.0} \\
\midrule
\rowcolor{black!8}
\multicolumn{9}{c}{\textit{\textbf{WAM}}} \\
\midrule
\textbf{Fast-WAM}~\citep{yuan2026fast} & -- & -- & -- & -- & -- & -- & 4.7 & 7.6 \\
\midrule
\textbf{\openwamalpha{}} & \underline{30.0} & \underline{44.2} & \textbf{67.5} & \textbf{72.0} & \textbf{44.3} & \textbf{72.6} & \textbf{49.4} & \textbf{64.7} \\
\bottomrule
\end{tabular}
\end{table}

\begin{table}[H]
\centering
\small
\setlength{\tabcolsep}{7pt}
\renewcommand{\arraystretch}{1.15}
\captionsetup{justification=centering,singlelinecheck=true}
\caption{\textbf{Evaluation Results on RoboCasa365.} Bold denotes best values, underline second best.}
\label{tab:robocasa365}
\begin{tabular}{lcccc}
\toprule
 & \textbf{Atomic} & \textbf{Comp.-Seen} & \textbf{Comp.-Unseen} & \textbf{Avg} \\
\midrule
\rowcolor{black!8}
\multicolumn{5}{c}{\textit{\textbf{VLA}}} \\
\midrule
\textbf{Diffusion Policy}~\citep{chi2025diffusion} & 15.7 & 0.2 & 1.3 & 6.1 \\
\rowcolor{black!4}
\textbf{\boldmath $\pi_0$}~\citep{black2024pi_0} & 36.3 & 5.2 & 0.7 & 15.0 \\
\textbf{\boldmath $\pi_{0.5}$}~\citep{intelligence2025pi_} & 39.6 & 7.1 & 1.2 & 16.9 \\
\rowcolor{black!4}
\textbf{GR00T-N1.5}~\citep{nvidia2025gr00tn1} & 50.7 & 14.8 & 2.7 & 23.9 \\
\textbf{\textsc{Qwen-RobotManip}}~\citep{yuan2026qwen} & 68.6 & 20.1 & \underline{14.9} & 35.9 \\
\rowcolor{black!4}
\textbf{RLDX-1}~\citep{kim2026rldx} & 67.6 & 27.9 & 8.5 & 36.0 \\
\textbf{Xiaomi-Robotics-1}~\citep{team2026xiaomi} & \textbf{80.2} & \textbf{57.1} & \textbf{32.1} & \textbf{57.4} \\
\midrule
\rowcolor{black!8}
\multicolumn{5}{c}{\textit{\textbf{WAM}}} \\
\midrule
\textbf{GigaWorld-Policy}~\citep{ye2026gigaworld} & 44.4 & 11.8 & 2.9 & 20.7 \\
\rowcolor{black!4}
\textbf{ABot-M0.5}~\citep{chen2026abot} & \underline{75.9} & \underline{38.3} & 2.7 & \underline{40.4} \\
\midrule
\textbf{\openwamalpha{}} & 69.7 & 32.1 & 8.9 & 38.2 \\
\bottomrule
\end{tabular}
\end{table}

\begin{table}[H]
\centering
\small
\setlength{\tabcolsep}{9pt}
\renewcommand{\arraystretch}{1.15}
\captionsetup{justification=centering,singlelinecheck=true}
\caption{\textbf{Evaluation Results on RoboCasa-GR1.} Bold denotes best values, underline second best.}
\label{tab:robocasa-gr1}
\begin{tabular}{lc}
\toprule
 & \textbf{SR (\%)} \\
\midrule
\rowcolor{black!8}
\multicolumn{2}{c}{\textit{\textbf{VLA}}} \\
\midrule
\textbf{\boldmath $\pi_0$}~\citep{black2024pi_0} & 13.6 \\
\rowcolor{black!4}
\textbf{\boldmath $\pi_{0.5}$}~\citep{intelligence2025pi_} & 37.0 \\
\textbf{GR00T-N1.5}~\citep{nvidia2025gr00tn1} & 48.0 \\
\rowcolor{black!4}
\textbf{StarVLA}~\citep{community2026starvla} & 48.8 \\
\textbf{GR00T-N1.6}~\citep{nvidia2025gr00tn1} & 49.9 \\
\rowcolor{black!4}
\textbf{VP-VLA}~\citep{wang2026vp} & 53.8 \\
\textbf{Being-H0.5}~\citep{luo2026being} & 53.9 \\
\rowcolor{black!4}
\textbf{Qwen-VLA-Instruct}~\citep{wang2026qwen} & 56.7 \\
\textbf{RLDX-1}~\citep{kim2026rldx} & 58.7 \\
\rowcolor{black!4}
\textbf{PhysBrain 1.0}~\citep{physbrain} & \textbf{64.5} \\
\midrule
\rowcolor{black!8}
\multicolumn{2}{c}{\textit{\textbf{WAM}}} \\
\midrule
\textbf{UWM}~\citep{zhu2025unified} & 20.0 \\
\rowcolor{black!4}
\textbf{Being-H0.7}~\citep{luo2026beingh07} & 49.2 \\
\textbf{DiT4DiT}~\citep{ma2026dit4dit} & 50.8 \\
\rowcolor{black!4}
\textbf{LDA-1B}~\citep{lyu2026lda} & 55.4 \\
\midrule
\textbf{\openwamalpha{}} & \underline{60.5} \\
\bottomrule
\end{tabular}
\end{table}

\end{document}